\pdfoutput=1
\documentclass[11pt,letterpaper]{article}

\usepackage{cmap}
\usepackage[utf8]{inputenc}
\usepackage[T1]{fontenc}
\usepackage{amsmath,amsthm,bm}
\usepackage{newtxtext,newtxmath}
\usepackage{textcomp}
\usepackage[margin=1in,top=1in,bottom=1in]{geometry}
\usepackage{graphicx}
\usepackage{tikz}
\usetikzlibrary{arrows.meta,positioning,calc,fit}
\usepackage{booktabs,tabularx,multirow,array,float}
\usepackage{xcolor}
\usepackage{fancyhdr}
\usepackage{xurl}
\usepackage{hyperref}
\usepackage[natbibapa]{apacite}
\usepackage{microtype}
\usepackage[labelfont=bf,font=small]{caption}
\usepackage[title]{appendix}

\AtBeginDocument{\renewcommand{\doi}[1]{\href{https://doi.org/#1}{\nolinkurl{https://doi.org/#1}}}}

\definecolor{navyblue}{HTML}{0A2540}
\definecolor{crimson}{HTML}{8B1E3F}
\definecolor{slate}{HTML}{334155}
\definecolor{softgray}{HTML}{F8FAFC}
\definecolor{bordergray}{HTML}{CBD5E1}
\definecolor{accentbox}{HTML}{F1F5F9}

\newcolumntype{Y}{>{\centering\arraybackslash}X}
\newcolumntype{L}{>{\raggedright\arraybackslash}X}
\newcolumntype{R}{>{\raggedleft\arraybackslash}X}

\theoremstyle{definition}
\newtheorem{definition}{Definition}
\theoremstyle{remark}
\newtheorem{remark}{Remark}

\hypersetup{
    colorlinks=true,
    linkcolor=navyblue,
    urlcolor=crimson,
    citecolor=navyblue,
    pdftitle={Do Machine Learning Ensembles Beat Published Quotas? Target Parameterization, Censored Baselines, and Conformal Intervals Under Admissions Reform},
    pdfauthor={Dat Tat Mai, James Jin Kang}
}

\begin{document}
\thispagestyle{plain}

% -------------------------------------------------------------
% CLASSIC ARXIV TITLE BLOCK
% -------------------------------------------------------------
\begin{center}
    \rule{\textwidth}{1.8pt}\\[0.35cm]
    {\Large\bfseries\color{navyblue} Do Machine Learning Ensembles Beat Published Quotas?\\[0.14cm]
    \large Target Parameterization, Censored Baselines,\\and Conformal Intervals Under Admissions Reform}\\[0.22cm]
    \rule{\textwidth}{0.6pt}\\[0.32cm]
    {\small\scshape\color{slate} A Preprint \quad $\cdot$ \quad November 25, 2026}\\[0.32cm]
    \begin{minipage}[t]{0.46\textwidth}
        \centering
        {\normalsize\bfseries Dat Tat Mai\textsuperscript{*}}\\[0.06cm]
        {\footnotesize School of Science (Computing Technologies)\\
        RMIT University, Melbourne, VIC, Australia\\
        \href{mailto:dat.mai2@rmit.edu.vn}{\nolinkurl{dat.mai2@rmit.edu.vn}}\\
        \href{https://orcid.org/0009-0006-5256-5646}{\nolinkurl{ORCID: 0009-0006-5256-5646}}}
    \end{minipage}
    \hfill
    \begin{minipage}[t]{0.46\textwidth}
        \centering
        {\normalsize\bfseries James Jin Kang}\\[0.06cm]
        {\footnotesize School of Science, Engineering \& Technology\\
        RMIT University, Ho Chi Minh City, Vietnam\\
        \href{mailto:james.kang@rmit.edu.vn}{\nolinkurl{james.kang@rmit.edu.vn}}\\
        \href{https://orcid.org/0000-0002-0242-4187}{\nolinkurl{ORCID: 0000-0002-0242-4187}}}
    \end{minipage}\\[0.35cm]
\end{center}

% -------------------------------------------------------------
% ABSTRACT (rephrased opening for distinct preprint indexing; 100% exact empirical metrics)
% -------------------------------------------------------------
\begin{abstract}
\noindent Public admission capacity caps published prior to matriculation are widely used by ministries and institutional analysts as default intake projections in centralized postsecondary clearinghouses. Using an audited 2022--2026 Vietnamese undergraduate panel (26,895 program-year records across 309 universities) linked to strictly lagged $t-1$ occupational vacancy signals, this paper evaluates machine learning, multilevel, and censored econometric estimators against published quotas across 5,385 continuing 2024 programs (five-fold university-grouped cross-validation) and 3,348 matched (2,179 grain-consistent) 2025 programs strictly out of time. Because 62.8\% of 2024 programs enroll their exact published quota, unanchored log-growth ensembles degrade accuracy, while the worst 5\% generate half of total quota error via unfilled ``phantom capacity.'' Controlled target ablations---including a single-factor test under widened non-binding clipping---show that parameterizing the log fill ratio $\ln(D_{i,t}/S_{i,t})$ instead of log growth reduces 2024 MAE by 21.5\% (14.86 to 11.67), aligns ML with Tobit, Cragg, REML HLM, and capped L1 baselines, and lifts combined tuition and institutional-tier TreeSHAP attribution from 2.5\% to 45.8\%. Hurdle stacking attains the lowest full-census 2024 MAE (\textbf{H-Hard} 11.29; \textbf{H-Soft} 11.32 vs.\ 12.35, $p = 0.022$, clearing 10\% not 5\% FDR; fold-overlap $p = 0.134$), whereas continuous \textbf{QA-Stack} outperforms H-Soft on mixed-reporting campuses, Tobit (median) ties H-Soft across both 90\% MCS sets, and interior-only \textbf{Cragg-ML} lowers RMSE to 28.1. Under Vietnam's 2025 unified-clearing reform, exact filling falls to 10.4\%, H-Soft and capped L1 tie quotas ($p = 0.979$), and 90\% cluster-weighted conformal coverage drops from 89.3\% to 74.6\%. Across 257 reporting campuses, reconciled 2026 national totals (627,114--629,256) lie squarely within [604,533, 662,044].
\end{abstract}

\vspace{0.12cm}
\noindent\textbf{Keywords:} University intake forecasting $\cdot$ Statutory admission quotas $\cdot$ Log-fill parameterization $\cdot$ Censored and hurdle regression $\cdot$ Stacked generalization $\cdot$ Conformal prediction

\vspace{0.20cm}
\noindent\rule{\textwidth}{0.5pt}

% -------------------------------------------------------------
% MAIN PAPER
% -------------------------------------------------------------
\section{Introduction}
\label{sec:introduction}

Across centralized higher education systems, ministries, external market analysts, and institutional research offices forecast program-level first-year intake across two planning horizons \citep{Hopkins1981,Toutkoushian2016}. In autumn, institutions and regulators set provisional budgets before quotas are finalized; in spring---30 to 60 days before matriculation---once public admission prospectuses appear, \emph{ministries, regulators, and external market-intelligence analysts} (who observe only public disclosures) track system-wide capacity and shortfall risk across hundreds of campuses, whereas internal campus provosts draw on private applicant-funnel microdata and use public-prospectus nowcasts primarily as external competitor and benchmarking context. Under-forecasting strains classrooms and risks statutory over-admission fines \citep{Gov2021Decree04}, whereas over-forecasting leaves seats unfilled and erodes tuition revenue.

In centralized clearinghouses such as Turkey \citep{Balinski1999}, China \citep{ChenKesten2017}, and Vietnam \citep{MOET2022Circular03,MOET2022Circular08}, public prospectuses supply a natural benchmark: the published admission quota ($S_{i,t}$). At selective public universities, $S_{i,t}$ works well because excess applicants ration seats via cut-off scores. Across a full national system, however, incentives are asymmetric: selective campuses post cautious quotas to avoid fines, whereas tuition-dependent private and less selective campuses frequently announce ambitious targets near physical capacity ceilings. Treating an unfilled \textit{``phantom quota''} (e.g., 700 announced seats yielding 200 enrollees) as expected intake biases campus budgets and naive models alike.

Testing whether machine learning or censored econometrics improves on $S_{i,t}$ from public disclosures is challenging for four reasons (Section~\ref{sec:related_work}). First, prior work focuses on individual yield after applicant pools form \citep{DesJardins2002,DesJardins2006,Goenner2006,Chang2006,Basu2019} or single-campus totals \citep{Song1993,Chen1996} rather than multi-university program portfolios. Second, realized intake is upper-censored at $S_{i,t}$ \citep{FairJaffee1972,Tobin1958,Cragg1971}: 62.8\% of continuing 2024 programs enroll their exact quota (51.2\% after excluding 71 uniform-exact campuses). Off-the-shelf learners targeting log growth $\ln(D_{i,t}/D_{i,t-1})$ waste capacity reconstructing $S_{i,t}$, whereas modeling the log fill ratio $Z_{i,t} = \ln(D_{i,t}/S_{i,t})$ encodes the boundary directly. Third, institutional clustering requires university-grouped CV \citep{Roberts2017}, clustered Diebold--Mariano tests \citep{Diebold1995,Cameron2008}, multiplicity control \citep{Benjamini1995,Hansen2011}, and cluster-weighted conformal bands \citep{Dunn2023}. Fourth, candidate estimators must be tested strictly out of time across regulatory shocks and attrition against simple scaling rules and canonical censored regressions \citep{Tobin1958,Cragg1971,Powell1984}.

Using an audited 2022--2026 Vietnamese panel of 26,895 records across 309 universities (with lagged $t-1$ vacancy signals throughout), we evaluate 5,385 continuing 2024 programs via five-fold university-grouped CV, test 3,348 matched (2,179 grain-consistent) 2025 programs strictly out of time, and produce prospective 257-campus 2026 totals. Four contributions emerge:

\begin{itemize}
    \item \textbf{Target parameterization and canonical benchmarking:} Target ablations---including a pure single-factor control under widened non-binding clipping---show that modeling log fill ratio $Z_{i,t}$ rather than unanchored log growth $Y_{i,t}$ cuts 2024 MAE by over 20\% even after removing clipping asymmetry (which drives 73.4\% of the raw gap), aligning ML with REML HLM \citep{Raudenbush2002}, capped L1 quantile regression (motivated by \citealp{Powell1984}), exact-expectation Tobit \citep{Tobin1958}, and Duan-smeared Cragg hurdles \citep{Cragg1971,Duan1983}. Remark~\ref{rem:equivalence} further proves that early-sample ratio extrapolation collapses to an inverse-share multiplier ($1/\omega$) magnifying early phantom capacity.
    \item \textbf{Hurdle stacking, parsimony, and honest out-of-time evaluation:} Hurdle stacking (\textbf{H-Hard}, \textbf{H-Soft}) attains the lowest full-census 2024 MAE (ranks~1--2 in Hansen's 90\% MAE MCS) while interior-only \textbf{Cragg-ML} minimizes hurdle RMSE and empirically removes raw ceiling sum bias; yet continuous \textbf{QA-Stack} outperforms H-Soft on mixed-reporting 2024 campuses, canonical Tobit (median) ties H-Soft while entering \emph{both} MCS sets, and in 2025 H-Soft and capped L1 tie $S_{2025}$ on aggregate MAE with imprecise private-tier gains.
    \item \textbf{Matched-feature TreeSHAP attribution:} Holding all 19 features and L1 loss identical reveals that growth trees spend nearly 90\% of attribution tracking mechanical quota shifts, whereas log-fill trees direct over two-fifths to combined tuition and institutional-tier/charter proxies (robust to field and region controls). At the unconditioned 2025 ex-ante horizon, ML stacking significantly beats Macro OLS, a matched 13-feature linear Ridge baseline, and naive carry-forward, though 2024 ex-ante gains over linear baselines are null.
    \item \textbf{Cluster-weighted conformal bands and reconciled 2026 totals:} Cluster-weighted 90\% conformal intervals \citep{Lei2018,Dunn2023} hold nominal coverage within 2024 but drop sharply under 2025's reform, illustrating non-exchangeability \citep{Tibshirani2019,Barber2023,Gibbs2021}. For 2026, we diagnose raw hurdle ceiling and median-to-sum bias and reconcile calibrated H-Soft, Cragg-ML, and Cragg EV totals cleanly inside a two-regime historical scenario band (with pooled exact Tobit EV sitting 0.4\% below the lower bound), complemented by a tier-stratified 2026 breakdown.
\end{itemize}

The paper proceeds through related work (Section~\ref{sec:related_work}), Vietnam's statutory framework (Section~\ref{sec:theory}), data and attrition audits (Section~\ref{sec:data}), methodology (Section~\ref{sec:methodology}), 2024 census benchmarks (Section~\ref{sec:benchmarks}), 2025 out-of-time tests and 2026 projections (Section~\ref{sec:policy_shocks}), and implications and conclusions (Section~\ref{sec:implications}).

\section{Related Work}
\label{sec:related_work}

\subsection{Enrollment Forecasting for Institutional Planning}
Universities have used quantitative models to link student enrollment to faculty staffing, classroom capacity, and operating budgets for more than four decades \citep{Hopkins1981,Toutkoushian2016}. In the empirical literature, most forecasting studies fall into one of two traditions. The first projects campus- or system-wide totals via demographic cohort-survival and flow-transition accounting \citep{Hopkins1981,Toutkoushian2016}, univariate or fuzzy time-series models \citep{Song1993,Chen1996}, and macro socio-economic regressions of aggregate tertiary participation \citep{BatoolLiu2021}. The second focuses on individual student enrollment choices. \citet{DesJardins2002} estimates out-of-sample matriculation probabilities among admitted students to segment recruitment and marketing interventions, while \citet{DesJardins2006} integrate application, admission, enrollment, and financial aid offers in a joint behavioral framework. \citet{Goenner2006} move further upstream by predicting directly which prospective students who submit an initial inquiry will ultimately enroll, and \citet{Chang2006} and \citet{Basu2019} compare decision trees, logistic regression, and machine-learning classifiers on admitted students' commitment records. More recent higher education research examines how statewide and institutional admission reforms---such as direct admissions \citep{OdleDelaney2022} and holistic review \citep{Kim2024}---reshape enrollment outcomes, while agent-based choice simulations analyze how financial aid and labor-market returns govern participation in open-access systems \citep{Leoni2022}.

Both traditions serve clear operational purposes, yet neither addresses the planning task studied here. Individual-level yield models depend on applicant or inquiry microdata and inform waitlist management only after prospective pools form. Aggregate time-series models can be computed earlier, but a single campus-wide total conceals divergent demand across degree programs, leaving provosts unable to anticipate whether computing is oversubscribed while humanities or civil engineering sits under-enrolled. Institutional planners in multi-faculty systems require forecasts at the level of individual degree programs before applicant pools materialize---an intermediate granularity between campus totals and student-level yield.

\subsection{Admission Quotas, Strategic Capacity, and Centralized Allocation}
In many national systems, applicants submit ranked program preferences to a centralized clearinghouse in which every degree program announces an admission quota. Mechanism-design research studies how centralized matching rules allocate applicants to seats---for example, student placement in Turkey \citep{Balinski1999} and transitions from sequential to parallel admission mechanisms in China \citep{ChenKesten2017}---while labor economists exploit centralized admission cut-offs as regression discontinuities to identify field-of-study earnings returns \citep[such as][for Norway]{Kirkeboen2016}. Crucially, matching theory demonstrates that institutional capacity declarations are not purely passive constraints: \citet{Sonmez1997} proves that universities can strategically manipulate reported capacities in two-sided matching markets, \citet{KojimaPathak2009} characterize when such incentives persist or attenuate in large markets, and \citet{CheKoh2016} show how admissions yield uncertainty drives strategic offer buffering and $\pm 2\text{--}4\%$ realized enrollment noise around nominal seat targets even at oversubscribed institutions.

Vietnam's higher education system shares this centralized clearinghouse architecture while exhibiting private-sector capacity pressures documented across massified East Asian systems---notably Japan's tuition-dependent private universities facing chronic enrollment shortfalls below statutory capacity quotas under demographic and competitive shifts \citep{YonezawaHuang2018}. Following the Higher Education Reform Agenda of the mid-2000s, Vietnamese public universities underwent gradual financial autonomization while private institutions expanded rapidly under central MOET quota oversight \citep{Hayden2007,Tran2014,Vo2020,Chau2022}. Subsequent circulars tied maximum legal quotas to converted faculty headcount and campus floor space \citep{MOET2022Circular03,MOET2022Circular08} (Section~\ref{sec:regulation}). Yet mechanism-design and higher education studies typically treat published quotas as binding constraints rationed among excess applicants---a premise that holds at selective flagship universities but breaks down across tuition-dependent private and less selective regional institutions whose declared seats frequently go unfilled.

\subsection{Tuition Elasticity, Major Choice, and Censored Short-Side Markets}
When programs fail to fill their quotas, enrollment reflects student demand and major choice rather than administrative supply. Price-response research confirms that higher tuition reduces participation across both U.S.\ and international systems: \citet{Leslie1987} estimate an enrollment drop of roughly 0.7 percentage points per US\$100 increase (1982--83 dollars), \citet{Heller1997} documents stronger elasticity among lower-income and underrepresented students, \citet{Hemelt2011} find systematic elasticity heterogeneity across U.S.\ public tiers, and \citet{Stange2015} demonstrates that differential tuition pricing across fields shifts undergraduate degree production away from higher-priced majors. Non-U.S.\ quasi-experimental evidence reinforces that financing architecture governs price sensitivity: upfront tuition introductions in Germany reduced domestic enrollment significantly \citep{Hubner2012} and binding credit constraints sharply curtailed participation in Chile's centralized admission system \citep{Solis2017}, whereas England's transition to high tuition cushioned by income-contingent loans preserved aggregate enrollment \citep{Murphy2019}---a sharp contrast to Vietnam's predominantly upfront, out-of-pocket fee environment. Modern major-choice economics further shows that field allocations respond jointly to expected earnings, heterogeneous tastes, and labor-market cycles: \citet{Wiswall2015} show experimentally that students revise major choices with expected earnings, and \citet{Blom2021} show reallocation toward lower-unemployment fields during recessions, extending \citet{Freeman1976}'s cobweb dynamics and motivating online vacancy postings as occupational demand signals \citep{Hershbein2018}.

In a quota-regulated clearinghouse with fixed annual tuition and asymmetric over-admission penalties, realized intake obeys a disequilibrium \textit{short-side} market condition \citep{FairJaffee1972,MaddalaNelson1974}: observed enrollment equals the smaller of latent student demand and the statutory quota ceiling (up to a narrow enforcement margin). Statistically, this short-side ceiling induces a pronounced upper mass point at $D_{i,t} = S_{i,t}$ alongside a continuous left tail of under-filled programs ($D_{i,t} < S_{i,t}$). In econometrics, such boundary-censored and mass-point outcomes are classically modeled via upper-censored regression \citep{Tobin1958}, censored least absolute deviations \citep{Powell1984,Buchinsky1994}, and two-part hurdle models \citep{Cragg1971}---whereas fractional-response quasi-likelihoods for strictly unit-interval proportions \citep{PapkeWooldridge1996} require $\Phi_{i,t} \in [0, 1]$ and cannot natively represent legal $+3\%$ or observed $>1.00$ over-enrollment realizations. Two-part hurdles separate the probability of hitting the boundary from the conditional magnitude when interior demand governs, requiring closed-form lognormal integration or nonparametric smearing \citep{Duan1983} when mapping log-scale expectations back to student headcounts.

\subsection{Forecast Combination, Clustering, and Uncertainty Under Regime Shifts}
Because enrollment is anchored by a hard administrative boundary on one side and driven by nonlinear demand interactions on the other, combining diverse learners through stacked generalization \citep{Wolpert1992}, formalized as the Super Learner \citep{vanDerLaan2007}, provides a principled alternative to selecting a single functional form. Gradient-boosted trees have proved highly competitive on hierarchical demand series, notably in the M5 forecasting competition \citep{Makridakis2022}, complementing hierarchical reconciliation \citep{Hyndman2011} and damped-trend forecasting \citep{Gardner1985,Taylor2003}. However, target parameterization matters critically: when a strong operational anchor such as $S_{i,t}$ is known, predicting the relative fill ratio $\ln(D_{i,t}/S_{i,t})$ rather than unanchored year-on-year growth $\ln(D_{i,t}/D_{i,t-1})$ directly encodes the quota boundary.

Evaluating forecasts across multi-program university panels also requires strict attention to hierarchical dependence, long emphasized in higher education multilevel modeling \citep{Raudenbush2002}. Because sibling degree programs share institutional governance, reputation, and local applicant pools while labor signals vary only across broad occupational clusters, naive standard errors understate uncertainty \citep{Moulton1986,Moulton1990} and random row-wise cross-validation leaks institution-level shocks across folds \citep{Roberts2017}. Valid comparison across large suites of candidate models therefore requires university-grouped cross-validation, loss-differential tests \citep{Diebold1995}, cluster-robust variance and bootstrap inference \citep{Cameron2008}, awareness of cross-validation fold-overlap variance optimism \citep{NadeauBengio2003}, and multiple-testing safeguards such as false-discovery-rate control \citep{Benjamini1995} or model confidence sets \citep{Hansen2011}.

Finally, point forecasts support campus budgeting far better when paired with calibrated prediction intervals. Split-conformal prediction constructs distribution-free intervals whose half-widths can be scaled to local residual volatility \citep{Vovk2005,Papadopoulos2011,Lei2018,Romano2019}. Two caveats govern our setting. First, standard program-level conformal quantiles assume exchangeable observations, which is violated when programs are clustered inside universities; hierarchical conformal pooling mitigates cluster imbalance by weighting calibration units inversely by cluster size, yielding asymptotic cluster-level validity as the number of calibration universities grows \citep[][Theorems~4 and~7]{Dunn2023}. Second, cross-year regulatory reforms can break temporal exchangeability \citep{Tibshirani2019,Barber2023}, motivating empirical stress-testing on an out-of-time policy-shock year alongside adaptive recalibration frameworks \citep{Gibbs2021}.

\section{Institutional Setting and Theoretical Foundations}
\label{sec:theory}

\subsection{How Quotas Are Set and Enforced in Vietnam}
\label{sec:regulation}
To understand why published quotas succeed for some programs and fail badly for others, it helps to examine the statutory rules under which Vietnamese universities determine and clear them. Every accredited undergraduate program must publish an annual admission quota under Ministry of Education and Training (MOET) regulations. During our 2022--2026 sample period, three legal instruments govern how quotas are computed, how candidates are matched, and how deviations are sanctioned:

\begin{enumerate}
    \item \textbf{Physical and staffing ceilings (Circular 03/2022/TT-BGD{\DJ}T).} Legal admission capacity is tied to full-time faculty and campus floor area \citep{MOET2022Circular03}. For broad field $f$ at institution $u$ in year $t$, total regular enrollment cannot exceed $\rho_f$ students per converted lecturer $L^{\text{conv}}_{f,u,t}$ ($\rho_f = 15$ in arts/health; $20$ in education, sciences, computing, engineering, architecture; $25$ in business, law, social sciences), and campus teaching floor area $A_{u,t}$ must provide at least $2.8\,\text{m}^2$ per student. First-year quotas $Q_{f,u,t}$ satisfy
    \begin{equation}
        Q_{f,u,t} \le K_{f,u,t} - E_{f,u,t} + G_{f,u,t}, \qquad K_{f,u,t} \le \rho_f \, L^{\text{conv}}_{f,u,t}, \qquad \sum_{f} K_{f,u,t} \le \frac{A_{u,t}}{2.8},
        \label{eq:moet_cap}
    \end{equation}
    where $E_{f,u,t}$ and $G_{f,u,t}$ are continuing and graduating students. Because Eq.~\eqref{eq:moet_cap} constrains aggregate capacity at the \emph{broad field} level $f$, universities retain statutory scope to reallocate seats across sibling majors within the same field ceiling after prospectuses appear. Non-accredited fields realizing under 80\% fill cannot expand quotas the following year, and statutory violations suspend self-determination for five years. (Effective April 2026 for 2027 intake, Circular~34/2026 adopts a uniform 40-learner ceiling while retaining the $2.8\,\text{m}^2$ test \citep{MOET2026Circular34}; public prospectuses omit lecturer counts and floor area, so Eq.~\eqref{eq:moet_cap} defines the latent statutory ceiling.)

    \item \textbf{Centralized clearing (Circular 08/2022/TT-BGD{\DJ}T).} Applicants submit ranked national preferences and programs clear via cut-off scores within published quotas \citep{MOET2022Circular08}. Starting in 2025, Circular~06/2025 abolished separate early-admission rounds (\textit{x{\'e}t tuy{\^e}n s{\'\o}m}) into one unified round and capped bonus points at 10\% without exceeding the maximum scale \citep{MOET2025Circular06}.

    \item \textbf{Asymmetric sanctions (Decree 04/2021/N{\DJ}-CP).} Exceeding authorized quotas triggers fines beginning at 3\% above quota (escalating at 10\%, 15\%, and 20\%) plus mandatory next-year quota deductions \citep{Gov2021Decree04}.
\end{enumerate}

These rules create a sharp asymmetry: over-admission incurs immediate fines and seat cuts, whereas under-enrollment carries no monetary fine (a program's license expires on under-enrollment grounds only after five consecutive zero-intake years, \citealp[Art.~12(7); distinct from compliance suspensions under Art.~12(6)]{MOET2024Circular12}). Published quotas therefore act as binding ceilings when demand is abundant, but provide no floor when demand is weak.

\subsection{Separating the Published Quota from Realized Intake}
To formalize this boundary, let $i \in \mathcal{I}$ index individual undergraduate degree programs observed across admission years $t \in \mathcal{T}$ (Section~\ref{sec:data}). Each program $i$ belongs to host university $u(i) \in \{1, \dots, U\}$, maps to an occupational cluster $k(i)$ (nine market clusters plus a state-regulated sector), an institutional governance tier $\tau(i) \in \{\text{Flagship Public}, \text{Other Public}, \text{Private}, \text{International}\}$, and a geographic region $r(i) \in \{\text{North}, \text{Central}, \text{South}, \text{National/other}\}$.

\begin{definition}[Quota, Short-Side Intake, and Fill Ratio]
For degree program $i$ in admission year $t$, let $S_{i,t} \in \mathbb{Z}^+$ denote the published admission quota subject to Eq.~\eqref{eq:moet_cap}, let $D_{i,t}^* \in \mathbb{R}_+$ denote latent student demand as a function of annual tuition $T_{i,t}$ and the institution's predetermined historical selectivity tier $C_i$ (a pre-clearing proxy for institutional prestige and applicant-pool quality, distinct from the endogenous year-$t$ market-clearing cut-off score), and let $D_{i,t} \in \mathbb{Z}_+$ denote realized student intake. Realized intake obeys a short-side clearing rule \citep{FairJaffee1972,MaddalaNelson1974}:
\begin{equation}
    D_{i,t} = \min\!\left( D_{i,t}^*\!\left(T_{i,t}, C_i\right), \; \bar{\phi} \cdot S_{i,t} \right), \qquad \text{with fill ratio } \Phi_{i,t} \equiv \frac{D_{i,t}}{S_{i,t}},
    \label{eq:fulfillment_ratio}
\end{equation}
where $\bar{\phi} \ge 1$ represents the effective enforcement tolerance around the published quota (statutory fines begin at $\bar{\phi} = 1.03$).
\end{definition}

Taking logarithms of Eq.~\eqref{eq:fulfillment_ratio} yields a direct structural link to censored and two-part econometric models \citep{Tobin1958,Cragg1971}:
\begin{equation}
    Z_{i,t} \equiv \ln\!\left(\frac{D_{i,t}}{S_{i,t}}\right) = \min\!\left( \ln\!\frac{D_{i,t}^*}{S_{i,t}},\; \ln\bar{\phi} \right).
    \label{eq:log_fill_censored}
\end{equation}
Programs therefore partition into two economic regimes:
\begin{itemize}
    \item \textbf{Supply-constrained programs ($\Phi_{i,t} \ge 1$, $Z_{i,t} \ge 0$)} attract more qualified applicants than available seats and clear by raising cut-off scores. Because over-admission is fined, intake concentrates heavily at or near $Z_{i,t} = 0$ ($D_{i,t} = S_{i,t}$), though a small fraction of programs still overshoot by more than 30\%, showing that $\bar{\phi}$ behaves in practice as a soft regulatory ceiling rather than an impassable wall (Section~\ref{sec:data}).
    \item \textbf{Demand-constrained programs ($\Phi_{i,t} < 1$, $Z_{i,t} < 0$)} fail to fill their seats even at minimum entry thresholds, generating a skewed left tail whenever tuition and opportunity costs outweigh perceived degree returns.
\end{itemize}
Two identification boundaries follow from Eq.~\eqref{eq:log_fill_censored}. First, because universities choose $S_{i,t}$ endogenously using private demand expectations and capacity constraints, the Tobit and Cragg models estimated below operate strictly as \emph{reduced-form conditional predictors} of realized intake given published quotas rather than structural identifiers of unconstrained latent demand $D_{i,t}^*$. Second, beyond mass accumulation at the statutory ceiling ($\Phi_{i,t} = 1$), Circular~03/2022's 80\% fill requirement for next-year quota expansion can also incentivize administrative bunching or conservative quota adjustment around $\Phi_{i,t} = 0.80$, reinforcing why fill-ratio distributions exhibit non-Gaussian kinks that motivate flexible censored and quantile estimators.

\subsection{The Phantom Quota Problem}
Whereas classical matching theory highlights strategic capacity under-reporting in oversubscribed clearinghouses \citep{Sonmez1997,KojimaPathak2009}, Vietnam's tuition-financed landscape shows that a university's published quota $S_{i,t}$ is an administrative target chosen under asymmetric regulatory and commercial incentives rather than an unbiased statistical expectation of $D_{i,t}$. We define unfilled \textit{phantom capacity} as $\mathrm{Phantom}_{i,t} \equiv \max(0, S_{i,t} - D_{i,t})$, and classify a program as exhibiting a \textit{severe phantom quota} when $D_{i,t}/S_{i,t} < 0.50$ and $S_{i,t} \ge 100$ seats. In practice, quotas diverge from intake in two opposing directions:
\begin{enumerate}
    \item[(i)] \textbf{Quota inflation ($S_{i,t} \gg D_{i,t}$):} Less selective institutions, particularly private and tuition-dependent universities competing for residual applicants after flagship campuses clear, often publish ambitious quotas near their physical and staffing limits in Eq.~\eqref{eq:moet_cap}, risking large unfilled tails when enrollment falls short.
    \item[(ii)] \textbf{Conservative quotas ($S_{i,t} < D_{i,t}$):} High-demand faculties occasionally announce cautious targets to maintain a buffer against yield surges and avoid Decree 04/2021 fines \citep{Gov2021Decree04}.
\end{enumerate}
Consequently, any uniform scaling heuristic (such as $0.95\,S_{i,t}$) systematically penalizes the large mass of compliant programs while barely touching severe phantom-quota misses, whereas unanchored regression models predicting year-on-year growth $\ln(D_{i,t}/D_{i,t-1})$ waste statistical capacity reconstructing the regulatory anchor $S_{i,t}$.

\subsection{Algebraic Fragility of Early-Sample Ratio Extrapolation}
A related forecasting pitfall arises early in the admission season, when a subset of early-reporting programs $\mathcal{S} \subset \mathcal{I}$ has published year-$t$ quotas $S_{i,t}$ while remaining institutions have not. A natural naive ratio heuristic---frequently encountered in institutional enrollment tracking before full national prospectuses are released---computes the early subset's baseline intake share $\omega = (\sum_{i \in \mathcal{S}} D_{i, t-1}) / (\sum_{j \in \mathcal{I}} D_{j, t-1}) \in (0, 1)$ and announced quota growth $g_{\mathcal{S}} = (\sum_{i \in \mathcal{S}} S_{i, t}) / (\sum_{i \in \mathcal{S}} D_{i, t-1}) - 1$, then adds the early quotas to non-reporting programs' prior intake scaled by $1 + g_{\mathcal{S}}$:
\begin{equation}
    \widehat{D}_{\text{total}, t}^{\text{heur}} = \sum_{i \in \mathcal{S}} S_{i, t} + \sum_{j \notin \mathcal{S}} D_{j, t-1}\left(1 + g_{\mathcal{S}}\right).
    \label{eq:heur_def}
\end{equation}
Although Eq.~\eqref{eq:heur_def} appears to stabilize early projections using non-reporting baselines, a direct algebraic reduction shows that it collapses identically to an unstabilized inverse-share multiplier.

\begin{remark}[Sample-Ratio Multiplier Equivalence]
\label{rem:equivalence}
For any non-empty early-reporting subset $\mathcal{S} \subset \mathcal{I}$ with baseline intake $D_{\mathcal{S}, t-1} = \sum_{i \in \mathcal{S}} D_{i, t-1} > 0$, population intake $D_{\mathcal{I}, t-1} = \sum_{j \in \mathcal{I}} D_{j, t-1}$, and share $\omega = D_{\mathcal{S}, t-1}/D_{\mathcal{I}, t-1} \in (0, 1)$, substituting $1 + g_{\mathcal{S}} = \big(\sum_{i \in \mathcal{S}} S_{i, t}\big) / D_{\mathcal{S}, t-1}$ into Eq.~\eqref{eq:heur_def} yields
\begin{equation}
    \widehat{D}_{\text{total}, t}^{\text{heur}}
    = \left(\sum_{i \in \mathcal{S}} S_{i, t}\right)\!\left[1 + \frac{\sum_{j \notin \mathcal{S}} D_{j, t-1}}{D_{\mathcal{S}, t-1}}\right]
    = \left(\sum_{i \in \mathcal{S}} S_{i, t}\right)\frac{D_{\mathcal{I}, t-1}}{D_{\mathcal{S}, t-1}}
    \equiv \frac{1}{\omega} \sum_{i \in \mathcal{S}} S_{i, t}.
    \label{eq:prop_equiv}
\end{equation}
\end{remark}

Historical intake among non-reporting programs cancels completely except through the aggregate share $\omega$, so any phantom-quota inflation $\epsilon_{\mathcal{S},t} = \sum_{i \in \mathcal{S}}(S_{i,t} - D_{i,t}) > 0$ among early reporters is magnified by $1/\omega$---five- to ten-fold when $\omega \in [0.10, 0.20]$---and compounded by early-reporting selection bias. Reliable national forecasting therefore requires modeling program-level fill behavior directly rather than extrapolating early quota announcements.

\section{Data}
\label{sec:data}

\paragraph{Sources, timing, and variables.}
Our panel compiles mandatory public admission prospectuses for 2022--2026, harmonized by  RMIT University Vietnam's market intelligence team{} (Table~\ref{tab:panel_overview}). Each record reports institution code, 7-digit MOET code, program title and track, published quota ($S_{i,t}$), realized intake ($D_{i,t}$), and annual tuition ($T_{i,t}$). Direct tuition is observed for 99.2\% of 2023 and 99.9\% of 2024 rows, dropping to 12.2\% in 2025 and 13.0\% in 2026; missing fees carry forward 2024 tuition (or campus median). Because program-level cut-off scores (\textit{{\dj}i{\^{e}}m chu{\^{a}}n}) were not uniformly digitized, selectivity tier ($C_i \in \{16.5, 21.0, 23.0, 25.5\}$) is a deterministic four-level institutional proxy (Appendix~\ref{app:upstream}; zero outcome leakage; Section~\ref{sec:benchmarks} verifies insensitivity to reclassifying three keyword-matched private campuses). External labor demand combines Lightcast and Vietnam National University full-year vacancy totals (2022--2025) across nine occupational clusters plus a state-regulated sector, always lagged $t-1$ to prevent look-ahead; two constant columns (wage and regional cohort growth) are excluded.

\begin{table}[t]
\centering
\scriptsize
\setlength{\tabcolsep}{3pt}
\caption{Descriptive overview of the audited 2022--2026 Vietnamese undergraduate panel (26,895 program-year records across 309 universities).}
\label{tab:panel_overview}
\begin{tabularx}{\textwidth}{c r r r r r c c c c r}
\toprule
\textbf{Year} & \textbf{Rows} & \textbf{Unis} & \textbf{Quotas} & \textbf{Intake $>0$} & \textbf{Seats (sum)} & \textbf{Wt.\ fill} & \textbf{Direct $T_{i,t}$} & \textbf{Exact} & \textbf{Near-fill} & \textbf{100\%-exact unis} \\
\midrule
2022 & 5,022 & 300 & 0 & 5,022 & --- & --- & --- & --- & --- & --- \\
2023 & 5,523 & 301 & 5,432 & 5,468 & 669,006 & 85.3\% & 99.2\% & 39.2\% & 42.3\% & 10 \\
2024 & 5,608 & 300 & 5,512 & 5,600 & 698,345 & 90.9\% & 99.9\% & 62.5\% & 64.7\% & 68 \\
2025 & 4,987 & 249 & 4,932 & 3,583 & 618,794 & 94.8\% & 12.2\% & 11.7\% & 31.7\% & 3 \\
2026 & 5,755 & 257 & 5,717 & 0 & 690,586 & --- & 13.0\% & --- & --- & --- \\
\bottomrule
\multicolumn{11}{@{}p{\textwidth}@{}}{\vspace{1pt}\scriptsize \textit{Notes:} \textit{Wt.\ fill} is seat-weighted aggregate fill $\sum D_{i,t}/\sum S_{i,t}$ across all rows with both quota and intake observed (2023: $N=5,379$; 2024: $N=5,504$; 2025: $N=3,569$). \textit{Direct $T_{i,t}$} is the share of rows with annual tuition reported directly in that year's prospectus (missing 2025/2026 values fall back to 2024 prospectus tuition). \textit{Exact} ($D=S$) and \textit{Near-fill} ($|D-S|\le 1$ or $0.98\le D/S\le 1.04$) are evaluated over all quota-and-intake-observed panel rows (e.g., 39.2\% in full 2023 panel vs.\ 38.7\% among the 5,385 continuing 2024 census links). \textit{100\%-exact unis} counts universities ($n\ge 3$ programs) reporting $D=S$ for every listed program.}
\end{tabularx}
\end{table}

\paragraph{Audited 2022--2026 panel and linkage diagnostics.}
The audited panel spans 26,895 rows across 309 universities, resolving quota bundling and alignment errors that distorted an earlier uncleaned extraction ($N = 4,434$, quota MAE 52.95, RMSE 250.8, $R^2 = -2.262$). Continuing tracks link across adjacent years via optimal bipartite matching on institution, 7-digit MOET code, track category, title similarity, and scale proximity (Appendix~\ref{app:upstream}).

Our primary 2024 census comprises 5,385 continuing programs across 298 universities with positive 2023--2024 intake and positive 2024 quota (5,286 also observe 2022 intake; seat-weighted fill 90.9\%; 2,172 North, 2,137 South, 1,073 Central, 3 national). Links are 70.6\% exact 1-to-1 ($N = 3,801$), 24.9\% multi-track bipartite, 2.3\% title fallback, and 2.2\% quota-share disaggregations. Published 2023 quotas appear for 98.4\% of rows (1.6\% initialized from 2023 intake for lags); 100\% observe $S_{i,2024}$, and $D_{i,t}$ is never imputed.

\begin{figure}[t]
    \centering
    \includegraphics[width=\textwidth]{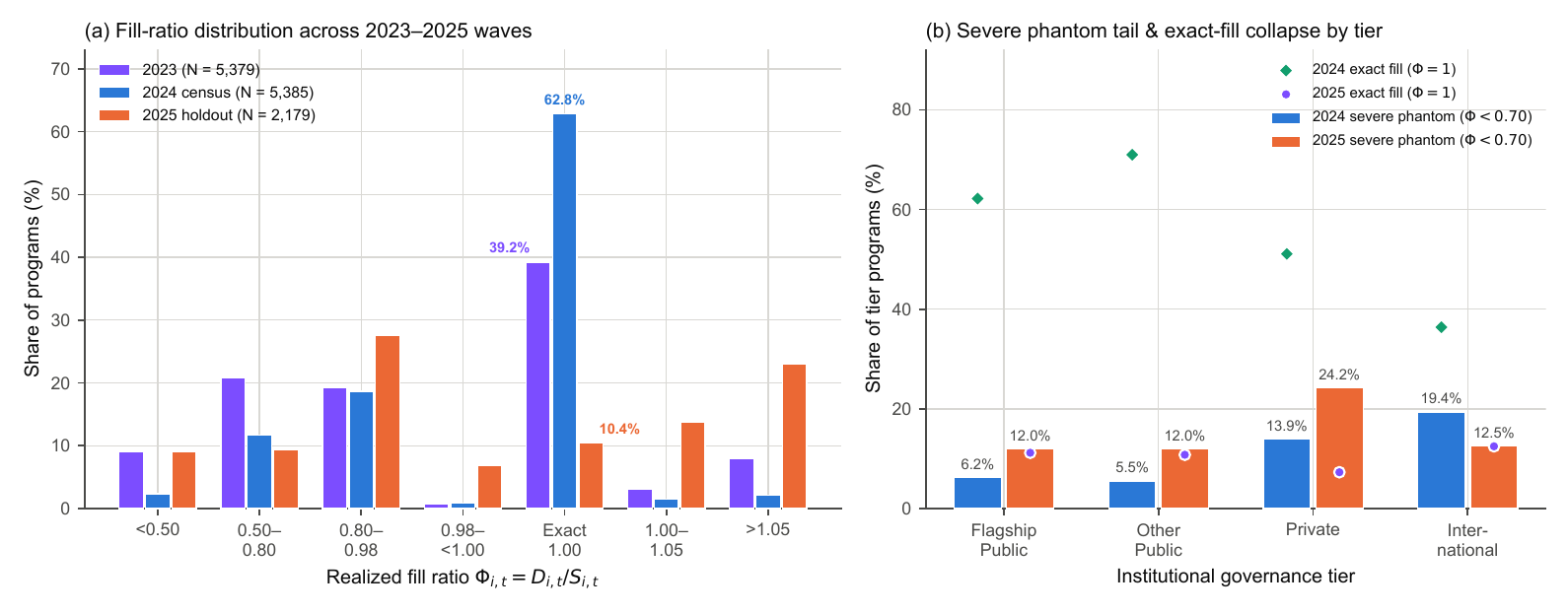}
    \caption{Empirical program fill-ratio distributions ($\Phi_{i,t} = D_{i,t}/S_{i,t}$) across 2023--2025: (a) full-wave fill-ratio bin shares illustrating the 2024 exact-fill spike (62.8\%) and 2025 post-reform collapse (10.4\%); (b) severe phantom-quota incidence ($\Phi_{i,t} < 0.70$, bars) and exact-fill shares (diamonds and circles) across institutional governance tiers}
    \label{fig:fill_dist}
\end{figure}

\paragraph{Exact-fill dynamics and reporting forensics.}
Table~\ref{tab:tier_rates} and Figure~\ref{fig:fill_dist} disaggregate compliance and phantom-quota tails, distinguishing \textit{strict exact fill} ($D_{i,t} = S_{i,t}$) from \textit{near-exact fill} ($|D_{i,t} - S_{i,t}| \le 1$ or $0.98 \le \Phi_{i,t} \le 1.04$):
\begin{itemize}
    \item Across quota-and-intake-observed panel rows (Table~\ref{tab:panel_overview}), strict exact fill rose from 39.2\% in 2023 ($N = 5,379$; 38.7\% in continuing 2024 links) to 62.5\% in 2024 ($N = 5,504$; 62.8\% continuing census, 65.0\% near-fill), then collapsed to 11.7\% in 2025 ($N = 3,569$; 31.7\% near-fill). Year-to-year persistence in 2024 is modest (65.8\% given 2023 exact fill vs.\ 60.9\% otherwise).
    \item The $2023 \to 2024$ surge occurred under unchanged early-admission rules as universities ($n \ge 3$) reporting 100\% exact fills across every listing jumped from 10 to 68 in the full 2024 panel---71 in the continuing census (1,277 rows, 23.7\%)---before plunging to 3 in 2025. Hence the 2025 drop reflects \emph{both} admission-round unification and a sharp decline in uniform quota-copying. Excluding all 71 uniform-exact campuses leaves \textbf{51.2\%} strict exact fill across 4,108 mixed-reporting programs (227 unis; Appendix Table~\ref{tab:app_attrition}, Panel~B): 12.6\% come from 16 near-uniform (90\%--99\%) and 35 small ($n < 3$) campuses, alongside broad exact fills across the remaining 176 mixed campuses (Other Public 58.6\%, Flagship 51.7\%, Private 41.4\%, International 34.0\%).
    \item Severe tails persist: 0.5\% of 2024 census programs over-enrolled beyond $1.30\,S_{i,t}$, 8.7\% contracted below $0.35\,D_{i,t-1}$, and 96 programs (1.8\%) exhibit severe phantom quotas ($\Phi_{i,2024} < 0.50, S_{i,2024} \ge 100$), where quota MAE spikes to 139.9 versus 10.0 on the remaining 98.2\% (twice as common at Private as at Other Public campuses: 2.8\% vs.\ 1.4\%; Table~\ref{tab:tier_rates}).
\end{itemize}

\begin{table}[t]
\centering
\scriptsize
\setlength{\tabcolsep}{3pt}
\caption{Empirical quota compliance and severe phantom-quota prevalence by institutional governance tier in the 2024 census ($N = 5,385$ continuing programs, 298 universities).}
\label{tab:tier_rates}
\begin{tabularx}{\textwidth}{L r r c c c c c c}
\toprule
\textbf{Institutional tier} & \textbf{$N$} & \textbf{Unis} & \textbf{Mean $\Phi_{2023}$} & \textbf{Mean $\Phi_{2024}$} & \textbf{Med.\ $\Phi_{2024}$} & \textbf{Strict exact} & \textbf{Near-fill} & \textbf{Severe phantom} \\
\midrule
Flagship Public & 1,483 & 77 & 0.892 & 0.949 & 1.00 & 62.2\% & 66.1\% & 1.6\% \\
Other Public & 2,452 & 155 & 0.875 & 0.948 & 1.00 & 71.0\% & 72.7\% & 1.4\% \\
Private & 1,285 & 58 & 0.820 & 0.886 & 1.00 & 51.1\% & 52.5\% & 2.8\% \\
International & 165 & 11 & 0.870 & 0.847 & 0.90 & 36.4\% & 37.0\% & 1.8\% \\
\bottomrule
\multicolumn{9}{@{}p{\textwidth}@{}}{\vspace{1pt}\scriptsize \textit{Notes:} Tiers are assigned purely by institutional charter rules (no hand-set inflation rates). $\Phi_t = D_{i,t}/S_{i,t}$ is clipped to $[0.25, 1.50]$ for program-unweighted means (seat-weighted census fill is 90.9\%). Strict exact: $D_{i,2024}=S_{i,2024}$; Near-fill: $|D_{i,2024}-S_{i,2024}|\le 1$ or $0.98 \le \Phi_{i,2024} \le 1.04$; Severe phantom: $\Phi_{i,2024} < 0.50$ with $S_{i,2024} \ge 100$. Three universities host both regular and international-track listings across tiers.}
\end{tabularx}
\end{table}

\paragraph{Out-of-time 2025 holdout, grain-consistency screening, and attrition audit.}
The 2025 wave contains 4,987 rows across 249 universities, with 3,583 reporting realized intake (415,652 students; 94.8\% seat-weighted fill across 3,569 quota-observed rows). Linking positive 2025 intake and quotas to 2024 baselines yields 3,348 matched programs across 158 universities (57.5\% exact 1-to-1, 22.6\% bipartite, 12.3\% title, 7.6\% quota-share; seat-weighted fill 95.2\%; strict exact 11.3\%, near-fill 31.8\%), which serves as our \emph{primary unconditioned ex-ante benchmark}. At the spring nowcast horizon, we evaluate both all 3,348 programs and a \textit{grain-consistency subset} ($N = 2,179$, 146 unis; strict exact 10.4\%, near-fill 31.0\%) satisfying $0.50 \le S_{i,2025}/S_{i,2024} \le 2.00$. Among the 1,169 excluded rows (34.9\%), 58.5\% (684) are exact 1-to-1 MOET-code links split 45.0\% $>2\times$ expansions and 54.9\% $<0.5\times$ contractions, confirming genuine large quota restructurings rather than linkage failure (on these 1,169 rows alone, H-Soft still edges $S_{2025}$: MAE 23.20 vs.\ 23.40).

Of the 298 universities in the 2024 census, 157 appear in all-matched 2025 disclosures and 141 (47.3\%) had not disclosed by our cutoff (one non-census 2024 campus brings 2025 to 158 and pooled 2023--2025 to 299). Appendix Table~\ref{tab:app_attrition} (Panel~A) shows that retained campuses skew public (80.4\% public, 18.4\% private, 1.2\% international; mean $\Phi_{24} = 0.939$) relative to dropped campuses (60.8\% public, 33.0\% private, 6.2\% international; mean $\Phi_{24} = 0.917$). This composition shift alongside post-reform overshoots resolves Table~\ref{tab:panel_overview}'s fill-rate paradox (aggregate fill rising to 94.8\% even as exact filling falls to 11.7\%): the 157 retained campuses already filled at 91.7\% in 2024 (vs.\ 89.6\% among dropped campuses), and within that balanced cohort fill climbed from 91.7\% to 95.3\% (95.2\% grain-consistent) as $>1.05$ overshoots surged tenfold---from 2.1\% in 2024 to 21.0\% (23.1\% grain-consistent; Figure~\ref{fig:fill_dist}(a)) in 2025 under unified clearing and broad-field seat reallocation (Eq.~\eqref{eq:moet_cap}). Section~\ref{sec:policy_shocks} therefore supplements unweighted holdout metrics with IPW estimates reweighted to 2024 campus composition (Appendix Table~\ref{tab:app_attrition}, Panel~C), noting that IPW adjusts for MAR observables but cannot rule out MNAR non-disclosure.

\paragraph{Information timing.}
\textit{Ex-ante forecasts} (autumn of year $t-1$) use no year-$t$ quota information; \textit{nowcasts} (spring of year $t$) condition on published year-$t$ quotas $S_{i,t}$. Both horizons strictly use lagged $t-1$ full-calendar-year vacancy-share growth ($g_{\text{job},k,t-1}$). Under Circular~08/2022 \citep[Art.~11]{MOET2022Circular08}, universities must publish admission prospectuses at least 30 days before preference registration opens; in 2025, registration ran 16--28 July and online enrollment confirmation closed 30 August \citep{MOET2025Decision1363}, making quotas observable at least two months before matriculation.

\section{Methodology}
\label{sec:methodology}

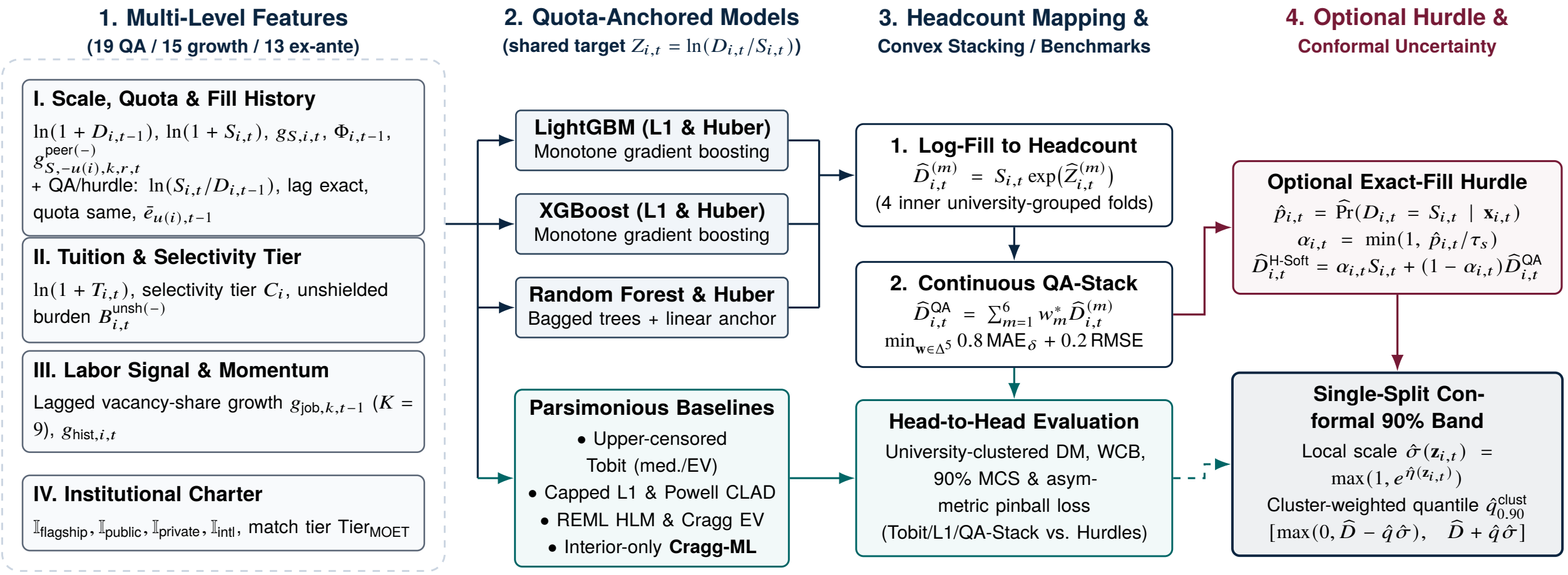
\begin{figure}[t]
\centering
\resizebox{\textwidth}{!}{%
\begin{tikzpicture}[
    font=\sffamily\footnotesize,
    >=Latex,
    coltitle/.style={
        font=\sffamily\bfseries\small, color=navyblue, align=center
    },
    box/.style={
        rectangle, rounded corners=3pt, draw=navyblue, line width=0.75pt,
        fill=white, align=center, inner sep=5.5pt
    },
    pillar/.style={
        rectangle, rounded corners=3pt, draw=slate!80, line width=0.65pt,
        fill=softgray, align=left, inner sep=4.5pt, text width=5.45cm
    },
    learner/.style={
        rectangle, rounded corners=2pt, draw=navyblue!85, line width=0.65pt,
        fill=accentbox, align=center, inner sep=3.6pt, text width=3.65cm
    },
    policy/.style={
        rectangle, rounded corners=3pt, draw=crimson!85!black, line width=0.8pt,
        fill=crimson!4, align=center, inner sep=5.2pt, text width=4.35cm
    },
    outbox/.style={
        rectangle, rounded corners=3pt, draw=navyblue, line width=1.0pt,
        fill=navyblue!7, align=center, inner sep=5.5pt, text width=4.35cm
    },
    arrow/.style={->, line width=0.8pt, color=navyblue},
    crimsonarrow/.style={->, line width=0.85pt, color=crimson!85!black}
]

% Column 1: Multi-Level Feature Space (19 quota-anchored / 15 growth / 13 ex-ante)
\node[coltitle] at (0, 4.75) {1. Multi-Level Features\\[-1pt]{\scriptsize (19 QA / 15 growth / 13 ex-ante)}};

\node[pillar] (p1) at (0, 2.95) {\textbf{I. Scale, Quota \& Fill History}\\[2pt]
{\scriptsize $\ln(1+D_{i,t-1})$, $\ln(1+S_{i,t})$, $g_{S,i,t}$, $\Phi_{i,t-1}$, $g^{\text{peer}(-)}_{S,-u(i),k,r,t}$}\\
{\scriptsize + QA/hurdle: $\ln(S_{i,t}/D_{i,t-1})$, lag exact, quota same, $\bar{e}_{u(i),t-1}$}};

\node[pillar] (p2) at (0, 1.05) {\textbf{II. Tuition \& Selectivity Tier}\\[2pt]
{\scriptsize $\ln(1+T_{i,t})$, selectivity tier $C_i$, unshielded burden $B_{i,t}^{\text{unsh}(-)}$}};

\node[pillar] (p3) at (0, -0.55) {\textbf{III. Labor Signal \& Momentum}\\[2pt]
{\scriptsize Lagged vacancy-share growth $g_{\text{job},k,t-1}$ ($K=9$), $g_{\text{hist},i,t}$}};

\node[pillar] (p4) at (0, -2.15) {\textbf{IV. Institutional Charter}\\[2pt]
{\scriptsize $\mathbb{I}_{\text{flagship}}, \mathbb{I}_{\text{public}}, \mathbb{I}_{\text{private}}, \mathbb{I}_{\text{intl}}$, match tier $\text{Tier}_{\text{MOET}}$}};

\node[draw=bordergray, rounded corners=7pt, dashed, line width=0.75pt, fit=(p1) (p2) (p3) (p4), inner sep=8pt] (featbox) {};

% Column 2: Quota-Anchored Base Learners & Parsimonious Econometric Track on Z_{i,t}
\node[coltitle] at (6.15, 4.75) {2. Quota-Anchored Models\\[-1pt]{\scriptsize (shared target $Z_{i,t} = \ln(D_{i,t}/S_{i,t})$)}};

\node[learner] (m1) at (6.15, 3.20) {\textbf{LightGBM (L1 \& Huber)}\\[-1.5pt]{\scriptsize Monotone gradient boosting}};
\node[learner] (m2) at (6.15, 2.00) {\textbf{XGBoost (L1 \& Huber)}\\[-1.5pt]{\scriptsize Monotone gradient boosting}};
\node[learner] (m3) at (6.15, 0.80) {\textbf{Random Forest \& Huber}\\[-1.5pt]{\scriptsize Bagged trees + linear anchor}};
\node[box, draw=teal!80!black, fill=teal!5, text width=3.65cm, inner sep=4pt] (econ) at (6.15, -1.65) {
    \textbf{Parsimonious Baselines}\\[1.5pt]
    {\scriptsize $\bullet$ Upper-censored Tobit (med./EV)}\\
    {\scriptsize $\bullet$ Capped L1 \& Powell CLAD}\\
    {\scriptsize $\bullet$ REML HLM \& Cragg EV}\\
    {\scriptsize $\bullet$ Interior-only \textbf{Cragg-ML}}
};

% Bus from Feature Space to ML & Econometric tracks
\coordinate (busL) at (3.65, 0);
\draw[line width=0.8pt, color=navyblue] (featbox.east |- m2.west) -- (busL |- m2.west);
\draw[line width=0.8pt, color=navyblue] (econ.west -| busL) -- (m1.west -| busL);
\foreach \m in {m1,m2,m3} {
    \draw[arrow] (\m.west -| busL) -- (\m.west);
}
\draw[arrow, color=teal!80!black] (econ.west -| busL) -- (econ.west);

% Column 3: Headcount Transformation FIRST, THEN Convex Simplex Stacking
\node[coltitle] at (11.35, 4.75) {3. Headcount Mapping \&\\[-1pt]{\scriptsize Convex Stacking / Benchmarks}};

\node[box, text width=4.15cm] (headconv) at (11.35, 2.70) {
    \textbf{1. Log-Fill to Headcount}\\[1pt]
    {\scriptsize $\widehat{D}_{i,t}^{(m)} = S_{i,t}\exp\!\big(\widehat{Z}_{i,t}^{(m)}\big)$}\\
    {\scriptsize (4 inner university-grouped folds)}
};

\node[box, text width=4.15cm] (qp) at (11.35, 0.70) {
    \textbf{2. Continuous QA-Stack}\\[1pt]
    {\scriptsize $\widehat{D}_{i,t}^{\text{QA}} = \sum_{m=1}^{6} w_m^* \widehat{D}_{i,t}^{(m)}$}\\
    {\scriptsize $\min_{\mathbf{w} \in \Delta^{5}} 0.8\,\text{MAE}_\delta + 0.2\,\text{RMSE}$}
};

\node[box, draw=teal!80!black, fill=teal!5, text width=4.15cm] (parsim) at (11.35, -1.65) {
    \textbf{Head-to-Head Evaluation}\\[1pt]
    {\scriptsize University-clustered DM, WCB,}\\
    {\scriptsize 90\% MCS \& asymmetric pinball loss}\\
    {\scriptsize (Tobit/L1/QA-Stack vs.\ Hurdles)}
};

% Collector bus merging ML learners into Step 1 (Headcount Transform), then Step 2 (Simplex Stack)
\coordinate (busR) at (8.55, 0);
\foreach \m in {m1,m2,m3} {
    \draw[line width=0.8pt, color=navyblue] (\m.east) -- (\m.east -| busR);
}
\draw[line width=0.8pt, color=navyblue] (m3.east -| busR) -- (m1.east -| busR);
\draw[arrow] (headconv.west -| busR) -- (headconv.west);
\draw[arrow] (headconv.south) -- (qp.north);
\draw[arrow, color=teal!80!black] (econ.east) -- (parsim.west);
\draw[arrow, color=teal!80!black] (qp.south) -- (parsim.north);

% Column 4: Optional Soft Hurdle Blend & Cluster-Weighted Conformal Intervals
\node[coltitle, color=crimson!85!black] at (16.85, 4.75) {4. Optional Hurdle \&\\[-1pt]{\scriptsize Conformal Uncertainty}};

\node[policy] (hurdle) at (16.85, 1.95) {
    \textbf{Optional Exact-Fill Hurdle}\\[1pt]
    {\scriptsize $\hat{p}_{i,t} = \widehat{\Pr}(D_{i,t}=S_{i,t} \mid \mathbf{x}_{i,t})$}\\
    {\scriptsize $\alpha_{i,t} = \min(1,\, \hat{p}_{i,t}/\tau_s)$}\\
    {\scriptsize $\widehat{D}_{i,t}^{\text{H-Soft}} = \alpha_{i,t} S_{i,t} + (1-\alpha_{i,t})\widehat{D}_{i,t}^{\text{QA}}$}
};

\node[outbox] (out) at (16.85, -1.45) {
    \textbf{Single-Split Conformal 90\% Band}\\[1pt]
    {\scriptsize Local scale $\hat{\sigma}(\mathbf{z}_{i,t}) = \max(1, e^{\hat{\eta}(\mathbf{z}_{i,t})})$}\\
    {\scriptsize Cluster-weighted quantile $\hat{q}_{0.90}^{\text{clust}}$}\\
    {\scriptsize $[\max(0, \widehat{D} - \hat{q}\hat{\sigma}),\;\; \widehat{D} + \hat{q}\hat{\sigma}]$}
};

\draw[crimsonarrow] (qp.east) -- ++(0.45,0) |- (hurdle.west);
\draw[crimsonarrow] (hurdle.south) -- (out.north);
\draw[arrow, dashed, color=teal!80!black] (parsim.east) -- ++(0.45,0) |- (out.west);

\end{tikzpicture}%
}
\caption{Parallel quota-anchored modeling architecture on the shared log fill-ratio target $Z_{i,t} = \ln(D_{i,t}/S_{i,t})$. Multi-level features across four pillars feed both (top track) six ML learners mapped to headcounts before convex simplex stacking (\textbf{QA-Stack}) with an optional LightGBM exact-fill hurdle (\textbf{H-Soft}) and (bottom track) parsimonious canonical econometric and censored baselines (\textbf{Tobit}, \textbf{Capped L1}, iterative \textbf{Powell CLAD}, \textbf{REML HLM}, \textbf{Cragg EV}, and interior-only \textbf{Cragg-ML}), evaluated head-to-head via clustered DM, wild cluster bootstrap, 90\% MCS, asymmetric pinball loss, and single-split cluster-weighted conformal intervals}
\label{fig:poise_arch}
\end{figure}

To isolate what genuinely improves program-level enrollment forecasts over the published quota, we compare four principled modeling families under an identical university-grouped validation protocol: (1) operational quota and simple scaling rules; (2) unanchored log-growth machine-learning ensembles; (3) canonical censored and two-part econometric baselines \citep{Tobin1958,Cragg1971}; and (4) quota-anchored log-fill stacking combined with a soft probability-weighted hurdle (\textbf{H-Soft}; Figure~\ref{fig:poise_arch}). Every equation below corresponds one-to-one with the accompanying replication code (\texttt{run\_analysis.py}).

\subsection{Target Parameterizations and Feature Architecture}
\label{sec:features}
A central methodological finding of this study is that how the forecasting target encodes the published quota matters far more than learner complexity. We evaluate two targets in parallel:
\begin{align}
    \text{Unanchored log-growth:} \quad & Y_{i,t} = \operatorname{clip}\!\left(\ln\!\frac{D_{i,t}}{D_{i,t-1}},\; -1.20,\; 1.00\right), \qquad \widehat{D}_{i,t}^{\text{G}} = D_{i,t-1}\exp\!\big(\widehat{Y}_{i,t}\big), \label{eq:target_growth} \\
    \text{Quota-anchored log-fill:} \quad & Z_{i,t} = \operatorname{clip}\!\left(\ln\!\frac{D_{i,t}}{S_{i,t}},\; -1.50,\; 0.50\right), \qquad \widehat{D}_{i,t}^{\text{QA}} = S_{i,t}\exp\!\big(\widehat{Z}_{i,t}\big). \label{eq:target_fill}
\end{align}
In Eq.~\eqref{eq:target_growth}, learners predict year-on-year enrollment growth and must implicitly reconstruct the identity $D_{i,t} = S_{i,t}$ from covariates. In Eq.~\eqref{eq:target_fill}, the published quota enters multiplicatively as the baseline anchor, so $Z_{i,t} = 0$ corresponds to exact quota fulfillment and negative values isolate institutional quota miscalibration or optimism conditional on endogenous $S_{i,t}$ (rather than exogenous demand alone). Both targets are winsorized to prevent extreme one-off restructurings from dominating estimation.

Explanatory variables are organized into four pillars (complete definitions appear in Appendix Table~\ref{tab:app_features}; no hand-set cluster priors are included):
\begin{enumerate}
    \item \textbf{Pillar I: Scale, quota dynamics, and peer competition.} Baseline scale uses $\ln(1 + D_{i,t-1})$ and lagged quota compliance $\Phi_{i,t-1} = \operatorname{clip}(D_{i,t-1}/S_{i,t-1},\; 0.25,\; 1.50)$. At the nowcast horizon, we add $\ln(1 + S_{i,t})$, clipped announced quota growth $g_{S,i,t} = \operatorname{clip}((S_{i,t} - S_{i,t-1})/\max(S_{i,t-1}, 1),\; -0.55,\; 0.85)$, and leave-one-institution-out peer quota growth within occupational cluster $k(i)$ and region $r(i)$:
    \begin{equation}
        g^{\text{peer}}_{S, -u(i), k, r, t} = \operatorname{clip}\!\left(\frac{\sum_{j \in \mathcal{C}_{k,r} \setminus \mathcal{I}_{u(i)}} (S_{j,t} - S_{j,t-1})}{\max\!\big(\sum_{j \in \mathcal{C}_{k,r} \setminus \mathcal{I}_{u(i)}} S_{j,t-1},\; 50\big)},\; -0.30,\; 0.50\right),
        \label{eq:peer_quota}
    \end{equation}
    where $\mathcal{C}_{k,r}$ denotes all programs in cluster $k$ and region $r$ and $\mathcal{I}_{u(i)}$ denotes all programs at university $u(i)$. Ex-ante models drop the three year-$t$ quota features and include $\ln(1 + S_{i,t-1})$ (13 features total; 15 features for growth nowcasts). For quota-anchored, econometric, and hurdle models (19 features), Pillar~I is augmented with four transparent quota-history indicators observable at prospectus release: clipped log quota jump $\ln(S_{i,t}/\max(D_{i,t-1},1)) \in [-1.50, 2.00]$, lagged exact-fill indicator $\mathbb{I}(D_{i,t-1}=S_{i,t-1})$, unchanged-quota indicator $\mathbb{I}(S_{i,t}=S_{i,t-1})$, and the host university's lagged exact-fill rate $\bar{e}_{u(i),t-1}$.

    \item \textbf{Pillar II: Tuition and institutional selectivity tier.} Direct price is measured by $\ln(1 + T_{i,t})$ (VND/year).\footnote{At the 2024--2025 exchange rate ($1\text{ USD} \approx 25,000\text{ VND}$), $50\text{M}$ VND $\approx$ USD 2,000, or roughly 47\% of Vietnam's GDP per capita ($\approx$ USD 4,300).} Selectivity tier $C_i \in \{16.5, 21.0, 23.0, 25.5\}$ is a predetermined institutional-tier lookup proxy (documented in Appendix~\ref{app:upstream}, distinct from program-level cut-off scores). To reflect stronger price elasticity at less selective campuses, we include an \textit{unshielded tuition burden} interaction:
    \begin{equation}
        B_{i,t}^{\text{unsh}} = \frac{T_{i,t}}{10^8} \cdot \max\!\left(0,\; \frac{24.5 - C_i}{6.0}\right),
        \label{eq:unshielded_burden}
    \end{equation}
    where annual tuition is scaled in hundred-million VND units ($10^8$), the $24.5$-point threshold sets unshielded burden to zero for selective flagship institutions ($C_i = 25.5$), and the denominator $6.0$ normalizes the gap across the active lower-to-upper examination score range on the 30-point national scale.

    \item \textbf{Pillar III: Enrollment momentum and lagged labor-demand signal.} Prior intake momentum is
    \begin{equation}
        g_{\text{hist}, i, t} = \begin{cases}
            \operatorname{clip}\!\left(\dfrac{D_{i,t-1} - D_{i,t-2}}{\max(D_{i,t-2}, 1)},\; -0.50,\; 0.75\right) & \text{if } D_{i,t-2} > 0, \\[6pt]
            0 & \text{if } D_{i,t-2} \text{ is unobserved.}
        \end{cases}
        \label{eq:ghist_clamp}
    \end{equation}
    In the 2024 census, 99 continuing programs lack 2022 intake ($5,385 - 5,286$) and receive $g_{\text{hist},i,t} = 0$. External labor demand uses combined Lightcast and Vietnam National University vacancy postings $V_{k,t}$ across $K = 9$ market occupational clusters. To prevent any temporal look-ahead across both autumn ex-ante and spring nowcast horizons (and ensure symmetric full-calendar-year windows across 2024--2026), every model uses strictly lagged $t-1$ Dirichlet-smoothed vacancy-share log growth ($\gamma = 20$, $\sum_{k=1}^{9} s_{k,t-1} = 1$):
    \begin{equation}
        s_{k,t-1} = \frac{V_{k,t-1} + 20}{\sum_{\ell=1}^{9} V_{\ell,t-1} + 180}, \qquad g_{\text{job}, k, t-1} = \operatorname{clip}\!\left(\ln\frac{s_{k,t-1}}{s_{k,t-2}},\; -0.35,\; 0.45\right).
        \label{eq:job_share}
    \end{equation}
    Programs link to clusters via specialty title keywords (Tier~1: 568 census programs) or 4-digit MOET field codes (Tier~2: 3,878 programs); 939 programs in state-regulated fields (Tier~3: medicine, law, teacher education, national defense) receive $g_{\text{job},k,t-1} = 0$.

    \item \textbf{Pillar IV: Institutional charter.} Binary charter indicators ($\mathbb{I}_{\text{flagship}}, \mathbb{I}_{\text{public}}, \mathbb{I}_{\text{private}}, \mathbb{I}_{\text{intl}}$) and match-granularity tier $\text{Tier}_{\text{MOET}} \in \{1, 2, 3\}$.
\end{enumerate}

\subsection{University-Grouped Convex Stacking and Baseline Model Families}
All out-of-sample evaluation, inner stacking-weight estimation, and hurdle threshold selection use five-fold cross-validation grouped by host university (\texttt{GroupKFold} across 298 universities) so sibling programs never straddle training and test splits \citep{Roberts2017}. Base-learner tree hyperparameters (Table~\ref{tab:stacking_weights}) are fixed \textit{a priori} at conservative regularization defaults without outer-fold grid search, ensuring that only low-dimensional meta-parameters ($\mathbf{w}^*$, $\hat{\kappa}$, $\tau$, $\tau_s$, $\theta$) are tuned inside inner folds.

\paragraph{Smooth convex simplex stacking.}
Within each training split, base learners generate headcount predictions $\widehat{D}_{i,t}^{(m)}$ on four inner university-grouped folds (applying Eqs.~\eqref{eq:target_growth}--\eqref{eq:target_fill} \textit{before} stacking; Figure~\ref{fig:poise_arch}). Following stacked generalization \citep{Wolpert1992,vanDerLaan2007}, simplex weights $\mathbf{w}^* \in \Delta^{M-1}$ minimize a $C^\infty$ convex pseudo-Huber (Charbonnier) composite loss combining 80\% smoothed absolute error ($\delta = 0.5$ students) and 20\% smoothed root mean squared error ($\varepsilon = 10^{-8}$):
\begin{equation}
    \mathbf{w}^* = \arg\min_{\mathbf{w} \ge 0,\; \sum_m w_m = 1} \; 0.8 \cdot \frac{1}{N}\sum_{i=1}^{N} \sqrt{e_i(\mathbf{w})^2 + \delta^2} \;+\; 0.2 \cdot \sqrt{\frac{1}{N}\sum_{i=1}^{N} e_i(\mathbf{w})^2 + \varepsilon},
    \label{eq:stacking_qp}
\end{equation}
where $e_i(\mathbf{w}) = D_{i,t} - \sum_{m=1}^{M} w_m \widehat{D}_{i,t}^{(m)}$. Because Eq.~\eqref{eq:stacking_qp} is continuously differentiable and globally convex on $\Delta^{M-1}$ (and strictly convex whenever the base learners' headcount prediction columns are linearly independent), exact analytical gradients guarantee global convergence without the subgradient discontinuities of unsmoothed L1 optimization.

\paragraph{Four nowcast families, controlled $2 \times 2$ factorial ablations, and ex-ante benchmarks.}
Table~\ref{tab:stacking_weights} summarizes the base learners and fold-averaged simplex weights across our four nowcast families:
\begin{enumerate}
    \item \textbf{Operational quota and simple rules:} Published quota $S_{i,t}$, uniform scaling $0.95\,S_{i,t}$, program-level lagged fill $\operatorname{clip}(\Phi_{i,t-1}, 0.10, 2.00)\,S_{i,t}$, and out-of-fold institutional tier-median fill $\text{med}_{\tau(i)}(\Phi_{t})\,S_{i,t}$.
    \item \textbf{Unanchored log-growth ML (\textbf{G0}--\textbf{G2}) and controlled $2 \times 2$ factorial cells (\textbf{G-19}, \textbf{QA-15}):} Five diverse learners trained on $Y_{i,t}$ with 15 features and Huber loss---monotone LightGBM \citep{Ke2017}, monotone XGBoost \citep{Chen2016}, Random Forest \citep{Breiman2001}, Extra-Trees \citep{Geurts2006}, and standardized Huber regression---with monotonicity constraints enforcing $\partial \widehat{Y}/\partial g_{S} \ge 0$, $\partial \widehat{Y}/\partial g^{\text{peer}}_{S} \le 0$, and $\partial \widehat{Y}/\partial B^{\text{unsh}} \le 0$. Their convex stack is \textbf{G0}. To test rule-based adjustments without hand-set inflation constants, \textbf{G1} blends \textbf{G0} with an out-of-fold empirically de-biased quota ($0.60\,\widehat{D}^{\text{G0}}_{i,t} + 0.40\,S_{i,t}\min(\Phi_{i,t-1}, \bar{\Phi}_{\tau(i)})$, where $\bar{\Phi}_{\tau}$ is the training-fold mean fill of charter tier $\tau$), and \textbf{G2} multiplies \textbf{G1} by a direct high-tuition decay $\exp(-\hat{\kappa}\max(0, \ln(T_{i,t} / 50\text{M})))$ for non-international programs with $T_{i,t} > 50\text{M}$ VND and $C_i < 24$, with $\hat{\kappa} \ge 0$ estimated inside each training fold by constrained least squares on $\ln(D/S)$ (full-census $\hat{\kappa} = 0.2114$ across $n = 308$ qualifying programs). Crucially, to separate target parameterization ($Y_{i,t}$ vs.\ $Z_{i,t}$) from feature/loss changes, Table~\ref{tab:census} includes two off-diagonal $2 \times 2$ controls---bundling feature count (15 vs.\ 19), learner/loss suite (5 Huber vs.\ 6 L1+Huber), and omitted monotonicity constraints---\textbf{G-19} (6-learner L1+Huber on $Y_{i,t}$ with all 19 QA features, under both default $[-1.20, 1.00]$ and widened non-binding $[-2.50, 2.00]$ clips) and \textbf{QA-15} (5-learner Huber on $Z_{i,t}$ with the 15 growth features), alongside a pure single-factor control (\textbf{QA-19-nomono}) holding all 19 features, all 6 L1+Huber learners, and zero monotonicity constraints identical to widened-clip \textbf{G-19}.
    \item \textbf{Canonical higher-education multilevel and censored econometric baselines on $Z_{i,t}$:} Reflecting the exact-fill mass point at $D_{i,t} = S_{i,t}$ ($Z_{i,t} = 0$), primary econometric baselines censor at $Z_{i,t} \le 0$ (Appendix Table~\ref{tab:app_robustness} checks $\ln(1.03)$); fractional-response logit \citep{PapkeWooldridge1996} is excluded because its $[0,1]$ support cannot accommodate legal $+3\%$ or observed $>1.00$ overshoots. First, we fit a \textbf{random-intercept hierarchical linear model (HLM)} via REML \citep{Raudenbush2002} on $\min(Z_{i,t}, 0)$ with university random intercepts (filtering exact charter-dummy collinearity via QR rank screening and optimizing via Powell--BFGS: unconditional null $\widehat{\text{ICC}} = 0.492$; conditional residual $\widehat{\text{ICC}} = 0.479$ in both 5-fold CV and full-2024 fits), predicting via population-averaged fixed effects ($\hat{u}_g = 0$) in 2024 CV and 2025 OOT (with Empirical Bayes BLUPs $\hat{u}_g$ checked in Appendix Table~\ref{tab:app_robustness}). Second, motivated by \citet{Powell1984}, we fit a convex \textbf{capped L1 median quantile regression} (\texttt{QuantileRegressor}) on pre-capped $\min(Z_{i,t}, 0)$ alongside an iterative \textbf{Powell (1984) CLAD} specification using \citet{Buchinsky1994}'s trimming algorithm (iteratively restricting estimation to observations with predicted latent index $< 0$; Appendix Table~\ref{tab:app_robustness}). Third, we estimate an \textbf{upper-censored Tobit} \citep{Tobin1958} via $L_2$-penalized MLE ($\lambda = 1$) on $Z_{i,t} = \min(Z_{i,t}^*, 0)$, evaluating both the censored median $S_{i,t}\exp(\min(\hat{\mu}_i, 0))$ and the \emph{exact} closed-form headcount expectation free of Jensen bias:
    \begin{equation}
        \mathbb{E}\!\left[S_{i,t} e^{\min(Z_{i,t}^*, 0)} \mid \mathbf{x}_i\right] = S_{i,t}\!\left[\Phi\!\left(\frac{\hat{\mu}_i}{\hat{\sigma}}\right) + \exp\!\left(\hat{\mu}_i + \frac{\hat{\sigma}^2}{2}\right)\Phi\!\left(-\frac{\hat{\mu}_i + \hat{\sigma}^2}{\hat{\sigma}}\right)\right].
        \label{eq:tobit_exact_ev}
    \end{equation}
    Fourth, we estimate a regularized lognormal \textbf{Cragg-style two-part hurdle} \citep{Cragg1971} ($L_2$-logistic $\hat{\pi}_i = \widehat{\Pr}(D_{i,t} \ge S_{i,t} \mid \mathbf{x}_i)$ plus Ridge on $D_{i,t} < S_{i,t}$) with within-fold \citet{Duan1983} nonparametric smearing $\hat{\psi}_{\text{smear}}$, yielding $\hat{\pi}_i S_{i,t} + (1-\hat{\pi}_i) S_{i,t} \min(1,\, \hat{\psi}_{\text{smear}} e^{\widehat{Z}^{\text{under}}_i})$ alongside a $\hat{\pi}_i > 0.5$ threshold rule.
    \item \textbf{Quota-anchored log-fill ML (\textbf{QA-Stack}) and hurdles (\textbf{Jump-QA}, \textbf{H-Hard}, \textbf{H-Soft}):} Six base learners trained directly on $Z_{i,t} = \ln(D_{i,t}/S_{i,t})$---LightGBM (L1), monotone LightGBM (Huber), monotone XGBoost (L1), monotone XGBoost (Huber), Random Forest, and standardized Huber regression---combined in headcount space via Eq.~\eqref{eq:stacking_qp} to form \textbf{QA-Stack} (Table~\ref{tab:stacking_weights}, Panel~B). Because the L1-loss LightGBM and XGBoost trees produce highly collinear log-fill forecasts, the convex objective in Eq.~\eqref{eq:stacking_qp} is flat along their face of the simplex, explaining their wide fold-to-fold weight standard deviations ($0.35 \pm 0.35$ and $0.30 \pm 0.29$ in Table~\ref{tab:stacking_weights}); likewise, Eq.~\eqref{eq:stacking_qp}'s 20\% RMSE term intentionally trades a modest MAE increment relative to pure L1 base learners for lower squared error. To test sensitivity to cross-year covariate shift in host-university exact-fill history, Appendix Table~\ref{tab:app_linkage_agg} also evaluates \textbf{QA-Stack (no $\bar{e}_{u,t-1}$)} and \textbf{H-Soft (no $\bar{e}_{u,t-1}$)}.
\end{enumerate}

At the ex-ante horizon (no year-$t$ quota), we benchmark the five-learner ex-ante stack (\textbf{E0}) against naive carry-forward $D_{i,t-1}$, $1.05\,D_{i,t-1}$, and a three-predictor \textbf{Macro OLS} benchmark---ordinary least squares regression of $Y_{i,t}$ on standardized $[\ln(1 + D_{i,t-1}),\, g_{\text{hist}, i, t},\, g_{\text{job}, k, t-1}]$ using strictly lagged $t-1$ vacancies.

\begin{table}[t]
\centering
\footnotesize
\setlength{\tabcolsep}{4pt}
\caption{Base learners and smooth-convex simplex stacking weights (mean $\pm$ SD across five university-grouped folds in the 2024 census).}
\label{tab:stacking_weights}
\begin{tabularx}{\textwidth}{>{\raggedright\arraybackslash}p{3.3cm} L c c}
\toprule
\textbf{Learner} & \textbf{Fixed regularization configuration} & \textbf{Nowcast weight} & \textbf{Ex-ante weight} \\
\midrule
\multicolumn{4}{@{}l}{\textit{Panel A: Log-growth stacks} (target $Y_{i,t} = \ln(D_{i,t}/D_{i,t-1})$; 15 nowcast / 13 ex-ante features)} \\
Monotone LightGBM & Huber loss ($\alpha=0.14$), 260 trees, 23 leaves, depth 5, $\eta=0.03$, min leaf 18 & $0.00 \pm 0.00$ & $0.34 \pm 0.16$ \\
Monotone XGBoost & Pseudo-Huber (slope 0.14), 260 rounds, depth 4, $\eta=0.03$, min child weight 15 & $0.66 \pm 0.04$ & $0.24 \pm 0.11$ \\
Random Forest & 180 trees, max depth 9, min leaf 10, 70\% features per split & $0.00 \pm 0.00$ & $0.30 \pm 0.28$ \\
Extra-Trees & 180 trees, max depth 10, min leaf 10, 75\% features per split & $0.00 \pm 0.00$ & $0.06 \pm 0.06$ \\
Huber regression & Standardized linear Huber ($\epsilon=1.35$, $\alpha=1.0$) & $0.34 \pm 0.04$ & $0.06 \pm 0.05$ \\
\midrule
\multicolumn{4}{@{}l}{\textit{Panel B: Quota-anchored log-fill stack (\textbf{QA-Stack})} (target $Z_{i,t} = \ln(D_{i,t}/S_{i,t})$; 19 features)} \\
QA-LightGBM (L1) & L1 loss, 260 trees, 15 leaves, depth 4, $\eta=0.03$, min leaf 25 & $0.35 \pm 0.35$ & --- \\
QA-LightGBM (Huber) & Monotone Huber ($\alpha=0.10$), 260 trees, 15 leaves, depth 4, $\eta=0.03$, min leaf 25 & $0.09 \pm 0.15$ & --- \\
QA-XGBoost (L1) & Monotone L1 loss, 260 rounds, depth 4, $\eta=0.03$, min child weight 20 & $0.30 \pm 0.29$ & --- \\
QA-XGBoost (Huber) & Monotone pseudo-Huber (slope 0.10), 260 rounds, depth 4, $\eta=0.03$, min child weight 20 & $0.22 \pm 0.10$ & --- \\
QA-Random Forest & 180 trees, max depth 9, min leaf 12, 70\% features per split & $0.00 \pm 0.01$ & --- \\
QA-Huber regression & Standardized linear Huber ($\epsilon=1.35$, $\alpha=1.0$) & $0.04 \pm 0.08$ & --- \\
\bottomrule
\multicolumn{4}{@{}p{\textwidth}@{}}{\vspace{1pt}\scriptsize \textit{Notes:} Base-learner tree hyperparameters are fixed \textit{a priori} at conservative regularization defaults (no outer-fold grid search). Simplex weights are estimated in headcount space by minimizing the $C^\infty$ convex pseudo-Huber (Charbonnier) composite loss (Eq.~\eqref{eq:stacking_qp}, $\delta=0.5$ students) over four inner university-grouped folds within each outer training split. Monotone constraints enforce $\partial/\partial g_S \ge 0$ (growth nowcast), $\partial/\partial g^{\text{peer}}_S \le 0$, and $\partial/\partial B^{\text{unsh}} \le 0$.}
\end{tabularx}
\end{table}

\subsection{Quota-Anchored Hurdle Forecasts}
\label{sec:hurdle}
Even when continuous learners predict $Z_{i,t}$ directly, nonzero tree residuals can slightly perturb compliant programs away from $S_{i,t}$. To test whether explicit boundary routing adds durable value over continuous \textbf{QA-Stack} and parsimonious censored baselines (Tobit median and Capped L1)---or merely exploits uniform exact-fill reporting---we compare four mechanisms that preserve $S_{i,t}$ on high-confidence programs while deploying log-fill predictions on at-risk targets:
\begin{itemize}
    \item \textbf{Explicit quota-jump screen (\textbf{Jump-QA}):} Keeps $S_{i,t}$ unless $\ln(S_{i,t}/\max(D_{i,t-1},1)) > \theta$, in which case it switches to $\widehat{D}^{\text{QA}}_{i,t}$, with $\theta \in [0.20, 1.30]$ tuned on inner university-grouped folds ($\theta = 0.20$ across folds).
    \item \textbf{Hard classifier hurdle (\textbf{H-Hard}):} Fits a LightGBM classifier predicting $\hat{p}_{i,t} = \widehat{\Pr}(D_{i,t} = S_{i,t} \mid \mathbf{x}_{i,t})$ on the 19-feature quota-anchored set, keeping $S_{i,t}$ if $\hat{p}_{i,t} > \tau$ and switching to $\widehat{D}^{\text{QA}}_{i,t}$ otherwise, with $\tau \in [0.05, 0.60]$ selected on inner folds ($\tau \in [0.20, 0.40]$, full-census $\tau = 0.30$).
    \item \textbf{Soft probability-weighted hurdle (\textbf{H-Soft}):} Replaces the discontinuous binary switch of H-Hard with a piecewise-linear probability blend:
    \begin{equation}
        \widehat{D}^{\text{H-Soft}}_{i,t} = \alpha_{i,t}\, S_{i,t} + (1 - \alpha_{i,t})\, \widehat{D}^{\text{QA}}_{i,t}, \qquad \alpha_{i,t} = \min\!\left(1,\; \frac{\hat{p}_{i,t}}{\tau_s}\right),
        \label{eq:hurdle}
    \end{equation}
    where $\tau_s \in [0.20, 1.00]$ is tuned on inner university-grouped folds to minimize MAE ($\tau_s \in \{0.60, 0.70\}$ across folds; full-census $\tau_s = 0.60$). Programs with $\hat{p}_{i,t} \ge \tau_s$ remain anchored identically at $S_{i,t}$, whereas uncertain programs shrink smoothly toward $\widehat{D}^{\text{QA}}_{i,t}$, eliminating abrupt switching jumps under regime shifts.
    \item \textbf{Interior-only ML hurdle (\textbf{Cragg-ML}):} Because Eq.~\eqref{eq:hurdle} blends $S_{i,t}$ with an \emph{unconditional} QA-Stack trained on all observations (including exact fills), H-Soft double-counts the exact-fill mass point---preserving sharp median/L1 accuracy on compliant programs at the cost of higher squared error and upward aggregate sum bias (Section~\ref{sec:pooled_2026}). To isolate this mechanism directly, we also benchmark a Cragg-consistent ML hurdle trained strictly on interior shortfalls ($D_{i,t} < S_{i,t}$):
    \begin{equation}
        \widehat{D}^{\text{Cragg-ML}}_{i,t} = \hat{\pi}_{i,t}^{\text{ML}}\, S_{i,t} + \big(1 - \hat{\pi}_{i,t}^{\text{ML}}\big)\,\min\!\left(S_{i,t},\; \widehat{D}^{\text{QA,int}}_{i,t}\right),
        \label{eq:cragg_ml}
    \end{equation}
    where $\hat{\pi}_{i,t}^{\text{ML}} = \widehat{\Pr}(D_{i,t} \ge S_{i,t} \mid \mathbf{x}_{i,t})$ is the regularized LightGBM ceiling classifier and $\widehat{D}^{\text{QA,int}}_{i,t}$ fits the six-learner convex stack strictly on uncensored programs $D_{i,t} < S_{i,t}$. Hard-capping $\widehat{D}^{\text{QA,int}}_{i,t}$ at $S_{i,t}$ without Duan smearing eliminates ceiling double-counting automatically---matching conditional-mean RMSE and raw 2026 aggregate totals---while introducing downside bias on programs that overshoot $S_{i,t}$ during reform years (such as 2025's $>1.05\,S_{i,t}$ surge).
\end{itemize}

\subsection{Single-Split Cluster-Weighted Conformal Prediction Intervals}
\label{sec:conformal}
To quantify uncertainty around point forecasts $\widehat{D}_{i,t}$, we implement locally weighted conformal prediction \citep{Vovk2005,Papadopoulos2011,Lei2018} extended to hierarchical university clusters \citep{Dunn2023}. Within each outer census training split (and across the full 2024 census when calibrating for 2025), universities are partitioned once into two disjoint equal halves: a scale-fitting set $\mathcal{D}_{\text{fit}}$ and a calibration set $\mathcal{D}_{\text{cal}}$ containing $G_{\text{cal}}$ universities (no nested double-splitting). Because the primary 2024 CV calibration uses inner-CV out-of-fold residuals on $\mathcal{D}_{\text{fit}} \cup \mathcal{D}_{\text{cal}}$ (which share training universities with the outer point learner), Section~\ref{sec:benchmarks} also verifies coverage under a strict three-way university split in which point models are fitted on a disjoint two-thirds subset of each training fold so that calibration universities never enter point estimation. On $\mathcal{D}_{\text{fit}}$, a standardized Ridge regression ($\alpha = 10$) predicts $\ln(|D_{i,t} - \widehat{D}_{i,t}| + 1)$ from five local volatility covariates $\mathbf{z}_{i,t} = [\ln(1+\widehat{D}_{i,t}),\, \ln(1+D_{i,t-1}),\, \ln(1+S_{i,t}),\, |\Phi_{i,t-1}-1|,\, \ln(1+\text{SD}_{\text{QA}, i, t})]$, where $\text{SD}_{\text{QA},i,t}$ is the standard deviation across the six quota-anchored learners, yielding the local error scale $\hat{\sigma}(\mathbf{z}_{i,t}) = \max(1.0,\; \exp(\hat{\eta}(\mathbf{z}_{i,t})))$.

On the disjoint calibration set $\mathcal{D}_{\text{cal}}$, we compute locally normalized residuals $r_i = |D_{i,t} - \widehat{D}_{i,t}| / \hat{\sigma}(\mathbf{z}_{i,t})$ and define all 90\% ($\alpha = 0.10$) calibration quantiles conservatively via the right-continuous empirical inverse CDF $\hat{q}_{0.90} = \inf\{q : \sum_{i \in \mathcal{D}_{\text{cal}}} w_i \mathbb{I}(r_i \le q) \ge p_{\text{target}}\}$ so exact-fill ties at $r_i = 0$ cannot cause under-coverage, comparing three schemes:
\begin{enumerate}
    \item \textbf{Standard program-level quantile ($\hat{q}^{\text{prog}}_{0.90}$):} Uniform weights $w_i = 1/|\mathcal{D}_{\text{cal}}|$ at $p_{\text{target}} = \min(1,\, \lceil (|\mathcal{D}_{\text{cal}}| + 1)(1-\alpha) \rceil / |\mathcal{D}_{\text{cal}}|)$, valid under program-level exchangeability \citep{Lei2018}.
    \item \textbf{University-cluster-weighted hierarchical quantile ($\hat{q}^{\text{clust}}_{0.90}$):} Following the CDF-pooling approach of \citet[][Theorems~4 and~7]{Dunn2023}, each calibration program receives inverse-cluster-size weight $w_i = 1 / (G_{\text{cal}}\, n_{u(i)})$ so every university contributes equally, evaluated at $p_{\text{target}} = \min(1,\, \lceil (G_{\text{cal}} + 1)(1-\alpha) \rceil / G_{\text{cal}})$. Under exchangeable university clusters, this inverse-size pooling provides asymptotic cluster-level marginal validity as $G_{\text{cal}} \to \infty$, with empirical behavior under exact-fill ties audited across size quintiles.
    \item \textbf{Constant-width quantile ($\hat{q}^{\text{const}}_{0.90}$):} Unscaled empirical quantile of $|D_{i,t} - \widehat{D}_{i,t}|$.
\end{enumerate}
For any test program, the resulting 90\% prediction interval is truncated below at zero:
\begin{equation}
    \widehat{\mathcal{C}}_{0.90}(\mathbf{x}_{i,t}) = \left[\max\!\left(0,\; \widehat{D}_{i,t} - \hat{q}_{0.90}\,\hat{\sigma}(\mathbf{z}_{i,t})\right),\;\; \widehat{D}_{i,t} + \hat{q}_{0.90}\,\hat{\sigma}(\mathbf{z}_{i,t})\right].
    \label{eq:conformal_interval}
\end{equation}
Symmetric MAE/RMSE point metrics and Eq.~\eqref{eq:conformal_interval}'s symmetric locally scaled bands provide neutral multi-stakeholder benchmarks (with upper endpoints allowed above $S_{i,t}$ to accommodate over-enrollment shocks); to reflect Decree~04/2021's asymmetric cost structure directly, Section~\ref{sec:policy_shocks} and Appendix Table~\ref{tab:app_linkage_agg} (Panel~D) also evaluate 2:1 and 3:1 asymmetric pinball-loss ratios penalizing unfilled budgeted seats versus over-admission risk, and local offices can extend the same framework to asymmetric conformalized quantile regression \citep{Romano2019}. In the 2024 census, each outer fold is evaluated strictly out of sample using the scale model and quantiles calibrated on its disjoint outer training universities. For the 2025 holdout, intervals are calibrated solely on 2024 census out-of-fold residuals, providing an unvarnished test of coverage under cross-year regulatory shift \citep{Tibshirani2019,Barber2023}. All statistical modeling decisions, econometric derivations, estimation scripts, and verification checks were designed, executed, and audited by the authors; general-purpose large language model tools were used only under direct author supervision for language copy-editing, LaTeX formatting, and code refactoring, with all empirical outputs independently verified.

\section{Results on the 2024 Census}
\label{sec:benchmarks}

\subsection{Forecasting Accuracy, Target Ablations, and Model Confidence Sets}
\begin{figure}[t]
    \centering
    \includegraphics[width=\textwidth]{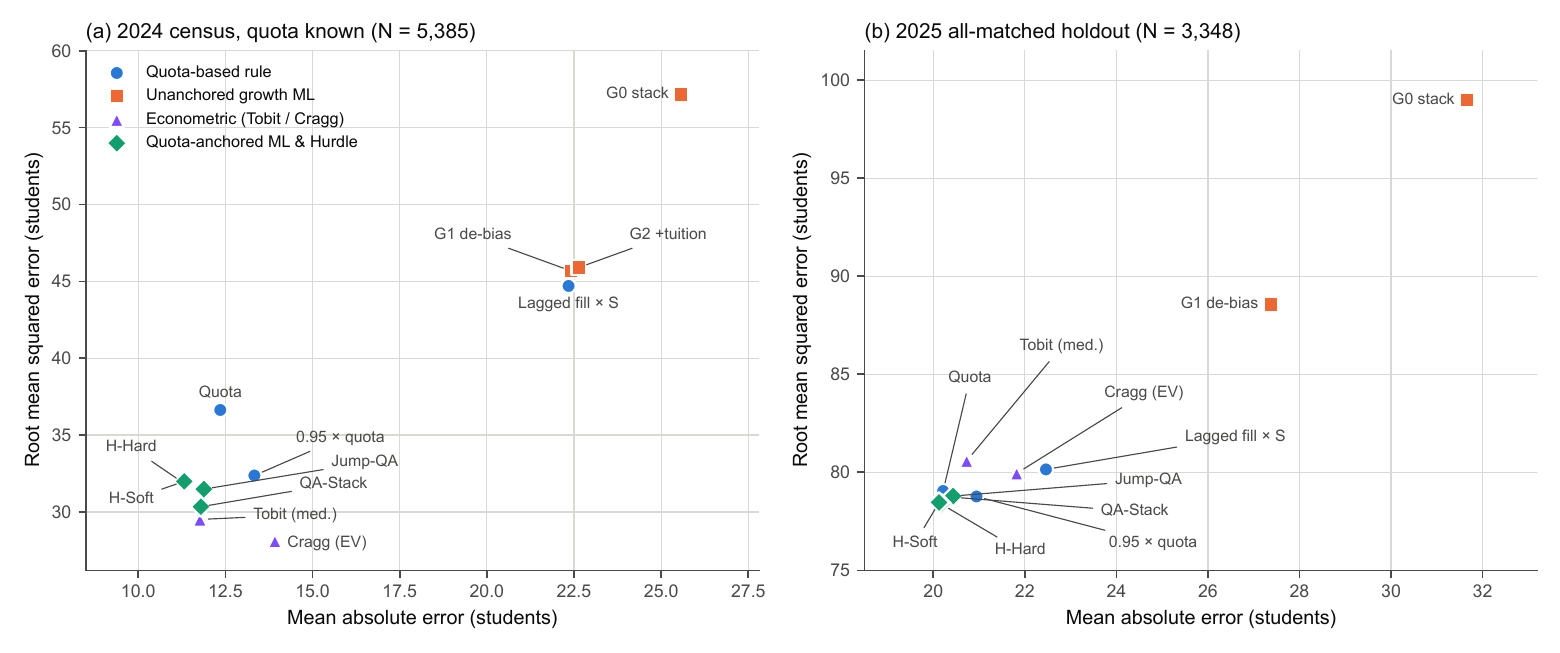}
    \caption{Mean absolute error against root mean squared error for nowcasts (published quota known): (a) 2024 census ($N = 5,385$), five-fold university-grouped cross-validation; (b) primary unscreened 2025 all-matched holdout ($N = 3,348$, quota RMSE 79.0; diagnostic grain-consistent subset RMSE 41.2 appears in Table~\ref{tab:holdout} Panel~C). Circles denote quota-based rules, squares denote unanchored growth ML, triangles denote econometric Tobit/Cragg baselines, and diamonds denote quota-anchored ML and hurdles; points lower and further left are better}
    \label{fig:accuracy}
\end{figure}

Table~\ref{tab:census} and Figure~\ref{fig:accuracy}(a) report five-fold university-grouped CV accuracy across the 5,385 continuing 2024 programs (298 universities). Every specification is tested against its reference via two-sided university-clustered Diebold--Mariano tests \citep{Diebold1995,Cameron2008} ($df = 297$), complemented by \citet{NadeauBengio2003} fold-overlap $t$-tests ($df = 4$), Benjamini--Hochberg FDR \citep{Benjamini1995}, and university-cluster-bootstrapped \citet{Hansen2011} 90\% Model Confidence Sets ($\widehat{\mathcal{M}}_{90\%}^*$, marked inline by $^\dagger$ for MAE and $^\ddagger$ for RMSE). The 16-model MCS candidate set is defined by explicit inclusion criteria as all primary quota-anchored specifications and simple quota scaling rules in Table~\ref{tab:census}, excluding strictly dominated log-growth models, diagnostic ablations, and Cragg hard-threshold.

\begin{table}[!ht]
\centering
\scriptsize
\setlength{\tabcolsep}{2.8pt}
\caption{Out-of-sample accuracy on the 2024 census (5,385 continuing programs, 298 universities; five-fold cross-validation grouped by university).}
\label{tab:census}
\begin{tabularx}{\textwidth}{L c r r r r r r}
\toprule
\textbf{Model} & \textbf{MAE [95\% CI]} & \textbf{RMSE} & \textbf{MedAE} & \textbf{$R^2$} & \textbf{DM $t$} & \textbf{$p_{\text{abs}}$} & \textbf{$p_{\text{sq}}$} \\
\midrule
\multicolumn{8}{@{}l}{\textit{Nowcast: published 2024 quota known} (reference for DM tests: published quota $S_t$)} \\
\multicolumn{8}{@{}l}{\quad \textit{1. Operational quota \& simple scaling rules}} \\
Published quota $S_t$$^{\dagger}$ & 12.35 $[9.9, 15.0]$ & 36.6 & 0.0 & 0.886 & --- & --- & --- \\
$0.95 \times S_t$$^{\ddagger}$ & 13.33 $[11.5, 15.4]$ & 32.4 & 5.0 & 0.911 & $-2.61$ & 0.009 & $<$0.001 \\
Lagged fill $\times S_t$ & 22.34 $[20.1, 24.7]$ & 44.7 & 10.0 & 0.831 & $-8.48$ & $<$0.001 & 0.002 \\
Tier-median fill $\times S_t$ (OOF)$^{\dagger}$ & 12.27 $[9.8, 14.9]$ & 36.5 & 0.0 & 0.887 & $+0.97$ & 0.333 & 0.195 \\
\addlinespace[2pt]
\multicolumn{8}{@{}l}{\quad \textit{2. Unanchored log-growth ML} ($Y_{i,t} = \ln(D_{i,t}/D_{i,t-1})$) \& \textit{2$\times$2 factorial controls}} \\
G0: Growth stack (15 feat, Huber) & 25.57 $[23.0, 28.2]$ & 57.2 & 9.0 & 0.723 & $-7.57$ & $<$0.001 & $<$0.001 \\
G1: G0 + empirical tier de-biasing & 22.42 $[20.2, 24.6]$ & 45.7 & 9.2 & 0.823 & $-6.61$ & $<$0.001 & 0.018 \\
G2: G1 + direct tuition decay ($T_{i,t} > 50$M) & 22.64 $[20.5, 24.8]$ & 45.9 & 9.3 & 0.822 & $-6.76$ & $<$0.001 & 0.015 \\
G-19: Growth stack (19 feat, default clip $[-1.20, 1.00]$) & 23.29 $[20.8, 26.0]$ & 55.5 & 5.9 & 0.739 & $-6.75$ & $<$0.001 & $<$0.001 \\
G-19 (widened non-binding clip $[-2.50, 2.00]$) & 14.86 $[13.0, 17.0]$ & 38.3 & 4.0 & 0.876 & $-2.82$ & 0.005 & 0.602 \\
QA-19-nomono (pure single-factor $Z_{i,t}$ control)$^{\text{a}}$ & 11.67 $[9.9, 13.6]$ & 30.3 & 2.7 & 0.922 & $+1.20$ & 0.231 & 0.001 \\
QA-15: Log-fill stack (15 feat, Huber)$^{\ddagger}$ & 12.15 $[10.5, 14.2]$ & 30.1 & 4.0 & 0.923 & $+0.39$ & 0.700 & $<$0.001 \\
\addlinespace[2pt]
\multicolumn{8}{@{}l}{\quad \textit{3. Canonical higher-education \& censored econometric baselines} ($Z_{i,t} = \ln(D_{i,t}/S_{i,t})$)} \\
HLM random-intercept (REML; cond.\ $\text{ICC}=0.479$)$^{\ddagger}$ & 14.74 $[13.3, 16.2]$ & 28.5 & 7.3 & 0.931 & $-2.65$ & 0.008 & 0.009 \\
Capped L1 median quantile ($\min(Z,0)$)$^{\dagger}$ & 12.14 $[9.7, 14.7]$ & 35.8 & 0.3 & 0.891 & $+2.09$ & 0.037 & $<$0.001 \\
Upper-censored Tobit (median)$^{\dagger\ddagger}$ & 11.77 $[9.9, 13.9]$ & 29.5 & 0.0 & 0.926 & $+0.99$ & 0.323 & 0.044 \\
Upper-censored Tobit (exact headcount mean)$^{\ddagger}$ & 14.39 $[13.0, 15.9]$ & 28.1 & 6.3 & 0.933 & $-2.22$ & 0.027 & 0.024 \\
Cragg-style two-part (EV, Duan smearing)$^{\ddagger}$ & 13.92 $[12.4, 15.5]$ & 28.1 & 6.1 & 0.933 & $-1.97$ & 0.050 & 0.005 \\
Cragg-style two-part (threshold) & 14.14 $[12.5, 16.0]$ & 31.9 & 0.0 & 0.914 & $-1.93$ & 0.054 & 0.151 \\
\addlinespace[2pt]
\multicolumn{8}{@{}l}{\quad \textit{4. Quota-anchored log-fill ML \& hurdles} ($Z_{i,t} = \ln(D_{i,t}/S_{i,t})$)} \\
QA-LightGBM (L1)$^{\dagger\ddagger}$ & 11.53 $[9.7, 13.5]$ & 30.9 & 1.9 & 0.919 & $+1.40$ & 0.163 & 0.005 \\
QA-XGBoost (L1)$^{\dagger\ddagger}$ & 11.58 $[9.7, 13.5]$ & 30.4 & 2.1 & 0.922 & $+1.40$ & 0.162 & 0.002 \\
QA-Stack (6-learner convex stack)$^{\dagger\ddagger}$ & 11.80 $[10.0, 13.7]$ & 30.3 & 2.9 & 0.922 & $+0.97$ & 0.332 & 0.001 \\
Jump-QA (explicit quota-jump screen) & 11.88 $[10.0, 14.0]$ & 31.5 & 1.9 & 0.916 & $+1.10$ & 0.273 & 0.001 \\
\textbf{H-Hard (hard classifier hurdle)}$^{\dagger}$ & 11.29 $[9.3, 13.3]$ & 31.9 & 0.0 & 0.914 & $+2.11$ & 0.036 & 0.010 \\
H-Soft (soft probability-weighted hurdle)$^{\dagger}$ & 11.32 $[9.2, 13.4]$ & 32.0 & 0.5 & 0.913 & $+2.31$ & 0.022 & 0.005 \\
Cragg-ML (interior-only $D<S$ hurdle)$^{\ddagger}$ & 12.79 $[11.3, 14.5]$ & 28.1 & 5.3 & 0.933 & $-0.61$ & 0.545 & 0.001 \\
\midrule
\multicolumn{8}{@{}l}{\textit{Ex-ante: no 2024 quota} (reference for DM tests: 3-predictor Macro OLS)} \\
Naive carry-forward $D_{t-1}$ & 65.12 $[60.8, 69.6]$ & 107.0 & 37.0 & 0.032 & $-3.47$ & $<$0.001 & 0.025 \\
$1.05 \times D_{t-1}$ & 66.49 $[62.1, 71.2]$ & 109.3 & 37.4 & -0.012 & $-4.27$ & $<$0.001 & 0.005 \\
Macro OLS (3 predictors) & 59.97 $[56.0, 63.5]$ & 99.7 & 34.6 & 0.158 & --- & --- & --- \\
Linear-13 (Ridge, same 13 features as E0) & 60.10 $[56.2, 63.7]$ & 99.9 & 34.4 & 0.156 & $-1.31$ & 0.191 & 0.281 \\
E0: Ex-ante stack (5-learner convex stack) & 59.11 $[55.5, 62.6]$ & 95.9 & 34.4 & 0.221 & $+1.20$ & 0.232 & 0.070 \\
\bottomrule
\multicolumn{8}{@{}p{\textwidth}@{}}{\vspace{1pt}\scriptsize \textit{Notes:} MAE, RMSE, and MedAE in students; 95\% CIs from university-cluster bootstrap ($B=1{,}000$); lagged $t-1$ vacancies throughout. $^{\text{a}}$Pure single-factor control holding all 19 features, 6 L1+Huber learners, and zero monotonicity identical to widened-clip G-19 (head-to-head vs.\ widened-clip G-19: $t=+4.99$, $p_{\text{abs}} < 0.001$, $p_{\text{sq}} = 0.005$; QA-Stack vs.\ widened-clip G-19: $t=+4.86$, $p_{\text{abs}} < 0.001$, $p_{\text{sq}} = 0.005$). DM $t$, $p_{\text{abs}}$, $p_{\text{sq}}$: two-sided university-clustered tests ($df = 297$; $t > 0$ beats reference). Inline markers denote 90\% Hansen--Lunde--Nason (2011) MCS ($B=1{,}000$) over the 16-model candidate set defined by explicit inclusion criteria (all primary quota-anchored specifications and scaling rules, excluding strictly dominated log-growth models, diagnostic ablations, and Cragg hard-threshold): $^{\dagger}$MAE MCS (H-Hard $p_{\text{MCS}}=1.000$, H-Soft $0.721$, $S_t$ $0.153$); $^{\ddagger}$RMSE MCS (Cragg EV $1.000$, Tobit EV $1.000$, Cragg-ML $1.000$; $S_t$ $0.003$, H-Soft $0.015$). Directional (one-sided superiority) BH-FDR across the 15 non-reference MCS candidates (and all 29 non-reference nowcast models, adding 5 growth base learners and 2 no-$\bar{e}_{u,t-1}$ rows): H-Soft clears 1\% squared-error FDR ($q_{\text{sup,sq}}=0.004$/$0.007$) and 10\% not 5\% candidate MAE FDR ($q_{\text{sup,abs}}=0.093$/$0.135$; Section~\ref{sec:benchmarks} reports two-sided sensitivity). Ex-ante head-to-head E0 vs.\ Linear-13 (Ridge): $t=+1.36$, $p_{\text{abs}} = 0.173$, $p_{\text{sq}} = 0.066$. Bold marks lowest MAE.}
\end{tabularx}
\end{table}

Five central findings emerge from Table~\ref{tab:census}:
\begin{enumerate}
    \item \textbf{Audited quotas anchor typical programs, while phantom-quota tails drive aggregate error.} Exact quota prediction yields MAE 12.35 and MedAE 0.0 (62.8\% exact fill), yet the worst 5\% generate half of quota error (41.8\% under QA-Stack; 46.5\% under H-Soft). On the 96 severe phantom-quota programs (1.8\%), quota MAE spikes to 139.9 versus 10.0 on the remaining 98.2\%. Simple scaling rules either degrade MAE or add negligible lift.
    \item \textbf{Target-parameterization ablations isolate log-fill gains beyond clipping.} Unanchored log-growth ensembles perform poorly under default clipping (G0: 25.57 MAE; G1--G2: 22.42--22.64). In the $2 \times 2$ controls, expanding to 19 features and L1+Huber (\textbf{G-19}) trims MAE only to 23.29, whereas switching to log fill ratio $Z_{i,t}$ with the 15-feature Huber suite (\textbf{QA-15}) halves error to 12.15, and \textbf{QA-Stack} reaches 11.80 (trading ~0.22--0.27 MAE against pure L1 trees because Eq.~\eqref{eq:stacking_qp} weights RMSE at 20\%). Widening $Y_{i,t}$'s clip to $[-2.50, 2.00]$ resolves 73.4\% of G-19's raw MAE gap relative to QA-Stack ($8.43/11.49$)---or 72.5\% relative to QA-19-nomono ($8.43/11.62$)---yet widened-clip G-19 (14.86 MAE) still loses to $S_t$ ($p_{\text{abs}} = 0.005$) and trails QA-Stack by over 20\% on MAE and RMSE ($p_{\text{sq}} = 0.005$). Most cleanly, Table~\ref{tab:census}'s pure single-factor control (\textbf{QA-19-nomono}: 11.67 MAE) holds all 19 features, 6 learners, and zero monotonicity fixed, cutting MAE by 21.5\% ($p < 0.001$) and RMSE by 20.9\% ($p = 0.005$) solely from switching $Y_{i,t} \to Z_{i,t}$; relaxing monotonicity lets QA-19-nomono slightly edge QA-Stack by fitting minor non-monotone tuition and peer interactions at a small trade-off against domain-sign regularization.
    \item \textbf{Canonical multilevel, censored, and count baselines perform strongly on $Z_{i,t}$.} Conditional-mean specifications deliver low tail error at higher MAE: full-rank REML HLM (null $\widehat{\text{ICC}} = 0.492$; conditional $\widehat{\text{ICC}} = 0.479$ across folds and full 2024) posts 28.5 RMSE ($p_{\text{sq}} = 0.009$), mirroring exact Tobit EV (28.1), Duan-smeared Cragg EV (28.1, $p_{\text{sq}} = 0.005$), and Poisson GLM (28.4; Appendix Table~\ref{tab:app_robustness}). Median-oriented baselines balance both metrics: Capped L1 reaches 12.14 MAE, iterative Powell (1984) CLAD sharpens this to 11.81 MAE and 30.4 RMSE ($p_{\text{sq}} = 0.013$; Appendix Table~\ref{tab:app_robustness}), and Tobit (median) attains 11.77 MAE and 29.5 RMSE.
    \item \textbf{Hurdle combination, multiplicity control, and parsimony.} In 2024 CV, the exact-fill classifier achieves AUC 0.685 with well-centered calibration (mean $\hat{p} = 63.2\%$ vs.\ 62.8\% base rate, Brier 0.214, slope 0.63). Combining it with QA-Stack yields \textbf{H-Hard} (11.29 MAE) and \textbf{H-Soft} (11.32 MAE, 32.0 RMSE; $p = 0.022/0.005$), cutting net error by 5,572 students across 2,454 adjusted programs, while trimming severe phantom-quota MAE only modestly (139.9 to 131.7) because prospectuses lack program-level cut-off scores (\textit{{\dj}i{\^{e}}m chu{\^{a}}n}) and private funnels. Under one-sided superiority BH-FDR across the 15 non-reference MCS candidates (and all 29 nowcast rows), H-Soft clears 1\% squared-error FDR ($q_{\text{sup,sq}} = 0.004/0.007$; two-sided $0.008/0.007$) and \citet{NadeauBengio2003} fold-overlap adjustment ($p = 0.047$, positive in all five folds). Full-census MAE clears 10\% not 5\% candidate FDR ($q_{\text{sup,abs}} = 0.093/0.135$; two-sided $0.093$, or $>0.16$ excluding inferior rejections) and is non-significant under overlap adjustment ($p = 0.134$). In Table~\ref{tab:census}'s 16-model MCS, \textbf{H-Hard} and \textbf{H-Soft} lead the 90\% MAE MCS yet exit the RMSE MCS due to ceiling blending, whereas interior-only \textbf{Cragg-ML} enters the RMSE MCS ($p_{\text{MCS}} = 1.000$) at 28.1 RMSE. Pairwise, H-Soft beats HLM, Capped L1, and QA-Stack on MAE ($p \le 0.033$) while tying H-Hard and canonical Tobit (median) ($p = 0.171/0.361$), which enters \emph{both} 90\% MCS sets alongside QA-LightGBM, QA-XGBoost, and QA-Stack.
    \item \textbf{Ex-ante horizon.} Every learner beats naive $D_{t-1}$ (65.12 MAE). Adding all 13 ex-ante covariates linearly (\textbf{Linear-13 Ridge}: 60.10) does not improve on three-predictor Macro OLS (59.97, $p = 0.191$), and \textbf{E0} (59.11) adds statistically insignificant gains in 2024 over both Macro OLS ($p = 0.232$) and Linear-13 Ridge ($p_{\text{abs}} = 0.173, p_{\text{sq}} = 0.066$).
\end{enumerate}

\subsection{Tier Heterogeneity, Structural Robustness, and 2024 Conformal Coverage}
Table~\ref{tab:tier_acc} (Panel~A) shows that 2024 gains concentrate in tuition-dependent tiers: across 58 Private campuses ($N = 1,285$), H-Soft cuts MAE from 21.61 to 17.88 and RMSE from 56.9 to 46.5 under both asymptotic ($p = 0.027/0.014$) and finite-sample Rademacher wild cluster bootstrap inference ($B = 2,000$: $p_{\text{WCB}} = 0.026/0.004$; Tobit RMSE 35.8). International programs show 27\%--29\% MAE reductions that remain imprecise across $G = 11$ clusters (asymptotic $p = 0.184/0.169$; Rademacher $p_{\text{WCB}} = 0.317/0.300$, exact $2^{11}$ enumeration $0.326/0.307$). Flagship Public campuses see RMSE relief ($p_{\text{sq}} = 0.034$), while in Other Public campuses (71.0\% exact fill) raw quotas edge H-Soft on MAE (7.93 vs.\ 8.08).

\begin{table}[t]
\centering
\scriptsize
\setlength{\tabcolsep}{1.5pt}
\caption{Nowcast accuracy stratified by institutional governance tier (MAE / RMSE in students) in the 2024 census ($N = 5,385$), the 2025 grain-consistent holdout ($N = 2,179$), and the full all-matched 2025 holdout ($N = 3,348$), with university-clustered Diebold--Mariano $p$-values for H-Soft vs.\ published quota $S_t$.}
\label{tab:tier_acc}
\begin{tabular*}{\textwidth}{@{\extracolsep{\fill}}l c c c c c c c c@{}}
\toprule
\textbf{Tier} & \textbf{$N$ (Unis)} & \textbf{Exact} & \textbf{Quota $S_t$} & \textbf{G0} & \textbf{Tobit} & \textbf{QA-Stack} & \textbf{H-Soft} & \textbf{H-Soft $p_{\text{abs}}/p_{\text{sq}}$} \\
\midrule
\multicolumn{9}{@{}l}{\textit{Panel A: 2024 census (5-fold university-grouped CV)}} \\
Flagship Public & 1,483 (77) & 62.2\% & 11.17 / 30.6 & 24.16 / 56.3 & 11.49 / 30.4 & 11.26 / \textbf{27.0} & \textbf{10.88} / 28.3 & 0.334 / 0.034 \\
Other Public & 2,452 (155) & 71.0\% & \textbf{7.93} / 24.9 & 26.21 / 60.2 & 8.41 / 25.4 & 8.84 / \textbf{23.6} & 8.08 / 24.4 & 0.162 / 0.143 \\
Private & 1,285 (58) & 51.1\% & 21.61 / 56.9 & 27.14 / 55.1 & 18.04 / \textbf{35.8} & 18.05 / 43.6 & \textbf{17.88} / 46.5 & 0.027 / 0.014 \\
International & 165 (11) & 36.4\% & 16.76 / 30.6 & 16.60 / 27.2 & 15.38 / 24.0 & \textbf{11.94} / \textbf{20.7} & 12.30 / 22.2 & 0.184 / 0.169 \\
\midrule
\multicolumn{9}{@{}l}{\textit{Panel B: 2025 grain-consistent holdout ($N = 2,179$; excludes quota restructurings)}} \\
Flagship Public & 740 (48) & 11.2\% & \textbf{13.36} / \textbf{24.8} & 16.11 / 27.2 & 13.83 / 25.6 & 14.32 / 25.8 & 13.79 / 25.2 & 0.026 / 0.057 \\
Other Public & 1,052 (72) & 10.8\% & \textbf{16.64} / \textbf{32.7} & 17.83 / 33.8 & 16.73 / 32.8 & 16.85 / 33.1 & 16.75 / 32.9 & 0.227 / 0.233 \\
Private & 355 (21) & 7.3\% & 35.38 / 76.8 & 35.26 / 74.1 & 35.36 / 76.3 & 34.45 / 71.3 & \textbf{33.93} / \textbf{70.6} & 0.323 / 0.269 \\
International & 32 (5) & 12.5\% & \textbf{11.62} / \textbf{25.7} & 15.51 / 33.2 & 16.13 / 37.6 & 14.34 / 31.5 & 13.72 / 30.6 & 0.408 / 0.433 \\
\midrule
\multicolumn{9}{@{}l}{\textit{Panel C: All matched 2025 programs ($N = 3,348$; includes all quota restructurings)}} \\
Flagship Public & 1,072 (50) & 11.6\% & \textbf{14.60} / \textbf{30.9} & 26.67 / 70.1 & 15.20 / 31.9 & 15.42 / 31.5 & 14.95 / 31.2 & 0.017 / 0.085 \\
Other Public & 1,651 (80) & 11.9\% & 18.04 / 38.5 & 29.09 / 59.8 & 18.08 / 38.5 & 18.01 / \textbf{38.1} & \textbf{17.98} / 38.3 & 0.574 / 0.418 \\
Private & 579 (22) & 9.0\% & 37.46 / 173.5 & 49.39 / 193.0 & 38.95 / 177.1 & 37.14 / 172.6 & \textbf{36.37} / \textbf{171.9} & 0.282 / 0.309 \\
International & 46 (6) & 10.9\% & \textbf{12.15} / \textbf{24.2} & 17.18 / 31.4 & 15.29 / 33.3 & 13.76 / 28.5 & 13.45 / 27.8 & 0.352 / 0.383 \\
\bottomrule
\multicolumn{9}{@{}p{\textwidth}@{}}{\vspace{1pt}\scriptsize \textit{Notes:} Each model cell reports MAE / RMSE in students; bold marks the single lowest MAE and lowest RMSE per row independently. Final column reports two-sided asymptotic university-clustered DM $p$-values comparing H-Soft against $S_t$ ($df = \text{Unis} - 1$). Finite-sample restricted wild cluster bootstrap (Rademacher weights, $B=2{,}000$) $p$-values ($p_{\text{WCB,abs}}/p_{\text{WCB,sq}}$) for small-cluster tiers ($G=5$--$22$): 2024 International ($G=11$): $0.317/0.300$ (exact: $0.326/0.307$); 2025 Panel~B Private ($G=21$): $0.346/0.357$, International ($G=5$): $0.508/0.508$ (Webb 6-point: $0.508/0.508$); 2025 Panel~C Private ($G=22$): $0.307/0.391$, International ($G=6$): $0.509/0.509$ (Webb 6-point: $0.469/0.485$). In 2024 Private universities ($G=58$), H-Soft beats $S_t$ under both asymptotic ($p = 0.027/0.014$) and wild-cluster-bootstrap inference ($p_{\text{WCB}} = 0.026/0.004$); in 2025 Private universities, directional gains remain statistically imprecise.}
\end{tabular*}
\end{table}

Four targeted robustness checks (full metrics in Appendix Tables~\ref{tab:app_attrition}, \ref{tab:app_robustness}, and~\ref{tab:app_linkage_agg}) confirm structural stability and clarify the hurdle mechanism:
\begin{itemize}
    \item \textbf{Mixed-reporting campuses ($N = 4,108$, 227 unis):} Excluding all 71 uniform-exact campuses raises quota error to 16.19/41.9 and strengthens every quota-anchored gain. Crucially, continuous \textbf{QA-Stack} (14.24/34.4, $p = 0.004/{<}0.001$) outperforms both \textbf{H-Soft} (14.45/36.5, $p = 0.002/0.003$) and \textbf{H-Hard} (14.49/36.4) on \emph{both} MAE and RMSE once uniform-exact reporters are removed (Tobit median: 14.61/33.1). This reversal shows candidly that the hurdle classifier's 2024 full-census MAE edge leans substantially on uniform exact-fill reporting behavior rather than purely on market clearing, reinforcing that log-fill target parameterization ($Z_{i,t}$) is the primary durable gain.
    \item \textbf{Broad MOET field and region controls:} Adding 10 broad field-cluster and 3 region dummies leaves 2024 CV and 2025 OOT accuracy unchanged (Appendix Table~\ref{tab:app_robustness}) while retaining Pillar~II as the top XGBoost TreeSHAP block at \textbf{30.9\%} (region 6.1\%, field 0.8\%).
    \item \textbf{Statutory $\bar{\phi} = 1.03$ censoring, weighting, winsorization, and $C_i$ checks:} Shifting censoring to $\ln(1.03)$ leaves Capped L1 invariant (12.14) but degrades Tobit median MAE to 14.44 by treating exact fills at $Z = 0$ as interior draws. Weighting log-ratio learners by $S_{i,t}$ preserves H-Soft's edge (11.44, $p = 0.027/0.005$); reclassifying the three keyword-matched private campuses (\texttt{DNC}, \texttt{DVB}, \texttt{KTD}) from $C_i = 25.5$ to $16.5$ preserves Private-tier gains (18.01, $p = 0.024/0.018$) and Pillar~II TreeSHAP (43.3\%); and widening $Z_{i,t}$ clips to $[-2.50, 1.00]$ leaves MAE unchanged (11.27).
    \item \textbf{Exact 1-to-1 linkage and broad-field $(u, f)$ aggregation:} Restricting evaluation to exact 1-to-1 MOET-code links (Appendix Table~\ref{tab:app_linkage_agg}, Panel~B) preserves all cross-family rankings in 2024 ($N = 3,801$) and 2025 ($N = 1,240$). Aggregating to 1,804 university-by-broad-field $(u, f)$ cells in 2024 (Panel~C) strengthens H-Soft's edge (32.02/86.8 vs.\ 36.40/103.6, $p = 0.004/0.008, R^2 = 0.969$).
\end{itemize}

\paragraph{Cluster-weighted conformal coverage in 2024.}
Under the strict three-way university split (Section~\ref{sec:conformal}), cluster-weighted locally scaled 90\% intervals around H-Soft \citep{Dunn2023} achieve conservative empirical coverage of \textbf{92.4\%} (width 85.8; quintiles 91.5\%--94.5\%; program-level 90.6\%, width 76.8). Under the inner-CV single-split calibration transferred to 2025, cluster-weighted intervals cover \textbf{89.3\%} (width 65.6; quintiles 86.5\%--91.6\%; Figure~\ref{fig:tails}(b)), whereas constant-width intervals cover 90.4\% overall by over-covering small programs (93.0\% in quintile~1) while under-covering large programs (86.8\% in quintile~5).

\subsection{Out-of-Time Check Within the Census (\texorpdfstring{2023 $\to$ 2024}{2023 to 2024})}
Training streamlined 13-feature models (Appendix~\ref{app:upstream}, setting $g_{\text{job},k,2022} = 0$) strictly on 2023 transitions (39.2\% exact fill) and testing on 2024 out of time reinforces both quota anchoring and intercept sensitivity (Appendix Table~\ref{tab:app_robustness}, Panel~A). Unanchored growth stacking trails severely (26.18/55.9 vs.\ quota 12.35/36.6, $p < 0.001$), whereas 2023-trained quota-anchored models slash squared error significantly: H-Soft reaches the lowest RMSE at 28.3 ($p_{\text{sq}} = 0.004, R^2 = 0.932$), QA-Stack posts 28.7 ($p_{\text{sq}} = 0.010$), and Tobit median reaches 30.1 ($p_{\text{sq}} = 0.033$). Because strict exact filling jumped from 39.2\% to 62.8\% in 2024, covariate-fitted 2023 regressions experience an upward intercept shift on MAE (Tobit 13.39, H-Soft 13.58; $p_{\text{abs}} \ge 0.115$; unhurdled QA-Stack 15.43, $p = 0.002$). Simple 2023 tier-median fill $\times S_{2024}$ attains the lowest MAE (11.95, $p_{\text{abs}} = 0.511$) because 2023 public tier medians already sat at $1.00$, while trailing H-Soft on tail RMSE (30.7 vs.\ 28.3).

\subsection{De-Confounded TreeSHAP Attribution: Why Target Choice Reveals Pricing and Institutional-Tier Signals}
Appendix Table~\ref{tab:app_features} reports exact TreeSHAP attributions \citep{Lundberg2020} across matched feature sets and losses. Holding all 19 features and L1 loss identical, unanchored log-growth trees (\textbf{G-19}) spend 87.9\% of XGBoost SHAP reconstructing mechanical quota shifts ($\ln(S_{i,t}/D_{i,t-1})$ 64.6\%, $g_{S,i,t}$ 23.4\%; 88.4\% in LightGBM), leaving Pillar~II at just \textbf{2.5\%} (2.3\% LGB). Switching only the target to $Z_{i,t}$ (\textbf{QA-19}) raises Pillar~II's combined tuition and institutional-tier attribution to \textbf{45.8\%} in XGBoost (\textbf{39.8\%} LGB): pure log tuition $\ln(1+T_{i,t})$ accounts for 11.7\% in XGBoost (5.3\% LGB), unshielded tuition burden $B_{i,t}^{\text{unsh}}$---which explicitly interacts tuition with the institutional-tier proxy $(24.5 - C_i)_+$ and therefore embeds both price and tier variation---accounts for 22.2\% (27.3\% LGB), and $C_i$ directly accounts for 11.9\% (7.1\% LGB), alongside lagged university exact-fill share $\bar{e}_{u(i),t-1}$ (24.9\% / 26.7\%) and quota scale $\ln(1+S_{i,t})$ (12.6\% / 9.7\%). The same tenfold jump holds in the 15-feature Huber pair (\textbf{G-15} \textbf{3.5\%} $\to$ \textbf{QA-15} \textbf{35.5\%}). Because $C_i$ and $B_{i,t}^{\text{unsh}}$ share variance with the four Pillar~IV charter dummies (which absorb only 3.7\% separately in XGBoost and 3.3\% in LightGBM; 4.0\% and 3.4\% including $\text{Tier}_{\text{MOET}}$), Pillar~II's 45.8\% share must be interpreted strictly as \emph{combined pricing and institutional-tier/charter predictive attribution} rather than pure price elasticity or causal demand response. Conversely, minor attribution on lagged vacancy-share growth $g_{\text{job},k,t-1}$ (0.3\%--0.8\%) reflects coarse nine-cluster pooling and universities already capitalizing expected labor demand into published quotas $S_{i,t}$ and peer seat expansions prior to prospectus release.

\section{Out-of-Time Evaluation in 2025 and Prospective 2026 Projections}
\label{sec:policy_shocks}

\subsection{The 2025 Admission Year}
To test whether census patterns survive a genuine temporal shock without look-ahead, we evaluate all specifications strictly out of time on 2025 using models trained solely on the 2024 census ($N = 5,385$), holding all stacking weights, econometric parameters, and hurdle thresholds fixed. Alongside steady national cohort expansion---MOET's high-school graduation exam candidate pool rose from roughly 1.01~million in 2022 and 1.02~million in 2023 to 1.07~million in 2024 and 1.17~million in 2025, reinforcing the system-wide 2023--2025 seat-weighted fill climb---three national policy shifts coincided in 2025:
\begin{enumerate}
    \item \textbf{Overhaul of graduation exams and admission rounds.} The 2025 cohort was the first to sit the four-subject reformed national exam under the 2018 General Education Programme \citep{MOET2023Decision4068,MOET2024Circular24}, while MOET abolished separate early-admission rounds so all methods cleared in one unified round and capped priority bonus points \citep{MOET2025Circular06}.
    \item \textbf{National semiconductor workforce initiative.} Decision 1017/Q{\DJ}-TTg (September 2024) established a target of training at least 50,000 semiconductor engineers by 2030 (including 5,000 AI specialists) backed by public laboratory investment \citep{PM2024Decision1017}.
    \item \textbf{Scheduled public tuition ceiling backdrop.} Decrees 97/2023/N{\DJ}-CP and 238/2025/N{\DJ}-CP stepped up statutory public tuition ceilings, allowing financially autonomous public universities to charge 2.0--2.5 times the standard cap \citep{Gov2023Decree97,Gov2025Decree238}. Because 87.8\% of 2025 prospectuses omit explicit fee tables and carry forward 2024 tuition (Table~\ref{tab:panel_overview}), these decrees form part of the macro-regulatory shock environment rather than directly observed program-level fee jumps in our 2025 panel.
\end{enumerate}

\subsection{Strictly Out-of-Time Accuracy in 2025}
\begin{table}[!ht]
\centering
\scriptsize
\setlength{\tabcolsep}{2.8pt}
\caption{Strictly out-of-time accuracy in 2025 (models trained on 2024 census only): Panel~A reports the unconditioned ex-ante benchmark on all 3,348 matched programs (158 universities) alongside the quota-stability diagnostic subset; Panels~B and~C report nowcast accuracy once $S_{2025}$ is published.}
\label{tab:holdout}
\begin{tabularx}{\textwidth}{L c r r r r r r}
\toprule
\textbf{Model} & \textbf{MAE [95\% CI]} & \textbf{RMSE} & \textbf{MedAE} & \textbf{$R^2$} & \textbf{DM $t$} & \textbf{$p_{\text{abs}}$} & \textbf{$p_{\text{sq}}$} \\
\midrule
\multicolumn{8}{@{}l}{\textbf{Panel A: Ex-ante horizon (no 2025 quota; reference: Macro OLS)}} \\
\multicolumn{8}{@{}l}{\quad \textit{Primary unconditioned all-matched sample} ($N = 3,348$, 158 universities)} \\
Naive carry-forward $D_{2024}$ & 72.86 $[64.7, 81.2]$ & 141.5 & 39.0 & 0.122 & $-2.05$ & 0.042 & 0.723 \\
$1.05 \times D_{2024}$ & 74.46 $[66.3, 82.9]$ & 143.3 & 39.4 & 0.100 & $-2.50$ & 0.014 & 0.456 \\
Macro OLS (3 predictors) & 67.86 $[61.0, 75.3]$ & 140.1 & 39.0 & 0.139 & --- & --- & --- \\
Linear-13 (Ridge, same 13 features as E0) & 68.43 $[61.6, 75.9]$ & 140.6 & 39.4 & 0.134 & $-3.48$ & $<$0.001 & 0.002 \\
\textbf{E0: Ex-ante stack} & 65.85 $[59.3, 73.1]$ & 134.0 & 38.6 & 0.213 & $+2.85$ & 0.005 & $<$0.001 \\
\multicolumn{8}{@{}l}{\quad \textit{Diagnostic subset conditioned on stable 2025 quotas} ($0.50 \le S_{25}/S_{24} \le 2.00$; $N = 2,179$, 146 unis)} \\
\textbf{Naive carry-forward $D_{2024}$} & 36.42 $[32.1, 41.1]$ & 62.0 & 22.0 & 0.748 & $+5.04$ & $<$0.001 & $<$0.001 \\
$1.05 \times D_{2024}$ & 36.93 $[32.7, 41.6]$ & 61.9 & 22.2 & 0.749 & $+4.46$ & $<$0.001 & $<$0.001 \\
Macro OLS (3 predictors) & 47.71 $[41.6, 54.5]$ & 95.0 & 25.2 & 0.409 & --- & --- & --- \\
Linear-13 (Ridge, same 13 features as E0) & 47.74 $[41.6, 54.6]$ & 95.0 & 25.3 & 0.408 & $-0.20$ & 0.843 & 0.906 \\
E0: Ex-ante stack & 43.25 $[37.9, 49.0]$ & 78.5 & 25.0 & 0.596 & $+4.68$ & $<$0.001 & $<$0.001 \\
\midrule
\multicolumn{8}{@{}l}{\textbf{Panel B: Primary unscreened nowcast on all matched 2025 programs} ($N = 3,348$, 158 unis; ref: $S_{2025}$)} \\
Published quota $S_{2025}$ & 20.21 $[16.6, 24.3]$ & 79.0 & 6.0 & 0.726 & --- & --- & --- \\
$0.95 \times S_{2025}$ & 20.94 $[17.6, 24.9]$ & 78.8 & 8.0 & 0.728 & $-2.28$ & 0.024 & 0.564 \\
Lagged fill $\times S_{2025}$ & 22.46 $[19.1, 26.3]$ & 80.1 & 9.0 & 0.718 & $-4.00$ & $<$0.001 & 0.272 \\
G0: Growth stack (15 feat, Huber) & 31.66 $[27.3, 36.2]$ & 99.0 & 12.1 & 0.571 & $-8.41$ & $<$0.001 & 0.001 \\
G-19: Growth stack (19 feat, L1+Huber) & 29.55 $[25.1, 34.2]$ & 97.2 & 10.0 & 0.586 & $-7.21$ & $<$0.001 & 0.002 \\
QA-15: Log-fill stack (15 feat, Huber) & 20.74 $[17.1, 24.8]$ & 78.7 & 7.7 & 0.728 & $-1.49$ & 0.137 & 0.668 \\
HLM random-intercept (REML) & 22.65 $[19.2, 26.8]$ & 80.3 & 9.5 & 0.717 & $-3.53$ & $<$0.001 & 0.416 \\
Capped L1 median quantile ($\min(Z,0)$) & 20.14 $[16.5, 24.2]$ & 78.9 & 6.1 & 0.727 & $+1.34$ & 0.182 & 0.151 \\
Upper-censored Tobit (median) & 20.73 $[16.9, 24.8]$ & 80.6 & 7.0 & 0.715 & $-1.33$ & 0.184 & 0.322 \\
Cragg-style two-part (EV, Duan) & 21.82 $[18.3, 26.0]$ & 79.9 & 8.8 & 0.720 & $-2.85$ & 0.005 & 0.584 \\
QA-Stack & 20.43 $[16.8, 24.5]$ & 78.7 & 7.2 & 0.728 & $-0.70$ & 0.487 & 0.681 \\
Jump-QA & 20.43 $[16.8, 24.5]$ & 78.8 & 7.0 & 0.728 & $-0.89$ & 0.376 & 0.716 \\
H-Hard & 20.17 $[16.6, 24.2]$ & 78.4 & 7.0 & 0.731 & $+0.19$ & 0.849 & 0.257 \\
\textbf{H-Soft} & 20.13 $[16.6, 24.1]$ & 78.5 & 6.7 & 0.730 & $+0.48$ & 0.632 & 0.323 \\
Cragg-ML (interior-only $D<S$) & 21.20 $[17.6, 25.4]$ & 79.3 & 8.2 & 0.724 & $-2.06$ & 0.041 & 0.808 \\
\midrule
\multicolumn{8}{@{}l}{\textbf{Panel C: Diagnostic grain-consistent 2025 nowcast} ($0.50 \le S_{25}/S_{24} \le 2.00$; $N = 2,179$, 146 unis; ref: $S_{2025}$)} \\
Published quota $S_{2025}$ & 18.51 $[15.1, 22.2]$ & 41.2 & 7.0 & 0.889 & --- & --- & --- \\
$0.95 \times S_{2025}$ & 19.48 $[16.2, 22.8]$ & 40.6 & 8.0 & 0.892 & $-2.84$ & 0.005 & 0.490 \\
Lagged fill $\times S_{2025}$ & 20.97 $[17.6, 24.5]$ & 42.9 & 9.0 & 0.879 & $-4.02$ & $<$0.001 & 0.423 \\
G0: Growth stack (15 feat, Huber) & 20.05 $[16.9, 23.4]$ & 41.4 & 8.6 & 0.888 & $-3.33$ & 0.001 & 0.933 \\
G-19: Growth stack (19 feat, L1+Huber) & 18.92 $[15.7, 22.4]$ & 40.3 & 7.3 & 0.894 & $-1.15$ & 0.251 & 0.593 \\
QA-15: Log-fill stack (15 feat, Huber) & 19.07 $[15.9, 22.4]$ & 40.1 & 7.5 & 0.895 & $-1.46$ & 0.147 & 0.553 \\
HLM random-intercept (REML) & 20.85 $[17.8, 24.0]$ & 41.5 & 9.6 & 0.887 & $-3.42$ & $<$0.001 & 0.907 \\
\textbf{Capped L1 median quantile ($\min(Z,0)$)} & 18.46 $[15.0, 22.0]$ & 40.9 & 6.6 & 0.890 & $+0.59$ & 0.556 & 0.243 \\
Upper-censored Tobit (median) & 18.77 $[15.3, 22.3]$ & 41.4 & 7.0 & 0.888 & $-0.93$ & 0.355 & 0.923 \\
Cragg-style two-part (EV, Duan) & 19.71 $[16.5, 23.0]$ & 40.0 & 8.6 & 0.895 & $-2.36$ & 0.019 & 0.546 \\
QA-Stack & 18.82 $[15.5, 22.2]$ & 40.0 & 7.3 & 0.895 & $-0.91$ & 0.364 & 0.492 \\
Jump-QA & 18.75 $[15.4, 22.2]$ & 40.1 & 7.0 & 0.895 & $-0.94$ & 0.348 & 0.497 \\
H-Hard & 18.56 $[15.2, 22.0]$ & 39.4 & 7.0 & 0.898 & $-0.17$ & 0.862 & 0.296 \\
H-Soft & 18.50 $[15.2, 21.9]$ & 39.6 & 7.0 & 0.897 & $+0.03$ & 0.979 & 0.341 \\
Cragg-ML (interior-only $D<S$) & 19.44 $[16.3, 22.7]$ & 40.0 & 8.1 & 0.895 & $-1.87$ & 0.063 & 0.567 \\
\bottomrule
\multicolumn{8}{@{}p{\textwidth}@{}}{\vspace{1pt}\scriptsize \textit{Notes:} As Table~\ref{tab:census} ($df = 157$ in all-matched, $df = 145$ in grain-consistent); all models use strictly lagged $t-1$ full-calendar-year vacancy growth. Bold marks the single lowest MAE in each panel (without asserting statistical superiority where $p > 0.05$). In Panel~A primary unconditioned ex-ante ($N=3,348$), E0 significantly beats Macro OLS ($t = +2.85, p = 0.005$), matched 13-feature Linear-13 Ridge ($t = +3.63, p_{\text{abs}} < 0.001, p_{\text{sq}} < 0.001$; grain-consistent $t = +4.74, p_{\text{abs}} < 0.001$), and Naive $D_{2024}$ ($p < 0.001$); conditioning on future quota stability in the diagnostic subset reverses the naive comparison to favor $D_{2024}$. In Panels~B and~C, H-Soft, H-Hard, QA-Stack, Tobit (median), and Capped L1 are statistically indistinguishable from $S_{2025}$ on aggregate MAE ($p \ge 0.18$), whereas unanchored growth stacks G0 and G-19 lose significantly ($p \le 0.001$ on primary Panel~B).}
\end{tabularx}
\end{table}

Table~\ref{tab:holdout} and Figure~\ref{fig:accuracy}(b) summarize strictly out-of-time 2025 performance across both planning horizons.

\paragraph{Ex-ante horizon (Panel~A: no 2025 quota; strictly lagged $t-1$ vacancies).}
On the primary unconditioned all-matched sample ($N = 3,348$, 158 universities), adding all 13 ex-ante covariates linearly (\textbf{Linear-13 Ridge}: 68.43/140.6) slightly underperforms three-predictor Macro OLS (67.86/140.1), whereas \textbf{E0} (65.85/134.0) significantly beats Macro OLS ($t = +2.85, p = 0.005$, $p_{\text{sq}} < 0.001$), matched Linear-13 Ridge ($t = +3.63, p_{\text{abs}} < 0.001, p_{\text{sq}} < 0.001$), and naive $D_{2024}$ (72.86/141.5; $p < 0.001$)---confirming that E0's unconditioned ex-ante lift reflects nonlinear ensemble gains rather than merely additional covariates. Conditioning diagnostically on stable future quotas ($N = 2,179$) favors inertia so Naive $D_{2024}$ (36.42) beats E0 (43.25), while E0 still beats both Macro OLS (47.71) and Linear-13 Ridge (47.74, $p < 0.001$).

\paragraph{Nowcast horizon (Panel~B primary all-matched $N = 3,348$; Panel~C diagnostic grain-consistent $N = 2,179$).}
As unified clearing and the collapse of uniform quota-copying (71 to 3 campuses) cut strict exact filling to 11.3\% in primary Panel~B (10.4\% in Panel~C), the audited quota posts 20.21 MAE and 79.0 RMSE in Panel~B (18.51/41.2 in Panel~C). Five lessons stand out:
\begin{enumerate}
    \item \textbf{Unanchored growth stacks remain dominated.} Across both Panel~B and Panel~C, G0, rule-adjusted G1--G2, G-19, and widened-clip G-19 all lose significantly to $S_{2025}$ ($p \le 0.001$; Table~\ref{tab:holdout} and Appendix Tables~\ref{tab:app_robustness}--\ref{tab:app_linkage_agg}), whereas quota-anchored QA-15 holds steady.
    \item \textbf{Quota-anchored models tie $S_{2025}$ on symmetric MAE while trimming RMSE, explained by classifier miscalibration.} In primary Panel~B, \textbf{H-Soft} (20.13 MAE, 78.5 RMSE; $p_{\text{abs}} = 0.632, p_{\text{sq}} = 0.323$), Capped L1 (20.14), and \textbf{H-Hard} (20.17) tie $S_{2025}$ alongside Tobit median and QA-Stack; diagnostic Panel~C repeats this tie (Capped L1 18.46, H-Soft 18.50, $p_{\text{abs}} = 0.979$, while H-Soft beats QA-Stack pairwise at $p = 0.021$). Classifier calibration diagnostics (Appendix Table~\ref{tab:app_robustness}) explain why H-Soft stays anchored at $S_{2025}$ on 69.2\% of Panel~C: as true exact filling collapsed from 62.8\% to 10.4\%, 2024's surge pushed mean predicted $\bar{\hat{p}}$ \emph{up} to 67.5\% (calibration-in-the-large $-57.1$~pp, Brier 0.452, slope 0.34).
    \item \textbf{Asymmetric economic loss sharply separates downside-protective models from ceiling anchors.} Evaluating 2:1 and 3:1 asymmetric pinball-loss ratios (Appendix Table~\ref{tab:app_linkage_agg}, Panel~D) resolves the symmetric 2025 tie cleanly. When over-forecasting (unfilled budgeted tuition seats) is penalized $2\times$--$3\times$ over-enrollment, downside-adjusted log-fill models significantly beat $S_{2025}$ (22.16/23.13) even out of time in 2025: interior-only \textbf{Cragg-ML} (20.75/20.53, $p = 0.011/{<}0.001$), \textbf{QA-Stack} (20.93/21.19, $p = 0.011/0.002$), \textbf{H-Soft} (21.39/22.01, $p = 0.050/0.031$), and Capped L1 ($p \le 0.001$). Conversely, when Decree~04/2021 over-admission crowding/fine risk is penalized $2\times$--$3\times$, ceiling-anchored $S_{2025}$ (18.27/17.30) outperforms every downside-adjusted model ($p \le 0.028$ at 3:1).
    \item \textbf{Sensitivity checks confirm stability and document regime-shift limits of hard caps and campus BLUPs.} Dropping $\bar{e}_{u,t-1}$, mixed-reporting training, $S$-weighting, pooled $2023+2024$ training, iterative Powell CLAD (20.81/79.3 in Panel~B), and near-fill $D \ge 0.98\,S$ targeting preserve broad parity (Appendix Tables~\ref{tab:app_robustness}--\ref{tab:app_linkage_agg}). By contrast, interior-only Cragg-ML (21.20 MAE in Panel~B, $p = 0.041$; 19.44 in Panel~C) and Poisson GLM ($p = 0.001$) incur symmetric MAE losses because Eq.~\eqref{eq:cragg_ml} hard-caps Cragg-ML at $S_{i,t}$ without Duan smearing, missing 2025's 21.0\% surge in $>1.05\,S_{i,t}$ overshoots. Likewise, adding 2024 REML HLM Empirical Bayes BLUPs ($\hat{u}_g$, conditional $\widehat{\text{ICC}} = 0.479$) for the 157 returning campuses slightly worsens 2025 accuracy relative to population-averaged HLM ($p < 0.001$ in both panels), confirming that unified clearing shifted campus intercepts.
    \item \textbf{Tier stratification, wild cluster bootstrap, IPW, and MNAR tipping-point bounds.} Table~\ref{tab:tier_acc} reports both grain-consistent (Panel~B) and all-matched (Panel~C) tier breakdowns. Public campuses show a modest quota edge at Flagship institutions ($p = 0.026$ Panel~B; $p = 0.017$ Panel~C), whereas Private campuses show directional gains in both panels (33.93/70.6 vs.\ 35.38/76.8 in Panel~B; 36.37/171.9 vs.\ 37.46/173.5 in Panel~C). Finite-sample restricted wild cluster bootstrap inference (Rademacher weights, $B = 2,000$; Table~\ref{tab:tier_acc} notes) confirms that all small-cluster 2025 subgroup differences remain imprecise: Private $p_{\text{WCB}} = 0.346/0.357$ ($G = 21$) and $0.307/0.391$ ($G = 22$); International $p_{\text{WCB}} = 0.508/0.508$ ($G = 5$; Webb 6-point $0.508/0.508$) and $0.509/0.509$ ($G = 6$; Webb 6-point $0.469/0.485$). IPW reweighting (Appendix Table~\ref{tab:app_attrition}, Panel~C; treating estimated first-stage propensity weights as fixed) preserves null aggregate hurdle differences ($p_{\text{abs}} \ge 0.39$) while G0 remains significantly worse ($p \le 0.013$). Finally, MNAR pattern-mixture tipping-point bounds (Appendix Table~\ref{tab:app_attrition} notes) show that overturning 2025's core target-parameterization gain (G0 vs.\ H-Soft: $+11.54$ MAE on $w_{\text{obs}} = 0.627$) would require the 141 missing campuses ($w_{\text{miss}} = 0.373$) to reverse by an implausible $-19.4$ MAE favoring unanchored growth, whereas in the Private tier ($w_{\text{obs}} = 0.484$), the $+1.09$ MAE directional edge reverses at just $-1.02$ MAE.
\end{enumerate}

\subsection{Where the Published Quota Fails}
\begin{figure}[t]
    \centering
    \includegraphics[width=\textwidth]{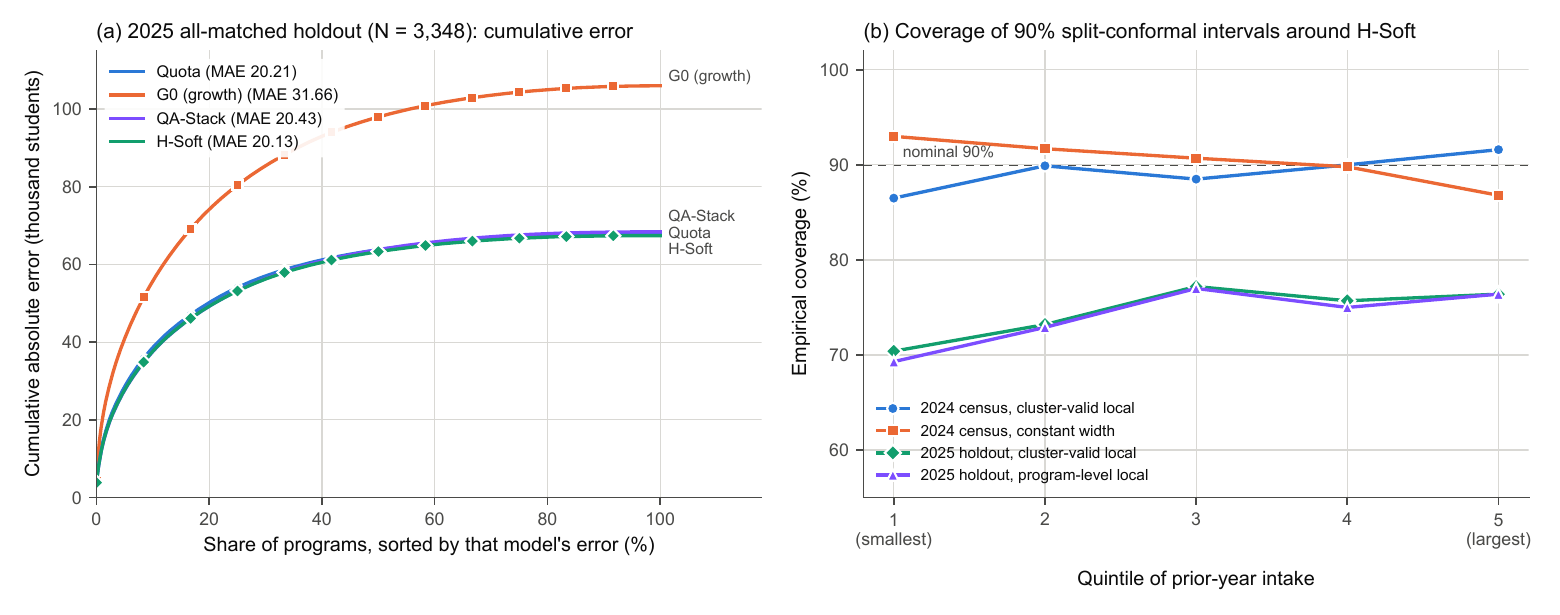}
    \caption{(a) Cumulative absolute error on the primary unscreened 2025 all-matched holdout ($N = 3,348$; quota MAE 20.21 vs.\ G0 31.66), sorted from largest to smallest miss (circles: Quota; squares: G0 growth; triangles: QA-Stack; diamonds: H-Soft). (b) Empirical coverage of 90\% split-conformal intervals around H-Soft across quintiles of prior-year intake: 2024 census (circles: cluster-valid local; squares: constant width) and 2025 holdout (diamonds: cluster-valid local; triangles: program-level local scaling)}
    \label{fig:tails}
\end{figure}

In the grain-consistent subset, the worst 5\% of programs generate 36.4\% of total quota absolute error (35.0\% under H-Soft, 34.5\% under QA-Stack, 33.0\% under G0); across the full all-matched 2025 population ($N = 3,348$; Figure~\ref{fig:tails}(a)), H-Soft and QA-Stack hug the quota cumulative-error curve while unanchored G0 rises roughly 38,000 students higher.

Table~\ref{tab:misses} inspects the ten widest 2025 quota misses alongside prior-year fill ratios $\Phi_{24}$. Eight are at private campuses---seven shortfalls (B--E, H--J) plus private Chinese language over-enrollment (A: $+75.5\%$, 1,578 vs.\ 899)---and two are at public campuses: business administration shortfall G (120 vs.\ 400) and agricultural economics over-enrollment F ($+370.0\%$, 376 vs.\ 80). Extreme single-program overshoots typically reflect post-prospectus seat reallocation across sibling majors within the same broad-field ceiling (Eq.~\eqref{eq:moet_cap}) or sub-track merging. Directly testing Eq.~\eqref{eq:moet_cap} by aggregating 2025 programs into 795 university-by-broad-field $(u,f)$ cells (Appendix Table~\ref{tab:app_linkage_agg}, Panel~C) cuts apparent $>1.05$ overshoots by one-third (from 23.1\% to 15.3\%; $>1.20$ from 8.1\% to 5.9\%), while H-Soft reduces $(u,f)$-level error from 36.35/82.6 to 34.12/71.4 ($R^2 = 0.982$, $p = 0.135/0.186$). The remaining 15.3\% broad-field overshoots align with strategic yield-hedging over-offers under 2025's newly unified clearing round \citep{CheKoh2016}, when universities faced higher matriculation-yield uncertainty after early admissions were abolished. At the program level, H-Soft and QA-Stack move \emph{all eight shortfalls} (B--E, G--J) closer to realized intake (cutting error by 253 students at E, 176 at B, 123 at C, 59 at I, and 45 at J) while leaving public over-enrollment F untouched at 80; Tobit (median) and E0 improve on seven of eight. Yet where prior-year compliance was high (H, J, G)---or at Program~D ($\Phi_{24} = 0.55$, where high host-university exact-fill probability kept H-Soft at 572 vs.\ 600 while Tobit shifted to 495)---2024-trained models under-correct abrupt drops, underscoring why automated screens complement rather than replace manual review.

\begin{table}[t]
\centering
\scriptsize
\setlength{\tabcolsep}{3pt}
\caption{The ten 2025 programs where the published quota missed realized intake by the widest margin (grain-consistent sample, $N = 2,179$).}
\label{tab:misses}
\begin{tabularx}{\textwidth}{L c r r r r r r r}
\toprule
\textbf{Field (institution type)} & \textbf{ID} & \textbf{$\Phi_{24}$} & \textbf{Quota} & \textbf{Intake} & \textbf{H-Soft} & \textbf{QA-Stack} & \textbf{Tobit} & \textbf{E0} \\
\midrule
Chinese language (private) & A & 1.00 & 899 & 1578 & 869 & 827 & 789 & 591 \\
Law (private) & B & 0.58 & 724 & 209 & 548 & 528 & 535 & 240 \\
Economics (private) & C & 0.43 & 513 & 124 & 390 & 379 & 374 & 168 \\
Medical laboratory technology (private) & D & 0.55 & 600 & 216 & 572 & 557 & 495 & 348 \\
Business administration (private) & E & 0.77 & 1040 & 712 & 787 & 769 & 743 & 555 \\
Agricultural economics (public) & F & 0.92 & 80 & 376 & 80 & 80 & 80 & 93 \\
Business administration (public) & G & 0.78 & 400 & 120 & 398 & 392 & 400 & 120 \\
Pharmacy (private) & H & 0.92 & 300 & 28 & 285 & 278 & 281 & 322 \\
Japanese language (private) & I & 0.73 & 324 & 62 & 265 & 256 & 262 & 162 \\
Information technology (private) & J & 1.00 & 960 & 709 & 915 & 882 & 817 & 506 \\
\bottomrule
\multicolumn{9}{@{}p{\textwidth}@{}}{\vspace{1pt}\scriptsize \textit{Notes:} Programs ranked by $|D_{2025}-S_{2025}|$. Letters anonymize host institutions; $\Phi_{24} = D_{2024}/S_{2024}$ is the prior-year fill ratio. H-Soft, QA-Stack, and Tobit (median) are 2024-trained nowcasts; E0 is the 2024-trained ex-ante stack.}
\end{tabularx}
\end{table}

\subsection{Prediction Intervals Under Regulatory Regime Shift}
Projected without updating onto 2025, 2024-calibrated 90\% cluster-weighted local conformal intervals around H-Soft drop from 89.3\% to \textbf{74.6\%} coverage on the grain-consistent subset (width 51.2; quintiles 70.4\%--77.2\%; Figure~\ref{fig:tails}(b)), edging program-level intervals (74.1\%, width 49.7; all-matched Panel~B: 74.0\% vs.\ 73.4\%). Transferring the strict three-way university-split 2024 calibration (zero inner-CV residual reuse) yields slightly lower 2025 cluster-weighted coverage of \textbf{71.7\%} on the grain-consistent subset (width 49.3; \textbf{71.3\%} on all-matched Panel~B, width 53.2), confirming that 2025 under-coverage stems from regulatory regime shift and volatility covariate shift rather than inner-CV calibration reuse. Constant-width bands reach 83.3\% average coverage (width 62.2) only by over-covering quintile~1 (92.0\%) while collapsing to \textbf{65.8\%} in quintile~5---over 10~pp below cluster-weighted scaling (76.4\%). Notably, local width \emph{shrank} from 65.6 to 51.2 even as H-Soft MAE rose from 11.32 to 18.50: 2024's exact-fill surge raised lagged exact fill from 38.7\% to 62.8\% and cut mean $|\Phi_{i,t-1} - 1|$ from 0.165 to 0.065, causing the 2024 volatility model to predict a 19.4\% smaller average scale ($\bar{\hat{\sigma}}$ falling from 5.27 to 4.25) right before reform-year dispersion widened---illustrating non-exchangeability \citep{Tibshirani2019,Barber2023}. Because Vietnam's centralized clearing resolves simultaneously once per year, online adaptive conformal updates \citep{Gibbs2021} operate at an \emph{inter-annual} rather than within-year cadence.

\subsection{Pooled 2023--2025 Retraining and Bias-Reconciled 2026 Aggregate Projections}
\label{sec:pooled_2026}
Refitting the quota-anchored pipeline on all $N = 8,733$ continuing transitions ($2023 \to 2024$ plus $2024 \to 2025$, 299 unis) under five-fold university-grouped CV---measuring cross-university generalization across pooled waves---yields MAE/RMSE of 15.37/56.8 ($R^2 = 0.799$) for published quotas, 15.01/54.3 for QA-Stack, and \textbf{14.81/54.8} for \textbf{H-Soft} ($p_{\text{abs}} = 0.055, p_{\text{sq}} = 0.023, R^2 = 0.813$).

Applying this pooled fit to the 2026 prospectus wave---5,755 programs across the 257 prospectus-reporting universities (43 fewer than the 300-university 2024 panel), totaling 690,586 published plus 2,585 imputed seats across 38 missing-quota rows (\textbf{693,171 effective seats}, filled via a strictly backward/published cascade: linked 2025 quota/intake $\to$ 2024 quota/intake $\to$ campus or 7-digit MOET median published quota $\to$ 60-seat default; zero 2026 outcome data)---Appendix Table~\ref{tab:app_attrition} (Panel~A) audits 2026 non-reporting composition: 250 of the 298 census universities reported 2026 prospectuses alongside 7 non-census campuses, while the 48 non-reporting census campuses are heavily private (49.5\% private program share vs.\ 20.3\% among reporters, with slightly smaller 2024 intakes of 107.1 vs.\ 114.7), underscoring why our 2026 aggregates are scoped strictly to the 257 reporting campuses. Appendix Table~\ref{tab:app_proj2026} summarizes two reconciliations:
\begin{itemize}
    \item \textbf{Ceiling and median-to-sum bias.} Summing raw pooled H-Soft forecasts yields \textbf{665,742 students} (\textbf{96.0\%} fill), above our scenario band's 95th percentile (\textbf{[604,533, 662,044]}). This $-5.48\%$ pooled deviation ($+5.80\%$ over-prediction; $-7.41\%$ in 2024, $-2.26\%$ in 2025) stems from left-skewed median summation ($\sum_i \operatorname{Med}(D_i) > \sum_i \mathbb{E}[D_i]$) and H-Soft blending $S_{i,t}$ with an unconditional QA-Stack. Confirming this directly, pooled interior-only \textbf{Cragg-ML} yields \textbf{629,077 (90.8\%)}---within 179 students of \textbf{macro-ratio-calibrated H-Soft (629,256; 90.8\%)} and matching conditional-mean \textbf{Cragg EV (627,114; 90.5\%)}, while 2025-only H-Soft gives \textbf{651,620 (94.0\%)} and pooled exact \textbf{Tobit EV} sits at \textbf{601,858 (86.8\%)}.
    \item \textbf{Within-regime CI, two-regime scenario band ($T = 2$), and tier breakdown.} University-residual resampling ($B = 2,000$) around calibrated H-Soft gives a \textit{within-regime} 90\% CI of \textbf{[614,756, 642,798]} (88.7\%--92.7\% fill; excluding macro shocks). Because pooled $-5.48\%$ calibration weights pre-reform 2024 ($N = 5,385$) more heavily than post-reform 2025 ($N = 3,348$), mixing both annual shifts yields our \textit{two-regime historical scenario band} of \textbf{[604,533, 662,044]} (87.2\%--95.5\% fill)---a sensitivity bound rather than a frequentist inter-annual interval---bracketing every calibrated, interior-hurdle, Cragg EV, and 2025-only estimate (627,114--651,620). Stratifying by institutional governance tier (Appendix Table~\ref{tab:app_proj2026}, Panel~B) reveals sharp operational divergence across uniform calibrated H-Soft and Cragg-ML: Flagship Public projects 93.7\% fill (167,723--167,804 of 179,084 seats), Other Public 92.2\%--92.4\% (293,699--294,605 of 318,711), Private 85.3\%--85.6\% (158,964--159,540 of 186,358 seats, ~27,000 unfilled seats), and International 88.3\%--89.1\% (7,963--8,034 of 9,018 seats), agreeing within 0.8~pp in every tier. Replacing uniform $-5.48\%$ scaling with tier-specific pooled CV ratio calibration (Panel~B) leaves the national total virtually unchanged (\textbf{629,233}, 90.8\%, within 23 students of 629,256) while sharpening the public (93.2\%--94.7\%) versus private (83.2\%) and international (84.1\%) fill gap.
\end{itemize}

\section{Implications, Limitations, and Conclusion}
\label{sec:implications}

\subsection{Implications for Regulators, Market Analysts, and Campus Planners}
Although drawn from one national system, the short-side trade-off we document---statutory capacity ceilings paired with heavy-tailed shortfalls at tuition-dependent private campuses---applies broadly across centralized higher education markets. Because spring public-prospectus nowcasts appear 30--60 days before matriculation without proprietary applicant funnels, their primary operational audience is \emph{ministries, regulators, and external market-intelligence analysts} monitoring system-wide capacity across hundreds of campuses, while internal provosts pair private applicant-funnel data with prospectus benchmarks for competitor context. Five operational principles follow:

\begin{enumerate}
    \item \textbf{Anchor forecasts to audited quotas, and prioritize parsimony.} Unanchored log-growth ML degrades accuracy even under identical features, L1 loss, and widened non-binding clipping. Analysts should model the log fill ratio $Z_{i,t} = \ln(D_{i,t}/S_{i,t})$ and match complexity to institutional objectives: single-equation \textbf{upper-censored Tobit (median)} (in both 90\% MCS sets) and \textbf{Capped L1} (matching H-Soft in 2025) offer transparent defaults, and continuous \textbf{QA-Stack} beats \textbf{H-Soft} on both MAE and RMSE once 2024 uniform-exact reporters are excluded. Duan-smeared Cragg EV and exact Tobit EV minimize squared error and provide conditional-mean aggregates, while interior-only \textbf{Cragg-ML} empirically removes raw hurdle ceiling sum bias (noting that hard-capping at $S_{i,t}$ without Duan smearing is not analytically unbiased and incurs downside bias under 2025 overshoots); \textbf{H-Soft} adds targeted value mainly when planners prioritize L1 headcount error under stable exact-quota reporting.
    \item \textbf{Differentiate budgeting by governance tier and price--prestige exposure.} At selective public campuses, audited quotas remain reliable defaults and review should target abrupt quota jumps. At private and international campuses---where matched-feature TreeSHAP assigns roughly two-fifths of shortfall attribution to tuition and institutional-tier prestige ($C_i$, a four-level charter/name proxy distinct from program-level admission cut-off scores \textit{{\dj}i{\^{e}}m chu{\^{a}}n})---quota-anchored models warrant downside stress testing before committing fixed operating overheads.
    \item \textbf{Separate autumn envelopes from spring adjustments.} Because ex-ante error is three to five times wider than spring nowcast error, autumn plans should serve as scenario bands while spring schedules incorporate published competitor quotas. Remark~\ref{rem:equivalence} further warns against early-sample ratio extrapolation that magnifies phantom capacity by $\omega^{-1}$.
    \item \textbf{Mandate fill-history disclosure while guarding against punitive automation.} Printing each program's three-year realized fill history ($\Phi_{i,t-3:t-1}$) beside $S_{i,t}$ under Circulars~03/2022 and~08/2022---and auditing uniform 100\%-exact reporters---would curb phantom-quota asymmetry at low cost. Regulators should treat shortfall flags strictly as audit triggers rather than automated grounds for sanctions without independent verification.
    \item \textbf{Pair adaptive, cost-aligned intervals with mean-consistent aggregates.} Given 2025's conformal coverage drop, planning offices should adopt inter-annual online adaptive updates \citep{Dunn2023,Gibbs2021}. Our 2:1 and 3:1 asymmetric pinball-loss sensitivity (Appendix Table~\ref{tab:app_linkage_agg}, Panel~D) confirms that model choice hinges on campus cost structure: tuition-dependent institutions facing fixed-overhead shortfall costs benefit significantly from downside-adjusted log-fill models (\textbf{Cragg-ML}, \textbf{QA-Stack}, \textbf{H-Soft}) even out of time in 2025, whereas selective public campuses facing Decree~04/2021 over-admission fines remain best served by ceiling-anchored $S_{i,t}$ paired with asymmetric conformalized quantile regression \citep{Romano2019}. Multi-campus totals should use macro-ratio calibration, Cragg-ML, or conditional-mean Cragg/Tobit EV sums rather than raw hurdle sums.
\end{enumerate}

\subsection{Limitations and Future Research}
Four limitations bound our findings:
\begin{itemize}
    \item \textbf{Out-of-time precision and prospective 2026 validation.} H-Soft's 2024 MAE gain clears 10\% not 5\% candidate FDR and is non-significant under fold-overlap adjustment; 2025 aggregate MAE differences are statistically null, Private-tier gains are imprecise across 21 retained clusters, and exact-fill base-rate collapse (62.8\% to 10.4\%) severely miscalibrates the hurdle classifier. Our prospective 2026 totals across 257 reporting campuses provide a direct future test.
    \item \textbf{Non-random 2025 attrition (MNAR) and 2024 reporting anomalies.} Nearly half of 2024 census universities (141/298, 47.3\%) had not disclosed 2025 intake by our cutoff, disproportionately dropping private and international campuses. IPW reweighting (treating first-stage weights as fixed) adjusts for MAR observables but cannot rule out MNAR non-disclosure; separating binding rationing from copy-forward reporting in 2024 requires ministerial audit microdata.
    \item \textbf{Measurement of selectivity, tuition, and labor signals.} Selectivity $C_i$ is a four-level institutional proxy collinear with charter dummies rather than digitized program-level cut-off scores, and 87.8\% of 2025 tuition carries forward 2024 values, precluding direct out-of-time tests of Decree~97/2023 fee hikes. Constant wage and cohort columns were excluded, and TreeSHAP reflects predictive attribution rather than causal elasticities.
    \item \textbf{Macro-shock variance ($T = 2$) and asymmetric intervals.} Observing only $T = 2$ transition waves restricts 2026 inter-annual uncertainty to a two-regime historical scenario band rather than a frequentist predictive interval. Implementing asymmetric conformalized quantile regression \citep{Romano2019} with online adaptive updates \citep{Gibbs2021,Barber2023} and extending to multi-year horizons or applicant-preference microdata remain priorities.
\end{itemize}

\subsection{Conclusion}
Across Vietnam's audited 2022--2026 panel (26,895 records, 309 universities) with strictly lagged $t-1$ vacancy signals, published quotas anchor compliant public programs while failing severely in shortfall tails at tuition-dependent private and international campuses. Target ablations---including a pure single-factor control under widened non-binding clipping---establish that modeling the log fill ratio $Z_{i,t} = \ln(D_{i,t}/S_{i,t})$ rather than unanchored log growth sharply lowers error, aligns ML with canonical multilevel and censored baselines, and redirects matched-feature TreeSHAP from mechanical quota tracking toward tuition and institutional-tier prestige.

Equally, our benchmark underscores model parsimony: hurdle stacking (\textbf{H-Hard}, \textbf{H-Soft}) leads the full 2024 90\% MAE MCS, yet continuous \textbf{QA-Stack} outperforms H-Soft on both MAE and RMSE once uniform-exact 2024 campuses are excluded, canonical \textbf{upper-censored Tobit (median)} ties H-Soft while entering \emph{both} MCS sets, interior-only \textbf{Cragg-ML} achieves low RMSE and empirically removes raw ceiling sum bias, and single-equation \textbf{Capped L1} matches H-Soft out of time in 2025. Across 257 prospectus-reporting universities in 2026 (693,171 effective seats), reconciled totals converge tightly inside our two-regime scenario band. For ministries, external market analysts, and campus planners, the operational recipe is concise: start from audited quotas, model deviations in log fill-ratio space with parsimonious censored or continuous/hurdle estimators, audit tuition-exposed tails, and pair forecasts with adaptive, cost-aligned intervals.

% -------------------------------------------------------------
% REFERENCES
% -------------------------------------------------------------

% -------------------------------------------------------------
% APPENDICES (Self-contained inside the arXiv preprint)
% -------------------------------------------------------------
\clearpage
\begin{appendices}
\section{Upstream Measurement, Bipartite Linkage, Streamlined Features, and IPW Diagnostics}
\label{app:upstream}

\paragraph{1. Predetermined selectivity-tier lookup ($C_i$).}
In the upstream institutional reference table (\nolinkurl{enriched_program_dataset_phase2.csv}, 7,427 program entries), entry selectivity $C_i$ maps deterministically from a predetermined institutional-name and charter-type tier column (\texttt{Selectivity\_Tier}) prior to any enrollment outcome observation (zero outcome leakage):
\begin{itemize}
    \item $C_i = 25.5$ (\textit{Tier~A Competitive}, 2,286 entries): 2,169 flagship/major public, 29 international, 15 specialized higher-education, and 73 private entries across three institutions (\texttt{DNC}, \texttt{DVB}, \texttt{KTD}) matched by the compiler's predetermined institutional-name keyword rule on regional-hub/economic terms prior to charter filtering;
    \item $C_i = 23.0$ (\textit{Tier~A International}, 109 entries): 109 international-campus program listings;
    \item $C_i = 21.0$ (\textit{Tier~B Standard Public}, 3,298 entries): 3,298 standard public university listings (also the default fallback when an institution code is unseen);
    \item $C_i = 16.5$ (\textit{Tier~C Open-Access/Floor}, 1,734 entries): 1,681 open-access private and 53 other private/college listings.
\end{itemize}
During panel assembly, each (university, 7-digit MOET) pair receives its modal lookup value (falling back to university median, then $21.0$). Because $C_i$ derives deterministically from institutional names and charter types rather than program-level cut-off scores, it is strongly collinear with Pillar~IV charter indicators.

\paragraph{2. Adjacent-year bipartite linkage scoring.}
Within each (university, 7-digit MOET) cell containing one current-year and one prior-year row, programs link directly (\texttt{1to1\_um7}). Where multiple tracks share a (university, 7-digit MOET) cell, optimal bipartite matching (\texttt{scipy.optimize.linear\_sum\_assignment}) minimizes the negative similarity cost between current row $i$ and prior row $j$:
\begin{align}
    \text{Cost}(i, j) = -\Big[&2.0\,\mathbb{I}(M^{\text{exact}}_i = M^{\text{exact}}_j) + 1.5\,\mathbb{I}(\text{PT}_i = \text{PT}_j) \nonumber \\
    &+ 3.0\,\text{sim}_{\text{tok}}(P_i, P_j) + \max\!\big(0,\; 1 - |\ln(S_i/S_j)|\big)\Big],
\end{align}
where $M^{\text{exact}}$ is the full sub-code, $\text{PT}$ is the normalized track category (standard, high-quality, joint/international, advanced), $\text{sim}_{\text{tok}}$ is Jaccard token similarity on accent-stripped titles, and $S$ uses quota (or intake if quota is missing). Excess unmatched current rows in a multi-track cell receive quota-proportional shares of the cell's prior total (\texttt{group\_share\_um7}). When a 7-digit MOET code changes between years, a within-university title fallback (\texttt{pname\_u}) links to the highest-scoring prior row at the same institution provided $\text{sim}_{\text{tok}}(P_i, P_j) + 0.5\,\mathbb{I}(\text{PT}_i = \text{PT}_j) \ge 0.65$.

\paragraph{3. Streamlined 13-feature specification for $2023 \to 2024$ and pooled $2023+2024 \to 2025$ checks.}
Because published 2022 quotas $S_{i,2022}$ are unobserved in the panel (Table~\ref{tab:panel_overview}), lagged quota growth $g_{S,i,t}$, lagged fill $\Phi_{i,t-1}$, lagged quota-history indicators, and two-year intake momentum $g_{\text{hist}}$ cannot be computed for 2023 transitions. Streamlined models therefore use the 13 covariates available identically across 2023, 2024, and 2025: $\ln(1 + D_{i,t-1})$, $\ln(1 + S_{i,t})$, clipped quota-to-lagged-intake change $r_{\text{quota},i,t} = \operatorname{clip}((S_{i,t} - D_{i,t-1})/\max(D_{i,t-1},1), -0.55, 0.85)$, leave-one-institution-out peer ratio growth $r^{\text{peer}}_{-u(i),k,r,t} = \operatorname{clip}(\sum_{j \notin \mathcal{I}_{u(i)}}(S_{j,t}-D_{j,t-1})/\max(\sum D_{j,t-1}, 50), -0.30, 0.50)$, $\ln(1+T_{i,t})$, $C_i$, $B_{i,t}^{\text{unsh}}$, lagged vacancy growth $g_{\text{job},k,t-1}$, the four charter dummies ($\mathbb{I}_{\text{flagship}}, \mathbb{I}_{\text{public}}, \mathbb{I}_{\text{private}}, \mathbb{I}_{\text{intl}}$), and $\text{Tier}_{\text{MOET}}$. Because full-year vacancy totals begin in 2022 (leaving 2021 unobserved), $g_{\text{job},k,t-1}$ is initialized at the neutral zero-growth fallback ($0.0$) for the auxiliary $t = 2023$ training wave---mirroring the $g_{\text{hist}} = 0.0$ fallback when $D_{i,t-2}$ is unobserved---so no contemporaneous 2023 vacancy information enters any 2023-trained model.

\paragraph{4. IPW propensity specification and Kish Effective Sample Size (ESS).}
Retention probabilities $\hat{e}_u = \widehat{\Pr}(\text{retained}_{2025} = 1 \mid \mathbf{v}_{u,2024})$ are estimated via $L_2$-regularized university-level logistic regression ($C = 0.5$) on nine 2024 campus characteristics ($\ln(1+n_u)$, $\ln(1+\bar{D}_{u,24})$, $\bar{\Phi}_{u,24}$, $\bar{e}_{u,24}$, $\mathbb{I}_{\text{flagship}}$, $\mathbb{I}_{\text{private}}$, $\mathbb{I}_{\text{intl}}$, North, South), clipped to $[0.15, 0.95]$ to prevent extreme leverage. Weight dispersion is modest: in the grain-consistent 2025 holdout ($N = 2,179$, 146 universities), Kish $\text{ESS} = (\sum w)^2 / \sum w^2$ equals $1{,}958.0$ programs (89.9\%) and $120.2$ universities (82.3\%), with university weights spanning $[1.13, 6.12]$ (median $1.69$); in the all-matched holdout ($N = 3,348$, 158 universities), Kish ESS equals $3{,}134.1$ programs (93.6\%) and $142.0$ universities (89.9\%), with weights in $[1.00, 4.01]$ (median $1.63$), where the $1.00$ minimum reflects the fallback weight assigned to the single non-census 158th university absent from the 2024 propensity fit (the 157 census-matched campuses satisfy $w_u \ge 1/0.95 = 1.053$). IPW clustered Diebold--Mariano tests treat first-stage propensity weights as fixed (a first-order approximation).

\section{Data Audits, Feature Dictionary, TreeSHAP, and Robustness Tables}
\label{app:feature_dict}

Table~\ref{tab:app_attrition} reports three empirical integrity checks referenced in Sections~\ref{sec:data}, \ref{sec:benchmarks}, and \ref{sec:policy_shocks}: (Panel~A) a baseline comparison of 2024 census universities retained versus dropped in 2025 disclosures; (Panel~B) a 2024 census sensitivity analysis excluding the 71 universities reporting 100\% exact fills across every listed program ($N = 4,108$ mixed-reporting programs); and (Panel~C) inverse-probability-weighted (IPW) 2025 out-of-time accuracy with weighted university-clustered Diebold--Mariano $p$-values.

\begin{table}[!ht]
\centering
\scriptsize
\setlength{\tabcolsep}{2.8pt}
\caption{Data-integrity diagnostics: (Panel~A) 2025 intake and 2026 prospectus university attrition audit relative to the 298-university 2024 census; (Panel~B) 2024 census sensitivity excluding 71 universities reporting 100\% exact fills ($N = 4,108$ mixed-reporting programs); and (Panel~C) inverse-probability-weighted (IPW) 2025 holdout accuracy with weighted clustered Diebold--Mariano $p$-values.}
\label{tab:app_attrition}
\begin{tabularx}{\textwidth}{L r r c c c c c c}
\toprule
\multicolumn{9}{@{}l}{\textbf{Panel A: 2025 \& 2026 university attrition audit (2024 baseline characteristics of 298 census unis)}} \\
\textbf{Sample / Status} & \textbf{Unis} & \textbf{2024 Progs} & \textbf{Public} & \textbf{Private} & \textbf{Intl.} & \textbf{Mean $D_{24}$} & \textbf{Mean $\Phi_{24}$} & \textbf{Exact $2024$} \\
\midrule
All matched 2025 (Retained) & 157 & 3,377 & 80.4\% & 18.4\% & 1.2\% & 115.5 & 0.939 & 65.5\% \\
All matched 2025 (Dropped) & 141 & 2,008 & 60.8\% & 33.0\% & 6.2\% & 110.9 & 0.917 & 58.2\% \\
Grain-consistent 2025 (Retained) & 146 & 3,334 & 80.4\% & 18.7\% & 0.9\% & 115.8 & 0.938 & 65.3\% \\
Grain-consistent 2025 (Dropped) & 152 & 2,051 & 61.2\% & 32.3\% & 6.5\% & 110.5 & 0.919 & 58.8\% \\
2026 prospectus wave (Reported) & 250 & 4,731 & 76.8\% & 20.3\% & 2.9\% & 114.7 & 0.932 & 62.8\% \\
2026 prospectus wave (Non-reporting) & 48 & 654 & 46.2\% & 49.5\% & 4.3\% & 107.1 & 0.920 & 62.5\% \\
\midrule
\multicolumn{9}{@{}l}{\textbf{Panel B: 2024 mixed-reporting sensitivity} ($N = 4,108$ programs, 227 unis; excl.\ 71 uniform-exact unis)} \\
\textbf{Model} & \multicolumn{2}{c}{\textbf{MAE [95\% CI]}} & \textbf{RMSE} & \textbf{MedAE} & \textbf{$R^2$} & \textbf{DM $t$} & \textbf{$p_{\text{abs}}$} & \textbf{$p_{\text{sq}}$} \\
\midrule
Published quota $S_t$ & \multicolumn{2}{c}{16.19 $[13.3, 19.8]$} & 41.9 & 0.0 & 0.858 & --- & --- & --- \\
$0.95 \times S_t$ & \multicolumn{2}{c}{15.74 $[13.4, 18.8]$} & 36.8 & 5.5 & 0.890 & $+1.11$ & 0.270 & $<$0.001 \\
G0: Growth stack & \multicolumn{2}{c}{24.52 $[21.9, 27.4]$} & 53.7 & 9.5 & 0.767 & $-4.64$ & $<$0.001 & 0.014 \\
HLM (REML) & \multicolumn{2}{c}{15.83 $[14.2, 18.0]$} & 30.8 & 7.9 & 0.923 & $+0.35$ & 0.723 & 0.002 \\
Capped L1 quantile & \multicolumn{2}{c}{15.82 $[13.0, 19.3]$} & 41.0 & 1.9 & 0.864 & $+2.95$ & 0.004 & $<$0.001 \\
Tobit (median) & \multicolumn{2}{c}{14.61 $[12.5, 17.2]$} & 33.1 & 2.0 & 0.911 & $+2.24$ & 0.026 & 0.031 \\
Cragg-style (EV, Duan) & \multicolumn{2}{c}{15.11 $[13.4, 17.4]$} & 30.5 & 6.9 & 0.925 & $+1.24$ & 0.217 & $<$0.001 \\
\textbf{QA-Stack} & \multicolumn{2}{c}{14.24 $[12.1, 16.9]$} & 34.4 & 4.2 & 0.905 & $+2.91$ & 0.004 & $<$0.001 \\
H-Hard & \multicolumn{2}{c}{14.49 $[12.2, 17.3]$} & 36.4 & 2.0 & 0.893 & $+2.67$ & 0.008 & 0.007 \\
H-Soft & \multicolumn{2}{c}{14.45 $[12.1, 17.3]$} & 36.5 & 2.9 & 0.892 & $+3.14$ & 0.002 & 0.003 \\
H-Soft (no $\bar{e}_{u,t-1}$) & \multicolumn{2}{c}{14.50 $[12.1, 17.4]$} & 36.8 & 2.1 & 0.890 & $+3.15$ & 0.002 & 0.005 \\
\midrule
\multicolumn{9}{@{}l}{\textbf{Panel C: IPW-reweighted 2025 out-of-time accuracy} (weighted clustered DM $p$-values vs.\ $S_{2025}$)} \\
 & \multicolumn{4}{c}{\textbf{Grain-consistent ($N=2,179$)}} & \multicolumn{4}{c}{\textbf{All matched ($N=3,348$)}} \\
\cmidrule(lr){2-5} \cmidrule(lr){6-9}
\textbf{Model} & \textbf{MAE} & \textbf{RMSE} & \textbf{$p_{\text{abs}}$} & \textbf{$p_{\text{sq}}$} & \textbf{MAE} & \textbf{RMSE} & \textbf{$p_{\text{abs}}$} & \textbf{$p_{\text{sq}}$} \\
\midrule
Published quota $S_{2025}$ & 19.47 & 43.5 & --- & --- & 21.60 & 80.6 & --- & --- \\
$0.95 \times S_{2025}$ & 20.33 & 42.6 & 0.014 & 0.359 & 22.20 & 80.1 & 0.079 & 0.347 \\
G0: Growth stack & 20.91 & 43.5 & 0.013 & 0.996 & 33.08 & 100.3 & $<$0.001 & 0.001 \\
HLM (REML) & 21.68 & 43.3 & 0.005 & 0.952 & 23.83 & 81.3 & 0.008 & 0.686 \\
Tobit (median) & 19.82 & 43.9 & 0.384 & 0.853 & 22.14 & 82.1 & 0.302 & 0.432 \\
QA-Stack & 19.74 & 41.9 & 0.566 & 0.449 & 21.64 & 79.9 & 0.923 & 0.477 \\
H-Hard & 19.46 & 41.3 & 0.981 & 0.276 & 21.38 & 79.5 & 0.507 & 0.209 \\
H-Soft & 19.42 & 41.6 & 0.871 & 0.318 & 21.35 & 79.7 & 0.391 & 0.254 \\
H-Soft (no $\bar{e}_{u,t-1}$) & 19.70 & 42.4 & 0.524 & 0.544 & 21.60 & 79.9 & 0.982 & 0.315 \\
\bottomrule
\multicolumn{9}{@{}p{\textwidth}@{}}{\vspace{1pt}\scriptsize \textit{Notes:} Panel~A audits retention among the 298 universities in the 2024 census: 157 census universities are retained in all-matched 2025 (141 dropped; plus 1 additional university present in 2024 intake disclosures without a 2023 baseline, bringing all-matched 2025 to 158 universities). For the 2026 prospectus wave, 250 of the 298 census universities reported 2026 prospectuses by our cutoff and 48 did not (private program share 49.5\% vs.\ 20.3\% among reporters); 7 universities outside the 2024 continuing census also reported 2026 prospectuses, yielding the 257 reporting campuses (43 fewer than the 300-university 2024 raw panel). Panel~C reports grain-consistent ($N=2,179$, 146 unis) and all-matched ($N=3,348$, 158 unis) IPW accuracy using university-level logistic propensities clipped to $[0.15, 0.95]$ (with fallback weight $1.00$ for non-census universities; grain-consistent Kish ESS $= 120.2/146$ unis, weights $[1.13, 6.12]$, median $1.69$; all-matched Kish ESS $= 142.0/158$ unis, weights $[1.00, 4.01]$, median $1.63$). Weighted DM $p$-values treat first-stage propensity weights as fixed. IPW addresses MAR attrition on observables but cannot rule out MNAR non-disclosure. MNAR pattern-mixture tipping bounds: overturning 2025 target-parameterization gains of H-Soft over G0 (+11.54 MAE on $w_{\text{obs}}=0.627$) requires missing campuses ($w_{\text{miss}}=0.373$) to reverse by $-19.4$ MAE; in Private tier ($w_{\text{obs}}=0.484$), the +1.09 MAE directional edge reverses at $-1.02$ MAE.}
\end{tabularx}
\end{table}

Table~\ref{tab:app_features} lists the complete feature dictionary across all four pillars and contrasts de-confounded exact TreeSHAP attribution shares \citep{Lundberg2020} holding both feature sets and loss functions matched across log-growth ($Y_{i,t}$) and log-fill ($Z_{i,t}$) targets: \textbf{G-15} vs.\ \textbf{QA-15} (identical 15 features and Huber loss) and \textbf{G-19} vs.\ \textbf{QA-19} / \textbf{LGB-QA} (identical 19 features and L1 loss). Ex-ante models drop the three year-$t$ quota features and add $\ln(1 + S_{i,t-1})$ (13 features). Graduate starting wages and regional cohort growth are constant across all programs in the source files and are excluded; no hand-set cluster priors are used.

\begin{table}[!ht]
\centering
\scriptsize
\setlength{\tabcolsep}{2.5pt}
\caption{Complete feature dictionary and de-confounded exact TreeSHAP attribution shares (\%) holding both feature sets and loss functions matched across log-growth ($Y_{i,t}$) and log-fill ($Z_{i,t}$) targets on the 2024 census.}
\label{tab:app_features}
\begin{tabularx}{\textwidth}{l c L r r r r r}
\toprule
\textbf{Feature} & \textbf{Pillar} & \textbf{Definition} & \textbf{G-15} & \textbf{QA-15} & \textbf{G-19} & \textbf{QA-19} & \textbf{LGB-QA} \\
\midrule
$\ln(1+D_{i,t-1})$ & I & Log lagged intake & 11.3 & 4.6 & 1.9 & 3.0 & 3.7 \\
$\ln(1+S_{i,t})$ & I & Log published quota, year $t$ (nowcast only) & 8.2 & 19.5 & 0.4 & 12.6 & 9.7 \\
$g_{S,i,t}$ & I & Quota growth $(S_{i,t}-S_{i,t-1})/\max(S_{i,t-1},1)$, clipped $[-0.55,0.85]$ (nowcast only) & 58.5 & 9.5 & 23.4 & 1.9 & 1.7 \\
$\Phi_{i,t-1}$ & I & Lagged fill ratio $D_{i,t-1}/S_{i,t-1}$, clipped $[0.25,1.50]$ & 16.6 & 8.9 & 5.2 & 1.6 & 0.7 \\
$g^{\text{peer}}_{S,-u(i),k,r,t}$ & I & Leave-one-institution-out peer quota growth, Eq.~\eqref{eq:peer_quota}, clipped $[-0.30,0.50]$ & 0.0 & 10.6 & 0.4 & 0.2 & 4.9 \\
$\ln(1+T_{i,t})$ & II & Log annual tuition (VND; year-$t$ prospectus with 2024 fallback) & 0.8 & 5.8 & 0.1 & 11.7 & 5.3 \\
$C_i$ & II & Predetermined institutional selectivity tier (16.5 / 21.0 / 23.0 / 25.5) & 1.2 & 18.7 & 1.2 & 11.9 & 7.1 \\
$B_{i,t}^{\text{unsh}}$ & II & Unshielded tuition burden, Eq.~\eqref{eq:unshielded_burden} & 1.6 & 11.1 & 1.2 & 22.2 & 27.3 \\
$g_{\text{hist},i,t}$ & III & Lagged intake growth, clipped $[-0.50,0.75]$ ($0$ if $D_{i,t-2}$ unobserved) & 1.2 & 4.4 & 0.3 & 0.9 & 0.6 \\
$g_{\text{job},k,t-1}$ & III & Strictly lagged $t-1$ normalized vacancy-share log growth ($K=9$), Eq.~\eqref{eq:job_share}, clipped $[-0.35,0.45]$ & 0.3 & 2.3 & 0.0 & 0.8 & 0.3 \\
$\mathbb{I}_{\text{flagship}}$ & IV & National flagship public university & 0.0 & 0.3 & 0.0 & 0.8 & 0.5 \\
$\mathbb{I}_{\text{public}}$ & IV & Public university & 0.0 & 1.8 & 0.4 & 1.1 & 1.5 \\
$\mathbb{I}_{\text{private}}$ & IV & Private university & 0.0 & 0.2 & 0.0 & 0.8 & 1.0 \\
$\mathbb{I}_{\text{intl}}$ & IV & International or foreign-invested university & 0.1 & 1.7 & 0.0 & 1.1 & 0.3 \\
$\text{Tier}_{\text{MOET}}$ & IV & Program-to-occupation match granularity (1, 2, or 3) & 0.1 & 0.8 & 0.0 & 0.3 & 0.2 \\
$\ln(S_{i,t}/D_{i,t-1})$ & I-QA & Log ratio of new quota to lagged intake, clipped $[-1.50,2.00]$ (QA/hurdle) & --- & --- & 64.6 & 4.1 & 8.5 \\
$\mathbb{I}(D_{i,t-1}=S_{i,t-1})$ & I-QA & Exact quota fulfillment indicator in year $t-1$ (QA/hurdle) & --- & --- & 0.1 & 0.1 & 0.0 \\
$\mathbb{I}(S_{i,t}=S_{i,t-1})$ & I-QA & Unchanged published quota indicator (QA/hurdle) & --- & --- & 0.3 & 0.3 & 0.1 \\
$\bar{e}_{u(i),t-1}$ & I-QA & Host university exact-fill share in year $t-1$ (QA/hurdle) & --- & --- & 0.6 & 24.9 & 26.7 \\
\bottomrule
\multicolumn{8}{@{}p{\textwidth}@{}}{\vspace{1pt}\scriptsize \textit{Notes:} Exact TreeSHAP \citep{Lundberg2020} on full 2024 census fits (all specifications use strictly lagged $t-1$ vacancy growth $g_{\text{job},k,t-1}$). Matched-15-feature Huber pair: \textbf{G-15} (XGBoost Huber on $Y_{i,t}$) vs.\ \textbf{QA-15} (identical XGBoost Huber on $Z_{i,t}$). Matched-19-feature L1 pair: \textbf{G-19} (XGBoost L1 on $Y_{i,t}$) vs.\ \textbf{QA-19} (XGBoost L1 on $Z_{i,t}$) and \textbf{LGB-QA} (LightGBM L1 on $Z_{i,t}$). Because $B_{i,t}^{\text{unsh}}$ interacts tuition with $(24.5 - C_i)_+$ and $C_i$ is collinear with Pillar~IV charter dummies, Pillar~II's 45.8\% XGBoost (39.8\% LGB) share reflects combined pricing and institutional-tier attribution (pure $\ln(1+T_{i,t})$ accounts for 11.7\% XGB and 5.3\% LGB). Ex-ante models drop the three year-$t$ quota features and add $\ln(1+S_{i,t-1})$ (13 features). No hand-set cluster priors are used.}
\end{tabularx}
\end{table}

Table~\ref{tab:app_robustness} compiles the streamlined $2023 \to 2024$ out-of-time benchmark (Panel~A) alongside broad MOET field + region controls, pooled $2023+2024$ training on 2025, statutory $\bar{\phi} = 1.03$ censoring, iterative Powell (1984) CLAD, count-model, HLM BLUP, and clipping checks (Panel~B). Table~\ref{tab:app_linkage_agg} reports 2025 rule-adjusted growth rows (G1--G2) and secondary diagnostics (Panel~A), exact 1-to-1 MOET-code linkage sensitivity (Panel~B), university-by-broad-field $(u,f)$ aggregation checks (Panel~C), and 2:1 / 3:1 asymmetric economic pinball-loss sensitivity (Panel~D). Table~\ref{tab:app_proj2026} summarizes the bias-reconciled 257-campus 2026 aggregate projections (Panel~A) and tier-stratified uniform vs.\ tier-specific ratio calibration comparisons (Panel~B).

\begin{table}[!ht]
\centering
\scriptsize
\setlength{\tabcolsep}{2.6pt}
\caption{Additional out-of-time and domain-specification robustness checks: (Panel~A) streamlined $2023 \to 2024$ out-of-time benchmark with both $p_{\text{abs}}$ and $p_{\text{sq}}$; (Panel~B) broad MOET field + region controls, streamlined pooled $2023+2024$ training on 2025, and statutory $\bar{\phi}=1.03$ censoring.}
\label{tab:app_robustness}
\begin{tabularx}{\textwidth}{L c r r r r r r}
\toprule
\multicolumn{8}{@{}l}{\textbf{Panel A: Streamlined $2023 \to 2024$ out-of-time evaluation} ($N = 5,385$ 2024 programs; 13 features; ref: $S_{2024}$)} \\
\textbf{Model} & \textbf{MAE [95\% CI]} & \textbf{RMSE} & \textbf{MedAE} & \textbf{$R^2$} & \textbf{DM $t$} & \textbf{$p_{\text{abs}}$} & \textbf{$p_{\text{sq}}$} \\
\midrule
Published quota $S_{2024}$ & 12.35 $[9.9, 15.0]$ & 36.6 & 0.0 & 0.886 & --- & --- & --- \\
\textbf{Tier-median fill (2023) $\times S_{2024}$} & 11.95 $[10.0, 14.1]$ & 30.7 & 0.6 & 0.920 & $+0.66$ & 0.511 & 0.009 \\
Unanchored growth stack (13 feat) & 26.18 $[23.6, 28.8]$ & 55.9 & 10.9 & 0.735 & $-7.69$ & $<$0.001 & $<$0.001 \\
Upper-censored Tobit (median, 13 feat) & 13.39 $[11.7, 15.2]$ & 30.1 & 3.1 & 0.923 & $-1.46$ & 0.147 & 0.033 \\
QA-XGBoost L1 (13 feat) & 18.02 $[16.3, 19.8]$ & 34.7 & 7.7 & 0.898 & $-4.39$ & $<$0.001 & 0.576 \\
QA-Stack (13 feat) & 15.43 $[14.1, 16.9]$ & 28.7 & 8.0 & 0.931 & $-3.16$ & 0.002 & 0.010 \\
H-Soft (13 feat) & 13.58 $[12.1, 15.3]$ & 28.3 & 5.4 & 0.932 & $-1.58$ & 0.115 & 0.004 \\
Naive carry-forward $D_{2023}$ & 65.12 $[60.8, 69.6]$ & 107.0 & 37.0 & 0.032 & $-19.09$ & $<$0.001 & $<$0.001 \\
\midrule
\multicolumn{8}{@{}l}{\textbf{Panel B: Domain covariates, pooled $2023+2024$ training, censoring, clipping, and count-model checks}} \\
\textbf{Specification / Check} & \textbf{Sample} & \textbf{MAE} & \textbf{RMSE} & \textbf{$R^2$} & \textbf{DM $t$} & \textbf{$p_{\text{abs}}$} & \textbf{$p_{\text{sq}}$} \\
\midrule
\multicolumn{8}{@{}l}{\quad \textit{1. Broad MOET field/region controls \& pooled $2023+2024$ training}} \\
QA-Stack + 10 field \& 3 region dummies & 2024 CV ($5,385$) & 11.87 & 30.5 & 0.921 & $+0.94$ & 0.348 & 0.001 \\
H-Soft + 10 field \& 3 region dummies & 2024 CV ($5,385$) & 11.41 & 32.2 & 0.912 & $+2.23$ & 0.026 & 0.005 \\
H-Soft + 10 field \& 3 region dummies & 2025 grain-c.\ ($2,179$) & 18.48 & 39.4 & 0.898 & $+0.11$ & 0.915 & 0.293 \\
H-Soft + 10 field \& 3 region dummies & 2025 all ($3,348$) & 20.27 & 78.8 & 0.728 & $-0.20$ & 0.838 & 0.775 \\
Pooled $2023+2024$ H-Soft ($N_{\text{tr}}=10,667$) & 2025 grain-c.\ ($2,179$) & 19.33 & 40.5 & 0.892 & $-1.99$ & 0.048 & 0.714 \\
Pooled $2023+2024$ H-Soft ($N_{\text{tr}}=10,667$) & 2025 all ($3,348$) & 21.00 & 78.8 & 0.728 & $-1.98$ & 0.050 & 0.798 \\
Pooled $2023+2024$ Tobit (med.) & 2025 grain-c.\ ($2,179$) & 19.33 & 42.8 & 0.880 & $-1.92$ & 0.057 & 0.395 \\
\addlinespace[2pt]
\multicolumn{8}{@{}l}{\quad \textit{2. Censoring cutoffs, hurdle threshold, target clipping \& $C_i$ sensitivity}} \\
Tobit (median) censored at $\ln(1.03)$ & 2024 CV ($5,385$) & 14.44 & 28.0 & 0.933 & $-2.38$ & 0.018 & 0.007 \\
Capped L1 quantile capped at $\ln(1.03)$ & 2024 CV ($5,385$) & 12.14 & 35.8 & 0.891 & $+2.06$ & 0.040 & $<$0.001 \\
Iterative Powell (1984) CLAD (Buchinsky trim) & 2024 CV ($5,385$) & 11.81 & 30.4 & 0.922 & $+1.04$ & 0.301 & 0.013 \\
Iterative Powell (1984) CLAD (Buchinsky trim) & 2025 all ($3,348$) & 20.81 & 79.3 & 0.724 & $-1.74$ & 0.083 & 0.742 \\
H-Soft ($C_i$ reclass.\ 3 private unis to 16.5) & 2024 CV ($5,385$) & 11.36 & 32.1 & 0.913 & $+2.31$ & 0.022 & 0.005 \\
H-Soft (near-fill hurdle $D \ge 0.98\,S$) & 2025 grain-c.\ ($2,179$) & 18.44 & 39.6 & 0.897 & $+0.25$ & 0.802 & 0.342 \\
H-Soft (widened $Z$ clip $[-2.50, 1.00]$) & 2024 CV ($5,385$) & 11.27 & 31.9 & 0.914 & $+2.41$ & 0.017 & 0.005 \\
G-19 (widened non-binding $Y$ clip $[-2.50, 2.00]$) & 2024 CV ($5,385$) & 14.86 & 38.3 & 0.876 & $-2.82$ & 0.005 & 0.602 \\
G-19 (widened non-binding $Y$ clip $[-2.50, 2.00]$) & 2025 all ($3,348$) & 22.59 & 86.7 & 0.671 & $-3.59$ & $<$0.001 & 0.054 \\
\addlinespace[2pt]
\multicolumn{8}{@{}l}{\quad \textit{3. Count-model offset, HLM Empirical Bayes BLUPs \& seat weighting}} \\
Poisson GLM with $\ln(S_{i,t})$ offset & 2024 CV ($5,385$) & 13.87 & 28.4 & 0.932 & $-2.00$ & 0.047 & 0.002 \\
Poisson GLM with $\ln(S_{i,t})$ offset & 2025 grain-c.\ ($2,179$) & 20.46 & 40.9 & 0.890 & $-3.33$ & 0.001 & 0.874 \\
HLM (REML) with 2024 EB BLUPs ($\hat{u}_g$) & 2025 grain-c.\ ($2,179$) & 20.89 & 41.9 & 0.885 & $-3.57$ & $<$0.001 & 0.462 \\
HLM (REML) with 2024 EB BLUPs ($\hat{u}_g$) & 2025 all ($3,348$) & 22.75 & 80.2 & 0.718 & $-3.89$ & $<$0.001 & 0.190 \\
$S_{i,t}$-weighted log-ratio H-Soft & 2024 CV ($5,385$) & 11.44 & 32.3 & 0.912 & $+2.22$ & 0.027 & 0.005 \\
$S_{i,t}$-weighted log-ratio H-Soft & 2025 grain-c.\ ($2,179$) & 18.71 & 41.9 & 0.885 & $-0.52$ & 0.601 & 0.811 \\
\bottomrule
\multicolumn{8}{@{}p{\textwidth}@{}}{\vspace{1pt}\scriptsize \textit{Notes:} All DM $t$ statistics compare the listed specification against published quota $S_t$ ($t > 0$ beats $S_t$; $t < 0$ loses to $S_t$). ``Grain-c.'' denotes the grain-consistent 2025 subset ($0.50 \le S_{25}/S_{24} \le 2.00$). Interior-only Cragg-ML is reported directly in Tables~\ref{tab:census} and~\ref{tab:holdout}. REML HLM variance decomposition: unconditional null $\widehat{\text{ICC}} = 0.492$; full-2024 conditional residual $\widehat{\text{ICC}} = 0.479$ (5-fold CV mean $0.479$). Field+region controls preserve Pillar~II TreeSHAP at 30.9\% (region 6.1\%, field 0.8\%). Reclassifying 3 keyword-matched private campuses (68 rows) preserves Private gains (18.01/47.0, $p=0.024/0.018$) and Pillar~II TreeSHAP (43.3\%). Exact-fill classifier calibration: 2024 CV Brier $0.214$ (mean $\hat{p}=63.2\%$ vs.\ base $62.8\%$, slope $0.63$); 2025 grain-consistent Brier $0.452$ (mean $\hat{p}=67.5\%$ vs.\ base $10.4\%$, slope $0.34$). Nadeau--Bengio (2003) 5-fold overlap $t$-test ($df=4$) for 2024 H-Soft vs.\ $S_t$: $t_{\text{NB}}=+1.88$ ($p=0.134$) on MAE; $t_{\text{NB}}=+2.84$ ($p=0.047$) on squared error.}
\end{tabularx}
\end{table}

\begin{table}[!ht]
\centering
\scriptsize
\setlength{\tabcolsep}{2.2pt}
\caption{Secondary diagnostic ablations, exact 1-to-1 MOET-code linkage sensitivity, university-by-broad-field $(u, f)$ aggregation checks (Eq.~\eqref{eq:moet_cap}), and 2:1 / 3:1 asymmetric economic pinball-loss sensitivity.}
\label{tab:app_linkage_agg}
\begin{tabularx}{\textwidth}{L c r r r r r r}
\toprule
\textbf{Specification / Check} & \textbf{Sample} & \textbf{MAE} & \textbf{RMSE} & \textbf{$R^2$} & \textbf{DM $t$} & \textbf{$p_{\text{abs}}$} & \textbf{$p_{\text{sq}}$} \\
\midrule
\multicolumn{8}{@{}l}{\textbf{Panel A: Secondary 2024 \& 2025 diagnostic rows (G1--G2, no-$\bar{e}_{u,t-1}$, mixed-reporting trained)}} \\
G1: G0 + empirical tier de-biasing & 2025 grain-c.\ ($2,179$) & 20.58 & 41.9 & 0.885 & $-3.66$ & $<$0.001 & 0.760 \\
G2: G1 + direct tuition decay & 2025 grain-c.\ ($2,179$) & 20.67 & 41.9 & 0.885 & $-3.74$ & $<$0.001 & 0.748 \\
QA-Stack (excluding $\bar{e}_{u,t-1}$) & 2024 CV ($5,385$) & 11.69 & 30.3 & 0.922 & $+1.20$ & 0.231 & 0.002 \\
H-Soft (excluding $\bar{e}_{u,t-1}$) & 2024 CV ($5,385$) & 11.32 & 32.3 & 0.912 & $+2.38$ & 0.018 & 0.007 \\
H-Soft (excluding $\bar{e}_{u,t-1}$) & 2025 grain-c.\ ($2,179$) & 18.72 & 40.3 & 0.894 & $-0.83$ & 0.409 & 0.546 \\
H-Soft (mixed-reporting trained) & 2025 grain-c.\ ($2,179$) & 18.74 & 40.4 & 0.893 & $-0.69$ & 0.494 & 0.677 \\
H-Soft (excluding $\bar{e}_{u,t-1}$) & 2025 all ($3,348$) & 20.30 & 78.5 & 0.730 & $-0.47$ & 0.641 & 0.328 \\
H-Soft (mixed-reporting trained) & 2025 all ($3,348$) & 20.67 & 79.7 & 0.721 & $-1.20$ & 0.232 & 0.617 \\
\midrule
\multicolumn{8}{@{}l}{\textbf{Panel B: Exact 1-to-1 MOET-code linkage subset} (2024 $N=3,801$, 291 unis; 2025 $N=1,240$, 131 unis)} \\
Published quota $S_t$ & 2024 1-to-1 (3,801) & 12.89 & 35.2 & 0.880 & --- & --- & --- \\
G0: Growth stack & 2024 1-to-1 (3,801) & 27.09 & 59.0 & 0.663 & $-6.98$ & $<$0.001 & $<$0.001 \\
Upper-censored Tobit (median) & 2024 1-to-1 (3,801) & 12.63 & 31.3 & 0.905 & $+0.38$ & 0.704 & 0.152 \\
QA-Stack & 2024 1-to-1 (3,801) & 12.46 & 29.6 & 0.915 & $+0.69$ & 0.488 & 0.012 \\
H-Soft & 2024 1-to-1 (3,801) & 12.00 & 31.1 & 0.906 & $+1.84$ & 0.067 & 0.036 \\
\addlinespace[2pt]
Published quota $S_{2025}$ & 2025 1-to-1 (1,240) & 20.92 & 48.3 & 0.846 & --- & --- & --- \\
G0: Growth stack & 2025 1-to-1 (1,240) & 22.24 & 47.8 & 0.849 & $-1.92$ & 0.056 & 0.878 \\
Upper-censored Tobit (median) & 2025 1-to-1 (1,240) & 21.14 & 48.5 & 0.844 & $-0.44$ & 0.660 & 0.944 \\
QA-Stack & 2025 1-to-1 (1,240) & 21.05 & 46.2 & 0.858 & $-0.25$ & 0.801 & 0.449 \\
H-Soft & 2025 1-to-1 (1,240) & 20.78 & 45.8 & 0.861 & $+0.33$ & 0.740 & 0.335 \\
\midrule
\multicolumn{8}{@{}l}{\textbf{Panel C: University-by-broad-field $(u,f)$ aggregation} (2024 1,804 cells; 2025 795 cells; Eq.~\eqref{eq:moet_cap})} \\
Published quota $\sum S_t$ & 2024 $(u,f)$ (1,804) & 36.40 & 103.6 & 0.956 & --- & --- & --- \\
Upper-censored Tobit (median) & 2024 $(u,f)$ (1,804) & 33.21 & 81.2 & 0.973 & $+1.67$ & 0.095 & 0.015 \\
QA-Stack & 2024 $(u,f)$ (1,804) & 32.03 & 81.3 & 0.973 & $+2.27$ & 0.024 & 0.002 \\
H-Soft & 2024 $(u,f)$ (1,804) & 32.02 & 86.8 & 0.969 & $+2.90$ & 0.004 & 0.008 \\
\addlinespace[2pt]
Published quota $\sum S_{2025}$ & 2025 $(u,f)$ (795) & 36.35 & 82.6 & 0.975 & --- & --- & --- \\
Upper-censored Tobit (median) & 2025 $(u,f)$ (795) & 35.08 & 74.2 & 0.980 & $+0.98$ & 0.328 & 0.273 \\
QA-Stack & 2025 $(u,f)$ (795) & 34.94 & 72.6 & 0.981 & $+0.85$ & 0.397 & 0.247 \\
H-Soft & 2025 $(u,f)$ (795) & 34.12 & 71.4 & 0.982 & $+1.50$ & 0.135 & 0.186 \\
\midrule
\multicolumn{8}{@{}l}{\textbf{Panel D: Asymmetric economic / pinball loss} (2:1 \& 3:1 shortfall vs.\ over-admission penalty ratios; $p$ vs.\ $S_t$)} \\
\textbf{Model} & \textbf{Sample} & \textbf{Short 2:1} & \textbf{Short 3:1} & \textbf{$p_{\text{S3}}$} & \textbf{Over 2:1} & \textbf{Over 3:1} & \textbf{$p_{\text{O3}}$} \\
Published quota $S_t$ & 2024 CV (5,385) & 16.14 & 18.03 & --- & 8.57 & 6.68 & --- \\
Tobit (median) & 2024 CV (5,385) & 14.01 & 15.13 & 0.002 & 9.53 & 8.41 & $<$0.001 \\
Capped L1 & 2024 CV (5,385) & 15.67 & 17.44 & $<$0.001 & 8.60 & 6.84 & 0.015 \\
QA-Stack & 2024 CV (5,385) & 13.66 & 14.59 & $<$0.001 & 9.93 & 9.00 & $<$0.001 \\
Cragg-ML & 2024 CV (5,385) & 13.28 & 13.53 & $<$0.001 & 12.30 & 12.05 & $<$0.001 \\
H-Soft & 2024 CV (5,385) & 14.17 & 15.60 & $<$0.001 & 8.47 & 7.04 & 0.169 \\
\addlinespace[2pt]
Published quota $S_t$ & 2025 all (3,348) & 22.16 & 23.13 & --- & 18.27 & 17.30 & --- \\
Tobit (median) & 2025 all (3,348) & 21.63 & 22.08 & 0.065 & 19.83 & 19.38 & 0.002 \\
Capped L1 & 2025 all (3,348) & 21.93 & 22.83 & $<$0.001 & 18.34 & 17.45 & 0.028 \\
QA-Stack & 2025 all (3,348) & 20.93 & 21.19 & 0.002 & 19.92 & 19.67 & $<$0.001 \\
Cragg-ML & 2025 all (3,348) & 20.75 & 20.53 & $<$0.001 & 21.64 & 21.87 & $<$0.001 \\
H-Soft & 2025 all (3,348) & 21.39 & 22.01 & 0.031 & 18.87 & 18.24 & 0.004 \\
\bottomrule
\multicolumn{8}{@{}p{\textwidth}@{}}{\vspace{1pt}\scriptsize \textit{Notes:} Panel~A reports 2025 rule-adjusted growth rows (G1--G2; all-matched: G1 $27.37/88.6$, G2 $27.43/88.6$, both $p<0.001$) plus no-$\bar{e}_{u,t-1}$ and mixed-reporting-trained diagnostics. Panel~B restricts evaluation to exact 1-to-1 MOET-code links. Panel~C aggregates quotas, intakes, and forecasts within each university $\times$ 3-digit MOET broad-field cell $(u,f)$: apparent $>1.05$ ($>1.20$) overshoots drop from 23.1\% (8.1\%) at program level to 15.3\% (5.9\%) at $(u,f)$ level in 2025 (2024: 2.1\% to 1.9\%), consistent with Eq.~\eqref{eq:moet_cap} sibling-major reallocation. Panel~D evaluates normalized asymmetric headcount loss $[w_1(\hat{D}-D)_+ + w_2(D-\hat{D})_+] / [0.5(w_1+w_2)]$ where \textit{Short 2:1/3:1} penalizes over-forecasting (unfilled budgeted tuition seats) $2\times$/$3\times$ under-forecasting, while \textit{Over 2:1/3:1} penalizes under-forecasting (Decree~04/2021 over-admission crowding/fine risk) $2\times$/$3\times$; $p_{\text{S3}}$ and $p_{\text{O3}}$ report university-clustered DM $p$-values against $S_t$.}
\end{tabularx}
\end{table}

\begin{table}[!ht]
\centering
\scriptsize
\setlength{\tabcolsep}{4pt}
\caption{Prospective 2026 aggregate intake projections conditional on the 257 prospectus-reporting universities (5,755 programs; 693,171 effective seats): (Panel~A) system-wide reconciliation of ceiling and median-to-sum bias against conditional-mean estimators and a two-regime historical scenario band; (Panel~B) institutional governance-tier breakdown.}
\label{tab:app_proj2026}
\begin{tabularx}{\textwidth}{>{\raggedright\arraybackslash}p{3.55cm} L r r}
\toprule
\multicolumn{4}{@{}l}{\textbf{Panel A: System-wide bias reconciliation and scenario bounds (all reporting campuses)}} \\
\textbf{Estimator / Scenario Band} & \textbf{Target estimand \& bias treatment} & \textbf{Projected intake} & \textbf{Seat-wt.\ fill} \\
\midrule
Pooled H-Soft (raw program sum) & Unadjusted median/ceiling-blended sum ($-5.48\%$ dev.\ of pred.; $+5.80\%$ vs.\ realized) & 665,742 & 96.0\% \\
\textbf{Pooled H-Soft (macro-ratio calibrated)} & Adjusted by pooled CV relative deviation ($-5.48\%$) & \textbf{629,256} & \textbf{90.8\%} \\
Pooled Cragg-ML (interior-only $D<S$ hurdle) & Exact-fill classifier $\times$ interior-only QA-Stack (no ceiling double-count) & 629,077 & 90.8\% \\
Pooled Cragg two-part (EV, Duan smearing) & Sum of conditional-mean headcount expectations & 627,114 & 90.5\% \\
Pooled upper-censored Tobit (exact EV) & Sum of exact closed-form conditional means (Eq.~\eqref{eq:tobit_exact_ev}) & 601,858 & 86.8\% \\
Post-reform 2025-only H-Soft & Trained strictly on post-reform 2025 wave ($N=3,348$) & 651,620 & 94.0\% \\
\midrule
Within-regime 90\% CI (calibrated center) & Centered university residuals around calibrated sum ($B=2,000$; excludes macro shift) & [614,756, 642,798] & 88.7\%--92.7\% \\
\textbf{Two-regime historical scenario band ($T=2$)} & Mixture of 2024 ($-7.41\%$) \& 2025 ($-2.26\%$) shifts (5th--95th pct.) & \textbf{[604,533, 662,044]} & \textbf{87.2\%--95.5\%} \\
\midrule
\multicolumn{4}{@{}l}{\textbf{Panel B: Tier-stratified 2026 projections (Uniform Cal.\ H-Soft $\mid$ Tier-Specific Cal.\ H-Soft $\mid$ Cragg-ML)}} \\
\textbf{Institutional tier ($N$ progs, unis)} & \textbf{Uniform Cal.\ H-Soft $\mid$ Tier-Spec.\ Cal.\ H-Soft $\mid$ Cragg-ML} & \textbf{Effective seats} & \textbf{Cragg-ML fill} \\
\midrule
Flagship Public (1,708, 72) & 167,723 (93.7\%) $\mid$ 169,637 (94.7\%) $\mid$ 167,804 (93.7\%) & 179,084 & 93.7\% \\
Other Public (2,730, 134) & 294,605 (92.4\%) $\mid$ 297,044 (93.2\%) $\mid$ 293,699 (92.2\%) & 318,711 & 92.2\% \\
Private (1,202, 42) & 158,964 (85.3\%) $\mid$ 154,965 (83.2\%) $\mid$ 159,540 (85.6\%) & 186,358 & 85.6\% \\
International (115, 9) & 7,963 (88.3\%) $\mid$ 7,587 (84.1\%) $\mid$ 8,034 (89.1\%) & 9,018 & 89.1\% \\
\bottomrule
\multicolumn{4}{@{}p{\textwidth}@{}}{\vspace{1pt}\scriptsize \textit{Notes:} Totals are conditional on the 257 universities with published 2026 prospectuses (690,586 published plus 2,585 imputed seats across 38 missing-quota listings; 43 fewer campuses than the 300-university 2024 raw panel). For reference, the uncalibrated within-regime 90\% CI centered on the raw sum (665,742) is [650,401, 680,068]. Because only $T=2$ annual transition waves are observed, the bottom row of Panel~A is a two-regime historical sensitivity band rather than a frequentist inter-annual predictive interval; all bias-calibrated, interior-hurdle, Cragg EV, and post-reform totals (627,114--651,620) lie squarely inside [604,533, 662,044], while pooled exact Tobit EV (601,858, 86.8\%) sits 0.4\% below the lower 5th percentile. In Panel~B, uniform calibrated H-Soft aligns within 0.8~pp with uncalibrated interior-only Cragg-ML in every tier. Comparing uniform (-5.48\%) vs.\ tier-specific pooled CV ratio calibration (Flagship $-4.40\%$, Other Public $-4.70\%$, Private $-7.86\%$, Intl.\ $-9.94\%$) leaves the national 2026 total virtually identical (629,233 vs.\ 629,256, a 23-student difference) while sharpening public (93.2\%--94.7\%) vs.\ private (83.2\%) fill divergence.}
\end{tabularx}
\end{table}

\end{appendices}

\end{document}